\PassOptionsToPackage{table,xcdraw}{xcolor}
\documentclass[10pt,compsoc]{IEEEtran}

\makeatletter \def\mdseries@tt{m} \makeatother 

\usepackage{makecell}
\usepackage{tcolorbox}
\usepackage{indentfirst}
\usepackage{booktabs}
\usepackage{graphicx}
\graphicspath{./figures}
\usepackage[singlespacing]{setspace}
\usepackage{threeparttable}
\usepackage{multirow}
\usepackage{subcaption}
\usepackage[utf8]{inputenc}
\usepackage[english]{babel}
\usepackage{amsthm}
\usepackage{changepage} 
\usepackage{mathtools}
\usepackage{textgreek}
\usepackage[compact]{titlesec}         
\usepackage{amsmath}
\usepackage{tikz}
\usetikzlibrary{shapes,positioning,arrows,calc}
\usepackage{multirow}
\usepackage[table,xcdraw]{xcolor}

\DeclareFontFamily{OT1}{pzc}{}
\DeclareFontShape{OT1}{pzc}{m}{it}{<-> s * [1.1] pzcmi7t}{}
\DeclareMathAlphabet{\mathpzc}{OT1}{pzc}{m}{it}
\theoremstyle{definition}
\newtheorem{definition}{Definition}
\DeclareMathAlphabet\mathbfcal{OMS}{cmsy}{b}{n}
\usepackage{centernot}
\usepackage{enumitem}
\makeatletter
\def\algbackskip{\hskip-\ALG@thistlm}
\usepackage{xcolor}
\usepackage{orcidlink}

\usepackage[ruled, vlined, linesnumbered]{algorithm2e}
\newcommand{\nonl}{\renewcommand{\nl}{\let\nl\oldnl}}
\SetKwInput{KwInput}{Input}                
\SetKwInput{KwOutput}{Output} 
\SetAlFnt{\small\sffamily}

\SetCommentSty{mycommfont}
\usepackage{amsmath, listings, amsthm, amssymb, proof, xspace}
\usepackage{hyperref}
\hypersetup{colorlinks,allcolors=black}
\theoremstyle{remark}

\newcommand{\added}[1]{\textcolor{black}{{#1}}}
\newcommand{\minor}[1]{\textcolor{black}{{#1}}}

\lstdefinelanguage{ocl1}{
  keywords={init,Message, Pre, Post, Client,Range,Invariant,Constraints},
  keywordstyle=\color{purple}\ttfamily\bfseries,
  keywords=[2]{not, and, or, state, transition, implies, component,dbg,receipt,:,|,;},
  keywordstyle=[2]\color{black}\ttfamily\bfseries,
  sensitive=false,
  comment=[l]{//},
  morecomment=[s]{/*}{*/},
  commentstyle=\color{blue}\ttfamily,
  stringstyle=\color{red}\ttfamily,
  morestring=[b]',
  morestring=[b]"
}

\lstdefinelanguage{antlr}{
  keywords={grammar, rule,ExecRule,EOF},
  keywordstyle=\color{Purple}\ttfamily\bfseries\footnotesize,
  keywords=[2]{script:, execRule:,when:, where:, statements:, scriptStatement:,interactiveStatement:,umlrtCmd:,random:,dbgCommands:,|},
  keywordstyle=[2]\color{black}\ttfamily\bfseries\footnotesize,
  otherkeywords = {-,|},
  morekeywords = [3]{-},
  morekeywords = [4]{|},
  keywordstyle = [3]{\color{blue}}\footnotesize,
  keywordstyle = [4]{\color{blue}}\footnotesize,
  identifierstyle=\color{black}\footnotesize,
  sensitive=true,
  comment=[l]{//},
  morecomment=[s]{/*}{*/},
  commentstyle=\color{green}\ttfamily,
  stringstyle=\color{blue}\ttfamily,
  morestring=[b]',
  morestring=[b]",
  alsodigit={:}
}

\lstdefinelanguage{z3}{
	sensitive=true,
	alsoletter={\-},
	comment=[l]{;},
	keywords=[1]{
apply, assert, assert-soft, check-sat, check-sat-using, compute-interpolant,
declare-const, declare-datatypes, declare-fun, declare-map, declare-rel,
declare-sort, declare-tactic, define-sort, display, echo, eval, exit,
fixedpoint-pop, fixedpoint-push, get-assertions, get-assignment, get-info, get-
interpolant, get-model, get-option, get-proof, get-unsat-core, get-user-tactics,
get-value, help, help-tactic, labels, maximize, minimize, pop, push, query,
reset, rule, set-info, set-logic, set-option, simplify
	},
	morekeywords=[2]{
check-sat-using, declare-var, declare-rel, rule, query, set-predicate-
representation, maximize, minimize, assert-soft, assert-weighted, compute-
interpolant
	},
}

\author{Amirhossein~Zolfagharian~\orcidlink{0000-0002-2411-7938}, Manel~Abdellatif~\orcidlink{0000-0002-8647-1676}, Lionel~C.~Briand~\orcidlink{0000-0002-1393-1010}, and~Ramesh~S~\orcidlink{S0000-0002-8501-7447}
\IEEEcompsocitemizethanks{\IEEEcompsocthanksitem Amirhossein Zolfagharian is affiliated with the School of Electrical Engineering and Computer Science (EECS), University of Ottawa, Ottawa, Canada.\protect\\
E-mail: A.zlf@uottawa.ca
\IEEEcompsocthanksitem Manel Abdellatif is with the Software and Information Technology Engineering Department, École de Technologie Supérieure, Montreal, Canada.\protect\\
E-mail: Manel.abdellatif@etsmtl.ca
\IEEEcompsocthanksitem Lionel Briand is with the School of Electrical Engineering and Computer Science (EECS), University of Ottawa, Ottawa, Canada, and also with the Lero SFI Research Center and University of Limerick, Ireland.\protect\\
E-mail: Lbriand@uottawa.ca
\IEEEcompsocthanksitem Ramesh S is with the Department of Research and Development, General
Motors, Warren, MI, USA.\protect\\
E-mail: Ramesh.s@gm.com
}}

\begin{document}

\title{SMARLA: A Safety Monitoring Approach for\\ Deep Reinforcement Learning Agents}

\IEEEtitleabstractindextext{%




\begin{abstract}
Deep Reinforcement Learning (DRL) has made significant advancements in various fields, such as autonomous driving, healthcare, and robotics, by enabling agents to learn optimal policies through interactions with their environments. However, the application of DRL in safety-critical domains presents challenges, particularly concerning the safety of the learned policies. DRL agents, which are focused on maximizing rewards, may select unsafe actions, leading to safety violations. Runtime safety monitoring is thus essential to ensure the safe operation of these agents, especially in unpredictable and dynamic environments.
This paper introduces \textit{SMARLA}, a black-box safety monitoring approach specifically designed for DRL agents. \textit{SMARLA} utilizes machine learning to predict safety violations by observing the agent's behavior during execution. The approach is based on Q-values, which reflect the expected reward for taking actions in specific states.
\textit{SMARLA} employs state abstraction to reduce the complexity of the state space, enhancing the predictive capabilities of the monitoring model. Such abstraction enables the early detection of unsafe states, allowing for the implementation of corrective and preventive measures before incidents occur.
We quantitatively and qualitatively validated \textit{SMARLA} on three well-known case studies widely used in DRL research. Empirical results reveal that \textit{SMARLA} is accurate at predicting safety violations, with a low false positive rate, and can predict violations at an early stage, approximately halfway through the execution of the agent, before violations occur. We also discuss different decision criteria, based on confidence intervals of the predicted violation probabilities, to trigger safety mechanisms aiming at a trade-off between early detection and low false positive rates.
\end{abstract}

}

\maketitle
\ifCLASSOPTIONcompsoc
\IEEEraisesectionheading{\section{Introduction}\label{Sec:Introduction}}
\else
\section{Introduction}
\label{Sec:Introduction}
\fi

\IEEEPARstart{R}{einforcement~Learning} (RL) has gained significant importance in addressing real-world challenges in different application domains such as autonomous driving, healthcare, and robotics. RL algorithms enable agents to learn optimal policies through interaction with their environments, aiming to optimize long-term cumulative rewards. Deep Reinforcement Learning (DRL) is an extension of RL where optimal policies are learned with Deep Neural Networks (DNNs). One of the advantages of DRL is its ability to handle high-dimensional inputs, allowing agents to learn directly based on raw data from the environment without the need for manual feature engineering~\cite{meng2021memory,arulkumaran2017deep}. 

While DRL has shown great promise, the increasing complexity and deployment of DRL agents in safety-critical domains have raised concerns regarding their safety. Indeed, one of the main challenges in DRL is the lack of guarantees on the safe behavior of the learned policies~\cite{turchetta2020safe,dulac2021challenges}. Since DRL agents focus on maximizing a reward signal, they might violate safety requirements during learning or execution~\cite{alshiekh2018safe} through the selection of unsafe actions. For example, a DRL agent controlling a self-driving vehicle can learn to drive at high speed to maximize the reward and reach its final goal earlier, violating traffic rules and endangering pedestrians in the environment. Moreover, pre-deployment testing of DRL agents, although crucial, remains insufficient for ensuring their safety, as it does not guarantee to uncover all possible corner cases. 

Even when dedicating significant effort to testing DRL agents across various scenarios~\cite{tappler2022search,zolfagharian2023search}, ensuring runtime safety remains challenging due to the extremely large agent state space and number of possible execution scenarios. As a result, the runtime monitoring of RL agents is crucial to guarantee their safety, as recommended by industry standards such as ISO 26262~\cite{iso26262} and ISO/PAS 21448~\cite{iso21448} in the automotive domain.  
\added{While such standards do not explicitly address RL agents, given the increasingly frequent integration of DRL in autonomous systems\minor{~\cite{kiran2021deep,tang2024deep}}, it is essential that such agents adhere to these established safety standards to maintain the safety integrity levels required in many safety-critical domains such as automotive.}

Runtime safety monitoring, however, entails the efficient, continuous observation and risk assessment of the states and actions performed by DRL agents. It must also provide a means to predict early and thus prevent unsafe behavior before it leads to catastrophic consequences. If the predictions of safety violations are sufficiently early, corrective measures or safety mechanisms can indeed be applied in most contexts (e.g., giving the control to a human driver or activating automatic emergency brakes in autonomous driving systems). This is highly important in safety-critical applications, such as autonomous vehicles, where even a single violation can have severe consequences. 
\minor{ 
Standards such as ISO 26262 (automotive), DO-178C (aviation), and IEC 60601 (medical devices) emphasize the need to predict known unsafe situations to achieve acceptable safety levels.} 
Moreover, runtime safety monitoring is essential when DRL agents are deployed in uncertain and dynamic environments. DRL agents often are trained on simulators and deployed in environments where the dynamics can change over time, and the agent's learned policy may no longer be optimal or safe. By monitoring the agents, it is not only possible to ensure safety but also to understand whether further training is required or not. Safety monitoring can also provide insights into the learning process of DRL agents. By analyzing the agent's behavior and the safety violations detected, it is possible to gain a better understanding of the agent's decision-making process and identify areas for improvement.

However, the runtime safety monitoring of DRL agents presents challenges. DRL agents often operate in environments with extremely large state spaces, which can make it challenging to monitor their behavior in real-time and predict potential safety violations~\cite{dulacarnold2019challenges}. Further, DRL agents learn and adapt their behavior based on interactions with the environment, which introduces uncertainty regarding their policies and resulting actions. Such uncertainty can make it difficult to predict and monitor the agents' behavior accurately. Safety monitoring needs to account for it and consider methods that can scale to large state spaces and can handle the dynamic and evolving nature of DRL agents' behavior.

Existing safety monitoring techniques for regular software systems often rely on formal verification to ensure compliance with safety constraints~\cite{708570}. However, when it comes to DRL policies, formally verifying their behavior to satisfy safety properties becomes a computationally expensive and an NP-complete problem~\cite{marchesini2023safe,katz2017reluplex}. 
Further, the black-box nature of DRL policies makes it challenging to analyze and verify them~\cite{dulacarnold2019challenges}. There are three distinct types of safety monitoring for DRL agents, each based on the type of information they require: white-box, grey-box, and black-box. White-box safety monitoring requires comprehensive access to internal information of the DRL agent, including its architecture, weights, reward function, and training dataset. Gray-box safety monitoring requires access to the agent's inputs, outputs, and training dataset.  Black-box safety monitoring, however, limits its analysis to the observed input-output behavior, including the agent's Q-values. Q-values assign numerical values to actions in states, reflecting their potential to lead to a reward or a penalty~\cite{10.1007/978-3-031-47994-6_26}. Monitoring DRL agents using a black-box approach is often practical, as testers and safety engineers usually lack access to the agent's internal structure or training dataset~\cite{zolfagharian2023search,aghababaeyan2023black,aghababaeyan2024deepgd}.


Moreover, most of the existing works on RL safety in the literature have primarily focused on safe exploration strategies to enhance the safety of RL agents during the learning process~\cite{pecka2014safe,okawa2023safe} and RL shielding strategies~\cite{alshiekh2018safe,elsayed2021safe,konighofer2022online} to block unsafe actions at runtime and force the agent to deviate from its policy. 
In contrast to these works, we propose a black-box safety monitoring approach, specifically tailored to DRL agents, where the goal is to predict safety violations at runtime, as early as possible, by monitoring the behavior of the agent.



In this paper, we present \textit{SMARLA}, a Safety Monitoring Approach for Reinforcement Learning Agents. \textit{SMARLA} is a monitoring approach that uses machine learning to predict safety violations, accurately and early,  during the execution of DRL agents. \minor{We train \textit{SMARLA} based on the testing results of DRL agents. This is important in practice, as testing data contain useful information (i.e., both known safe and unsafe situations) that can be exploited for monitoring. } We leverage state abstraction methods~\cite{pmlr-v80-abel18a,jiang2018notes, li2006towards} to reduce the large state space usually associated with DRL agents and thus increase the learnability of machine learning models to predict violations. \added{\textit{SMARLA} currently focuses on Q-learning algorithms. Q-learning is a widely used and common type of RL algorithms~\cite{jang2019q}, where the objective is to find the optimal action-selection policy through Q-values.} However, it is important to note that \textit{SMARLA} operates as a black-box monitoring method, as it does not require access to the internals of the Q-learning-based agent or the training dataset. \added{\textit{SMARLA} only requires access to the Q-values of the agent which represent the expected cumulative reward for taking an action in a given state, providing insight into the agent's decision-making process. As a result, \textit{SMARLA} is also agnostic to the type of DRL agent’s inputs, which is a practical advantage.} 

To evaluate the effectiveness of \textit{SMARLA}, we implemented our safety monitor for \added{three} well-known RL problems that serve as benchmarks in the RL community~\cite{Behzadan2019AdversarialEO,Behzadan2019SequentialTF,pattanaik2018robust,yan2024forgingvisionfoundationmodels,10565991,10.1007/978-3-031-48121-5_18,10077125}. Our experimental results suggest that \textit{SMARLA} can accurately predict safety violations, while maintaining a low false positive rate, long before violation occurrences. 
We thus provide evidence of the potential benefits of \textit{SMARLA} as it is a promising solution to prevent damages and mitigate risks associated with RL agents. The main contributions of our work are as follows:

\begin{itemize} 
    \item \added{ We introduce \textit{SMARLA}, a novel black-box safety monitoring approach specifically designed for DRL agents. \textit{SMARLA} is black-box as it only requires access to agents' Q-values as a proxy of its decision-making process. Based on these Q-values, we apply state abstraction to reduce the state space and increase the learnability of a model used to predict safety violations. \textit{SMARLA} therefore aims to predict DRL safety violations at an early stage, enabling proactive and preventive mechanisms for ensuring the safety of DRL-based systems.} 
    \item As part of \textit{SMARLA}, we propose a highly accurate machine learning solution to estimate the probability of safety violations at each time step of the DRL agent's execution, along with confidence intervals.
    \item We investigate alternative decision procedures, based on the confidence intervals of violation probability estimates, in terms of both accuracy and decision time.
    \item We implement \textit{SMARLA} on \added{three} widely used RL benchmark problems and investigate trade-offs between early and accurate predictions of safety violations. Results show the effectiveness of \textit{SMARLA} in accurately predicting safety violations and demonstrate its potential for real-world applications.
    \item \added{We conducted a qualitative analysis to provide insights into SMARLA's practical application and explain its prediction results, offering an understanding of the conditions under which the risk increases.} 
    \item We provide online~\cite{Replication} a prototype tool for our safety monitoring solution. We also include all the necessary data and configurations to replicate our experiments and results. 
\end{itemize}

The remainder of this paper is structured as follows. 
Section~\ref{Sec:Background} provides the necessary background information.
Section~\ref{Sec:Problem.Definition} defines our research problem and outlines the underlying assumptions. 
Section~\ref{Sec:Approach} describes our approach. 
Section~\ref{Sec:Empirical.Evaluation} presents our empirical evaluation and the corresponding results. 
Section~\ref{Sec:Discussion} discusses our empirical results. 
Section~\ref{Sec:Threats} investigates threats to the validity of our results. 
Section~\ref{Sec:RW} describes related works.
Finally, Section~\ref{Sec:Conclusion} concludes our work.

\section{Background}
\label{Sec:Background}
Reinforcement Learning (RL) is a method that trains a model or an agent to make a sequence of decisions to reach a final goal, making it an increasingly popular approach for complex autonomous systems. RL involves the use of trials and errors to explore the environment by choosing from a range of potential actions, with the goal of maximizing the resulting reward.


\subsection{Definitions}





\begin{definition}\textit{({RL Agent Behavior.}})
\label{def-MDP} The behavior of an RL agent can be expressed as a Markov Decision Process (MDP), denoted by $\langle S, A, T, R, \gamma \rangle$. In this formulation, $S$ and $A$ represent the set of possible states and actions, respectively. The function $T: S \times A \times S \xrightarrow[]{} [0,1]$ defines the transition probabilities between states. The reward function $R: S \times A \xrightarrow[]{} [0, R_{max}]$ calculates the immediate reward for each action taken in a given state. Finally, the discount factor $\gamma \in [0,1]$ accounts for the difference in short-term and long-term rewards~\cite{sutton2018reinforcement}.
\end{definition}

\begin{definition}\textit{({Episodes.}})
\label{def-episode} An episode $e$ refers to a finite sequence of state-action pairs, denoted as $[(s_j,a_j) | s_j \in S, a_j \in A, 0 \le j \le n, n \in \mathbb{N}]$. The initial state is represented by the state in the first pair, while the end state is represented by the state in the final pair. An end state is defined as a state where the agent cannot take any further actions.
\end{definition}

\begin{definition}\textit{({Unsafe State.}})
\label{def-faulty-state}
An unsafe state is a state in which at least one of the defined safety requirements (e.g., the autonomous vehicle must detect and avoid obstacles) is violated. An unsafe state is often an end state and may lead to damage or harm depending on the environment. In the context of autonomous vehicles, for example, an unsafe state is a state where a collision occurs.
\end{definition}

\begin{definition}\textit{({Unsafe Episode.}})
\label{def-faulty-episode} If an episode $e$ contains an unsafe state, it is considered an unsafe episode. An unsafe episode may lead to an unsafe situation in the context of safety-critical systems (e.g., hitting an obstacle).
    
\end{definition}

\subsection{State Abstraction}
\label{subsec:State abstraction}

State abstraction is a technique used to reduce the size of the state space through clustering to decrease the complexity of a problem~\cite{akrour2018regularizing,pmlr-v80-abel18a,zolfagharian2023search}. This is achieved by mapping an original state $s \in S$ to an abstract state $s^\phi \in S^\phi$ using an abstraction function $\phi : S \xrightarrow{} S^\phi$, where $S^\phi$ represents the set of abstract states. The abstract state space is often much smaller than the original state space. 

In general, there are three state-of-the-art state abstraction methods in the context of RL~\cite{jiang2018notes,li2006towards}.
These methods are $\pi^*$-irrelevance abstraction, model-irrelevance abstraction, and $Q^*$-irrelevance abstraction~\cite{jiang2018notes,zolfagharian2023search}. This work specifically concentrates on using $Q^*$-irrelevance state abstraction to reduce the state space of DRL agents, since it aligns well with our objective (e.g., Q-values capture the perception of the agent) and has been shown to work in previous studies~\cite{zolfagharian2023search,hao2021learning,liu2022soft}.  

\minor{Furthermore, $\pi^*$-irrelevance abstraction groups states based on the optimal policy, which often leads to overly coarse state abstraction, as it considers only the best action in a state. This lack of precision is particularly problematic in safety-critical environments. Model-irrelevance, on the other hand, requires access to the agent's policy and environment's transition model, which is impractical in black-box scenarios and introduces significant computational overhead, preventing real-time monitoring. In contrast, $Q^*$-irrelevance allows for flexible control over the level of abstraction, making it possible to adjust the granularity of state abstraction while efficiently leveraging Q-values to predict safety violations. This makes $Q^*$-irrelevance a more suitable choice for our approach, balancing precision and computational feasibility.}

The abstraction $\phi_d$ denotes a $Q^*$-irrelevance-based abstraction of a state and its formal definition is as follows: 
\begin{equation}
\small
   \phi_d(s_1)=\phi_d(s_2) \equiv \forall a \in A : \bigg\lceil \frac{Q^*(s_1,a)}{d} \bigg\rceil = \bigg\lceil \frac{Q^*(s_2,a)}{d} \bigg\rceil
\end{equation}

where $Q^*(s,a)$ is the optimal state-action function that returns the maximum expected reward from state $s$ up to the final state when selecting action $a$ in state $s$. Q-values $Q^*(s,a)$ are used to guide the RL agent's decision-making process. The control parameter $d$ determines the level of abstraction and the ceiling of $\frac{Q^{*}(s,a)}{d}$ is then calculated. It also allows for the grouping of more states together as $d$ increases. Such grouping results in a significant reduction in the size of the state space. Intuitively, this method discretizes the $Q^*$-values by using buckets of size $d$.

Let us illustrate this abstraction method with a simple example and show why it is adequate in our context. Assume, for example, two distinct concrete states where a trained RL agent has learned the same Q-values. Given that our objective is to monitor the agent, it is reasonable to assume that these states are similar since the agent has learned to assign the same Q-values to both states. This implies that the agent perceives both states to be identical~\cite{zolfagharian2023search}. Moreover, this abstraction method allows us to adjust the abstraction level, which gives us control over the granularity of the abstraction in different environments. Thus, we can adjust the abstraction level to achieve a higher accuracy in predicting safety violations for a specific RL agent. 

\section{Problem Definition}
\label{Sec:Problem.Definition}

We present in this section a comprehensive overview of our research problem and the assumptions that underlie our work.   

\subsection{Problem} 

RL has become an essential part of intelligent systems that can learn and adapt to complex and dynamic environments. It has been increasingly applied in various safety-critical areas such as autonomous driving, healthcare, and industrial control systems. 
Therefore, the development of a reliable runtime safety monitoring approach for RL agents is critical for ensuring the safety of the agent and other entities in the environment. 
A safety monitor aims to predict potential safety violations and trigger safety mechanisms to take corrective or preventive actions in real time. However, designing an algorithm that can learn to predict safety violations accurately and as early as possible presents a significant challenge. In this work, we propose to build a safety violation prediction approach based on \textit{Random Forest} algorithm and features based on abstract states to effectively predict unsafe episodes. Our safety monitoring approach will then use such model to monitor states and actions over time and predict the probability of safety violations before they occur. 


\subsection{Assumptions}

\added{In this work, we focus on RL agents with discrete actions and a deterministic policy. Indeed, our state abstraction method is designed for discrete action spaces. We acknowledge that this is a limitation of our approach. However, several studies have explored discretization approaches to effectively transform continuous action spaces into discrete ones~\cite{tang2020discretizing,tavakoli2018action}.  Addressing continuous action spaces is nonetheless beyond the scope of this paper. Moreover, we assume that a deterministic policy is realistic since, in many application domains (specifically in safety-critical domains), uncertainty is not acceptable and random actions should be avoided~\cite{zolfagharian2023search}. 
Further, we build our work on model-free Q-learning RL algorithms since our abstraction method relies on Q-values. Note that such algorithms are commonly used and have been extensively researched~\cite{swazinna2022comparing,OpenAi,jang2019q,van2016deep,wang2016dueling,Fortunato2017NoisyNF,hessel2017rainbow}, hence our choice to target them. Beyond that, Q-learning algorithms are versatile and applicable to a wide range of problems with discrete action spaces. Also, they are relatively sample efficient, and due to the off-policy setting, Q-learning algorithms can learn from experiences generated by any policy, making them valuable for safety-critical problems where exploration is costly or dangerous~\cite{fujimoto2019off,10.1007/978-3-031-33469-6_3,10.1007/978-3-030-46133-1_2}. Note that Q-values leveraged by Q-learning algorithms directly represent the expected cumulative reward for taking an action in a given state, providing insight into the agent's decision-making process. This interpretability can be valuable in domains where understanding the agent's behavior is crucial, such as safety-critical domains.}

 
\section{Approach}
\label{Sec:Approach}

We present in this section our safety monitoring approach for DRL agents, along with its various components. 

 \begin{figure*}[ht]
    \centering
    \includegraphics[width=\textwidth]{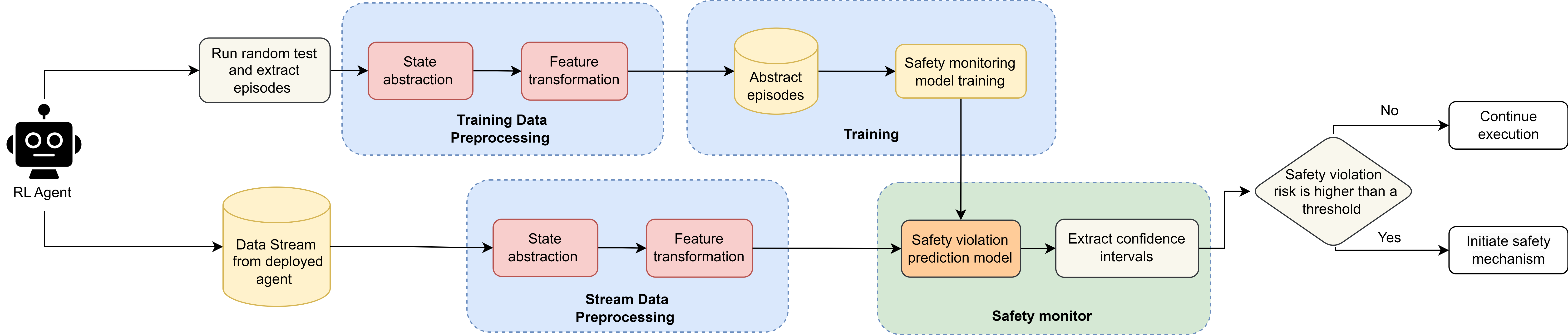}
    \caption{Overview of SMARLA}
    \label{fig:Approach}
  \end{figure*}

\subsection{Overview of the Approach}
\label{Sec:Overview}

As depicted in Figure~\ref{fig:Approach}, we propose a safety monitoring system to predict safety violations of DRL agents as early as possible. The early detection of unsafe episodes is indeed crucial for any safety-critical system to enable prompt corrective actions to be taken and thus prevent unsafe behavior of the agent. \added{Our approach predicts safety violations by observing the behavior of the RL agent rather than directly monitoring the state of the environment. Observing and processing such state information can be computationally expensive and inefficient, especially in environments with large or continuous state spaces. In contrast, our method is agnostic to the agent's input and leverages the agent's Q-values as features throughout the episodes. Q-values summarize the expected future rewards of all possible actions in given states, providing a more abstract and computationally efficient representation of the agent's behavior. By focusing on Q-values instead of raw state data,  \textit{SMARLA} can be seamlessly applied to agents with various types of input. Moreover, relying on Q-values significantly reduces the computational complexity involved in monitoring. This allows us to employ a lightweight machine learning model, such as \textit{Random Forest}, which is shown to be effective for this task (as reported in Section~\ref{Sec:RQ1}). While it might be possible to predict unsafe episodes using state information, the drawback is that it would require handling a higher-dimension and more complex dataset, potentially necessitating the development of resource-intensive models.}

To train the ML model we randomly execute the RL agent and extract the generated episodes. These episodes are labeled as either safe or unsafe. Since the state space is very large, we rely on state abstraction to reduce the state space and enhance the learnability of our ML model. Once we have trained the model on these labeled episodes, we use it to monitor the behavior of the agent and estimate the probability of encountering an unsafe state while an episode is being executed. We rely on the confidence intervals of such probability to accurately determine the optimal time step to trigger safety mechanisms.

\subsection{Training of the safety violation prediction model}

\added{We need a lightweight  ML model that can accurately classify RL episodes as safe or unsafe, and be effectively deployed on resource-constrained edge devices~\cite{buerkle2021fault}.} \minor{This is especially relevant in real-world scenarios where safety monitoring is essential, such as for RL agents operating on drones or small autonomous robotic systems~\cite{azar2021drone,feng2023dense}. Furthermore, even in resource-rich environments (e.g., some autonomous driving systems), having computational resources for core functionalities does not guarantee the availability of substantial unused capacity for implementing additional safety monitoring. Safety monitoring models should therefore be efficient and non-intrusive, ensuring they do not overburden the system's existing resources. } 
\added{Thus, we exclude DNN models and choose \textit{Random Forest} as a machine learning model because of (1) its ability to handle a large number of features, (2) its proven robustness to overfitting, and (3) its efficiency in providing classification results at inference time~\cite{10.1023/A:1010933404324,friedman2001greedy,zolfagharian2023search}.}

\added{Moreover, the empirical study presented by Caruana \textit{et al.}~\cite{caruana2006empirical} supports the use of \textit{Random Forest} due to its consistently high performance across multiple evaluation metrics, robustness to overfitting, and effective probability calibration. The authors report that \textit{Random Forest} maintains balanced performance in both traditional metrics, such as accuracy and F-score, and advanced metrics such as ROC Area and cross-entropy. Additionally, \textit{Random Forest} requires less parameter tuning compared to DNN models, making them practical for real-world applications.} 

\added{We also conducted comparison studies with other ML models, such as \textit{K-Nearest Neighbor} and \textit{Decision Trees}. However, \textit{Random Forest} provided the most accurate classification results. Since this aspect is not the primary focus of our work, we omitted the results of these experiments in this paper and reported the results in our replication package.}

\subsubsection{Training Data Collection}
\label{subsec:Prepare training data ML}

To build our safety violation prediction model, we collect the training data by randomly executing the RL agent starting from different initial states. Episodes are labeled unsafe if safety violation is observed within episodes; otherwise, they are labeled as safe. The goal of this process is to generate diverse episodes and form a training dataset that encompasses both safe and unsafe episodes. 
For each execution of the agent, we extract the corresponding episode (i.e., pairs of states and actions) along with the different $Q^*$-values in each state. Such data will be used to build the abstract states as we describe in the following.

\subsubsection{State Abstraction and Feature Transformation }\label{subsec: State Abstraction for Training}

Once we collect the training episodes, we need to build the set of abstract states and map each concrete state to its corresponding abstract state. This is meant to reduce the state space and thus enhance the learnability of the ML model. To build abstract states, we use the $Q^*$-irrelevance abstraction method, which was explained in Section~\ref{subsec:State abstraction}. The abstraction algorithm takes the concrete states as input and an abstraction level $d$ as a control parameter. The algorithm attempts to find the abstract state $s^{\phi_d} \in S_{^\phi}$ corresponding to each concrete state by evaluating the $Q^*$-values of all available actions, as explained in Section~\ref{subsec:State abstraction}. For each concrete state, if the algorithm detects a match with an abstract state from a previously processed concrete state, it assigns the same abstract state to the current concrete state. If no matching abstract state is found, a new abstract state is generated. 

After building the set of abstract states, we replace each concrete state in the training episodes with its corresponding abstract state to enable effective learning of the ML model. 
To train the model, we translate each episode into a feature vector that determines whether abstract states are present or not in episodes. Specifically, we represent each episode using a binary feature vector that encodes the presence (1) or absence (0) of each abstract state $S^\phi_{j}$ within the episode, where $1 \leq j \leq n$, and $n$ represents the total number of abstract states. An example of a feature vector is provided as follows: 

\[ 
\begin{matrix}
& S^\phi_1  & S^\phi_2  & \cdots  &  S^\phi_j & \cdots & S^\phi_{n-1} & S^\phi_n\\
episode_i & 1 & 0& \cdots &1&\cdots & 1 &0
\end{matrix}
\]

\minor{The main advantage of this representation is that it significantly reduces the feature space by relying on the presence or absence of the abstract states.} Specifically, considering $n$ different abstract states, the feature space of this representation is only $2^n$. \minor{Alternatively, if needed, we can rely on features based on the frequency of abstract states, which enhance the representation by accounting for how often each abstract state is observed during the execution of an episode, at the cost of course of a larger feature space.} It is worth noting that we only consider abstract classes that have been observed in the training dataset of the ML model, which we anticipate will be relatively comprehensive. 
However, the feature representation does not account for the order of abstract states within episodes, which may be a weakness if we are unable to accurately predict safety violations as a result. Empirical results will tell whether these two potential issues are significant in practice.

\subsection{Safety Violation Prediction Model in Operation}
\label{subsec: Hazard detection in action}

Runtime monitors are designed to detect and prevent safety violations while the system is in operation. Once the training is complete, we deploy the safety violation prediction model alongside the RL agent. The safety violation prediction model monitors episodes and estimates the probability of a safety violation at each state. We should note that the streamed episodes have different lengths. We then apply the stream data pre-processing step that generates binary vectors of length equal to the size of abstract states (i.e., equal to $n$). This process is necessary to be able to run the safety monitor at runtime. Stream data pre-processing is accomplished by using state abstraction and feature representation techniques, which are explained in detail in Section~\ref{subsec: State Abstraction for Training}.

\subsubsection{Estimating The Probability of Safety Violation} 
\label{Sec:FunctionalFault}

At each time step, the data stream is received and episodes are transformed into binary vectors based on state abstraction and feature representation. Each binary vector indicates the presence (1) or absence (0) of abstract states observed so far in the corresponding episode. The safety violation prediction model uses these binary inputs to estimate the probability of safety violation based on the abstract states visited so far.

Let us suppose that the agent is at time step $t$, and the running episode is as follows:

\begin{equation}
episode_{i}(t)= [(s_1,a_1),...,(s_{t-1},a_{t-1}),(s_{t},a_{t})]
\end{equation}

The stream data pre-processing process converts these episodes into binary feature vectors showing the presence and absence of the abstract states visited so far (until time step $t$). The transformed episode is as follows:

\[ 
\begin{matrix}
& S^\phi_1  & S^\phi_2  & \cdots  &  S^\phi_j & \cdots & S^\phi_{n-1} & S^\phi_n\\
episode_{i}(t) & 0 & 0& \cdots &1&\cdots&  1 &0
\end{matrix}
\]

At each time step $t$, the episode is fed into the safety violation prediction model. Then it estimates the probability of safety violation $P_{e_i}(t)$ at time step $t$. In other words, the safety violation prediction model captures the relationship between the presence of abstract states in an episode and the occurrence of safety violations. 

\added{We should note that state abstraction allows us not to depend on the specific occurrence of concrete states in RL episodes when predicting safety violations in operation. State abstraction thus generalizes patterns to concrete states that have possibly not been encountered in operation by analyzing change patterns in the agent’s Q-values, which are indicative of possible unsafe states. In other words, in our predictive model, we operate under the hypothesis that patterns of Q-value changes observed in training episodes that contain safety violations are likely to reappear in operation. This assumption allows \textit{SMARLA} to extend its predictive capability beyond the concrete episodes seen during training, by recognizing and responding to analogous change patterns, even in the presence of unseen concrete states. Nonetheless, during the prediction process, we may encounter concrete states belonging to unseen abstract states—those not present in the training data. Such situations require careful handling to ensure safety. Two strategies can be applied:
\begin{itemize}
    \item Stop the Execution: In the presence of an unforeseen situation, stopping the execution can prevent potential risks associated with unseen states. This conservative approach prioritizes safety by halting operations when the system encounters unfamiliar states that could lead to unknown hazards.
    \item Ignore the Unseen Abstract State: Alternatively, the system can ignore the unseen state and continue predicting safety violations based on other observed states in the episode. This approach assumes that the unseen state does not significantly impact the safety and leverages the information from other known states to make predictions. 
\end{itemize}}

\added{In our experiments, we observed that episodes had only a few unseen abstract states and could still be correctly predicted using only the seen abstract states. Therefore, we chose the latter approach.}

\added{We should note that the \textit{Random Forest} model used in this study not only generates class predictions but also calculates probabilities for these classifications, using an ensemble of decision trees (also called estimators). Each tree in the forest is constructed from a bootstrap sample of the dataset, and at each decision node, a randomly selected subset of features is considered. The trees individually predict whether a given episode is \textit{"safe"} or \textit{"unsafe"}. The overall probability for each class is derived by averaging the predictions across all trees in the forest.
By convention, a probability threshold of 50\% is applied, whereby probabilities of 50\% or higher result in an \textit{"unsafe"} classification, while probabilities below 50\% lead to a \textit{"safe"} classification. These output probabilities from a Random Forest are crucial as they reflect the confidence level of the model’s predictions, providing valuable guidance for decision-making, especially in scenarios where precise and timely detection of safety violations is paramount.}

\subsubsection{Safety Violation Prediction }
\label{Subsec: Safety Violation}

Based on the estimated probabilities of safety violations we compute the predictions' confidence intervals~\cite{easton1997statistics} to determine the time step at which a safety violation is likely to occur with high confidence. At this time step, corrective or preventive measures can be initiated to avoid damage and harm. 
To determine the confidence intervals of probabilities of safety violations, we rely on the predictions of the individual estimators in the \textit{Random Forest} model. At each time step for a given episode, each estimator predicts the probability of safety violation. Based on the mean prediction and standard deviation, we calculate the confidence intervals at each time step $t$ as follows:

\begin{equation}
    CI(t) = \Bar{X(t)} \pm Z \times \frac{\sigma}{\sqrt{m}} 
\end{equation}

where $\Bar{X(t)}$ represents the mean predicted probability at time step $t$, $Z$ is the critical value of the normal distribution with $Z=1.96$ for a 95\% confidence interval, $\sigma$ is the standard deviation, and m is the number of decision tree estimators~\cite{easton1997statistics}. The lower and upper bounds of the confidence intervals are as follows:
\begin{equation}
[Low(t), Up(t)] =  [\Bar{X(t)} - \frac{\sigma}{\sqrt{m}}, \Bar{X(t)} + \frac{\sigma}{\sqrt{m}}]
\end{equation}

\added{More specifically, a 95\% confidence interval indicates that, if such intervals were constructed from numerous samples (specifically, the output probabilities of each estimator in a \textit{Random Forest}), there is a 95\% likelihood that the probability of a safety violation predicted by the \textit{Random Forest} model will fall between the lower and upper bounds of this interval. The interval is centered on the predicted probability of a safety violation and extends symmetrically in both directions. The remaining 5\%  probability, not covered by the interval, is equally divided between the distribution's two tails, allocating 2.5\% to the lower tail and 2.5\%  to the upper tail. Consequently, at any time step t, there is a 2.5\% chance that the probability of a safety violation will be either lower than $Low(t)$ or higher than $Up(t)$.}

\added{\textit{SMARLA} supports three different decision criteria, based on the confidence intervals of the estimated violation probability, to determine the appropriate time step to classify the episode as unsafe and trigger safety mechanisms. We should note that all decision criteria rely on a threshold that may be adjusted according to the agent and the specific context. }  

\begin{itemize}
    
\item \added{\textbf{Upper Bound of the Confidence Interval:} This first decision criterion involves using the upper bound of the confidence interval to classify an episode as unsafe. If the upper bound of the confidence interval is equal to or greater than a certain prediction threshold, we classify the episode as unsafe. We consider this approach conservative because it allows for earlier predictions of safety violations compared to using the output probability or the lower bound of the confidence interval. Although this method provides an earlier prediction of safety violations, it might lead to a higher number of false positives. This is suitable for scenarios where early detection is critical, even at the cost of some false positives.}

\item \added{\textbf{Output Probability:} The second decision criterion is based on the actual output probability of safety violation being equal to or greater than a certain prediction threshold. This criterion provides a trade-off between timeliness and accuracy for predicting safety violations. It typically results in fewer false positives than using the upper bound but may delay the detection of unsafe episodes in comparison. This method is appropriate for environments where accuracy is more critical than early detection, minimizing unnecessary interventions.}

\item \added{\textbf{Lower Bound of the Confidence Interval:} The third decision criterion uses the lower bound of the confidence interval. If the lower bound is greater than a certain prediction threshold, the episode is predicted as unsafe. This decision criterion minimizes false positives but at the potential cost of detecting unsafe episodes later than the other two criteria. This approach is preferred in situations where it is critical to avoid false positives and where the consequences of late detection are not severe.}

\end{itemize}

\added{In our case studies, we rely on the upper bound because, to ensure safety, we take a conservative approach and rather err on the side of caution. In Section~\ref{Sec:Rq2Results}, we detail the decision criteria and report empirical results that confirm this is the best choice in our case studies. }

\section{Empirical Evaluation}
\label{Sec:Empirical.Evaluation}

This section describes the empirical evaluation of our monitoring approach, including our research questions, the used case studies, our experiments, and the obtained results. 

\subsection{Research Questions}
Our empirical evaluation is designed to answer the following research questions.

\noindent{\textit{\textbf{RQ1. How accurately and early can we predict safety violations during the execution of episodes?}}} 
This research question aims to investigate how accurately and early our approach can predict safety violations of the RL agent during its execution. Preferably, high accuracy should be reached as early as possible before the occurrence of safety violations to enable the effective activation of safety mechanisms. 

\noindent{\textit{\textbf{RQ2. How can the safety monitor determine when to trust the prediction of safety violations?}}}
This research question investigates the application of confidence intervals, based on \textit{Random Forest} models, to enable the safety monitoring model identify the earliest time step at which the prediction of violations can be trusted. 

\noindent\added{\textbf{RQ3. What is the effect of the prediction threshold on the safety monitoring system?}}
\added{This research question explores the impact of varying the prediction threshold values on our safety monitoring approach. We aim to understand how different prediction thresholds affect both the prediction time and the accuracy of our safety monitoring system during operation. Our goal is to understand the trade-offs between making early predictions of safety violations and maintaining high prediction accuracy. In other words, we want to assess how setting lower or higher thresholds influences SMARLA's ability to promptly detect safety violations while minimizing false positives and false negatives.}

\noindent{\textit{\textbf{RQ4. What is the effect of the abstraction level on the safety monitoring system?}}} We aim in this research question to investigate if and how different levels of state abstraction affect the safety violation prediction capabilities of the model. Specifically, we want to study the impact of state abstraction levels on (1) the accuracy of the safety violation prediction model after training, and (2) the accuracy of the ML model in operation.
Our goal is to gain insights into the possible trade-offs between the size of the feature space and the granularity of information captured by features, both determined by the abstraction level. Such an analysis aims to provide guidance in selecting proper abstraction levels in practice.

\subsection{Case Studies}\label{subsec:Case}
To validate our safety monitoring approach, we considered \added{three} well-known case studies widely used as benchmarks in RL-related research~\cite{Behzadan2019AdversarialEO,Behzadan2019SequentialTF,DBLP:journals/corr/abs-2006-05032,pattanaik2018robust,gajcin2023iterativerewardshapingusing}: the \textit{Cart-Pole} balancing problem~\cite{doya2000reinforcement}, the \textit{Mountain-Car} problem~\cite{moore1990efficient}, \added{and the \textit{Highway Driving} Problem}~\cite{highway-env} which we describe in detail in the following sections. 
\begin{figure}[ht]
        \centering
        \includegraphics[width=.5\linewidth]{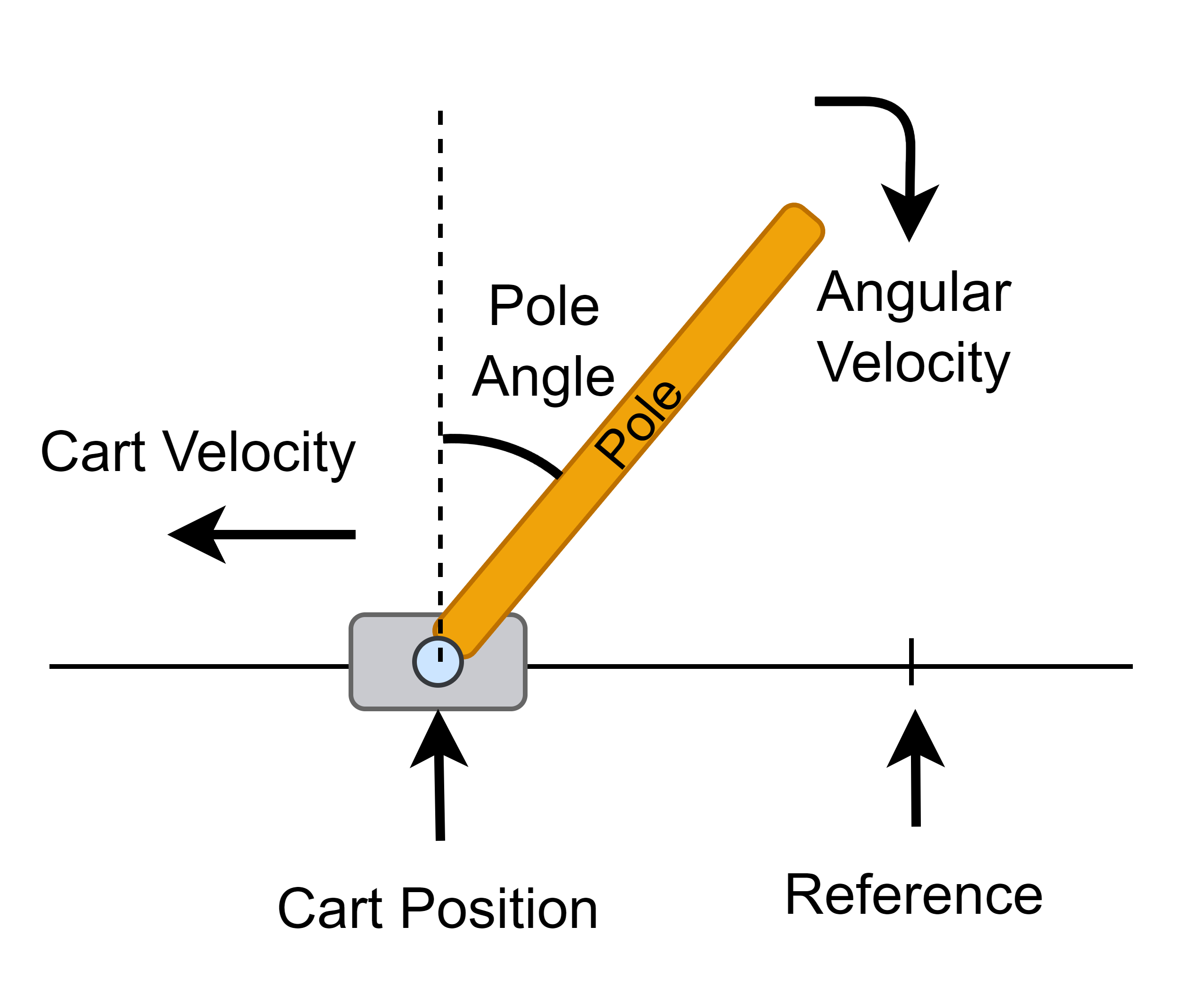}
    \caption{\textit{Cart-Pole} case study}
    \label{fig:CartpoleExample}
    \end{figure}
\subsubsection{Cart-Pole Balancing Problem}\label{subsec:Cartpole}

The \textit{Cart-Pole} balancing problem involves a pole that is attached to a cart moving on a track. The cart can move horizontally in both directions within a specific range. The objective is to keep the pole upright by moving the cart and adjusting its velocity. As shown in Figure~\ref{fig:CartpoleExample}, we characterize the state of the agent by four elements: (1) the position of the cart, (2) the velocity of the cart, (3) the angle of the pole, and (4) the angular velocity of the pole. There are two discrete actions that can be used to control the cart: move to the left, and move to the right.

A reward of +1 is granted for each time step when the pole is still upright. An episode terminates if (1) the cart is away from the center with a distance of more than 2.4 units, or (2) the angle of the pole is larger than 12 degrees, or (3) the pole remains upright for 200 time steps. An episode is considered unsafe if the cart moves away from the center with a distance above 2.4 units, regardless of the accumulated reward. In such a situation, the cart can pass the border and cause damage, which is therefore considered a safety violation.

\subsubsection{Mountain-Car Problem}\label{subsec:MTC}

\begin{figure}[ht]
        \centering
        \includegraphics[width=.5\linewidth]{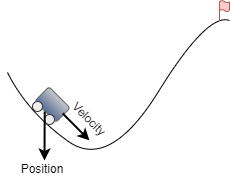}
    \caption{\textit{Mountain-Car} case study}
    \label{fig:MountainCarExample}
    \end{figure}

In the \textit{Mountain-Car} problem, an under-powered car is placed in a valley between two hills and tries to reach a goal state on the top of the right hill. Since the gravity is stronger than the engine of the car, the car cannot climb up the steep slope even with full throttle. The objective is to control the car in such a way that it can accumulate enough momentum to eventually reach the goal state on top of the right hill as soon as possible. We characterize the state of the agent with two elements: (1) the location of the car along the x-axis, and (2) the velocity of the car as illustrated in Figure~\ref{fig:MountainCarExample}. The agent controls the car with three actions: (1) accelerate to the left, (2) accelerate to the right, and (3) do not accelerate.

A penalty of -1 is applied for each time step until reaching the goal. 
An episode terminates if the car (1) reaches the goal state, (2) crosses the left border (in this case the reward is -200), or (3) exceeds the limit of 200 time steps. 
A safety violation is simulated by considering the crossing of the left border of the environment as an irrecoverable unsafe state that poses potential damage to the car. Consequently, when the car crosses the left border of the environment, it triggers a safety violation, leading to the termination of the episode. This modification allows us to assess the effectiveness of \textit{SMARLA} in predicting safety violations.

\subsubsection{The Highway Driving Problem} \label{Sec:CaseHighway}

\begin{figure}[ht]
        \centering
        \includegraphics[width=\linewidth]{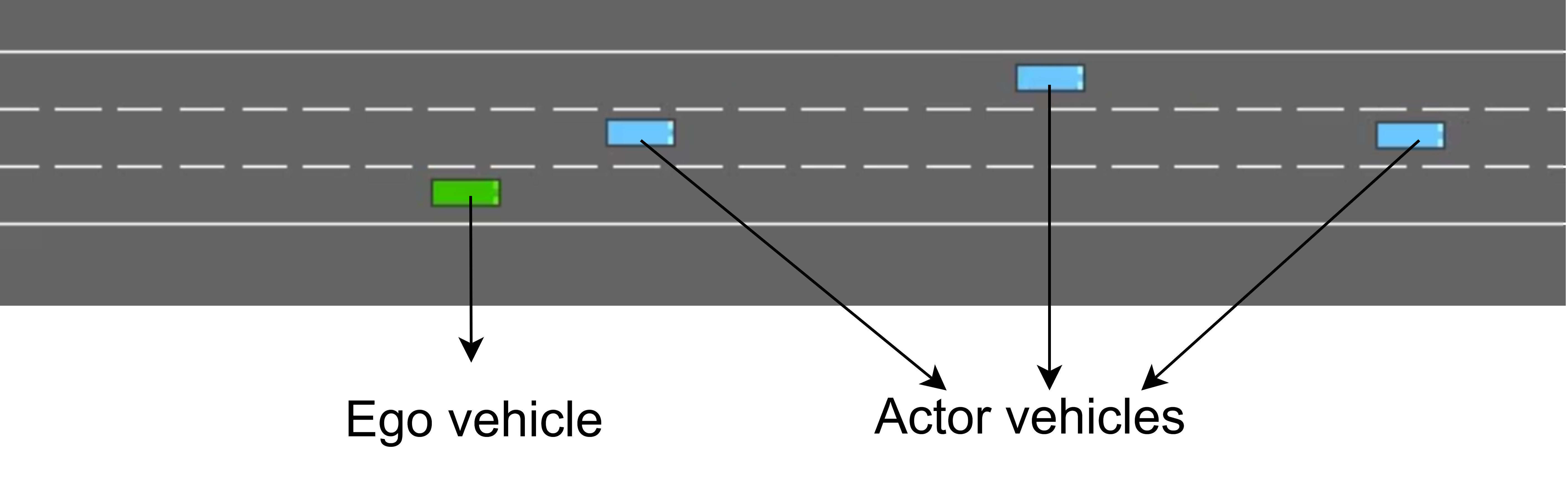}
    \caption{\added{\textit{Highway Driving} case study}}
    \label{fig:HighwayDrivingExample}
    \end{figure}

\added{In the \textit{Highway Driving} problem, an RL agent is responsible for driving a car on a highway with three lanes, as shown in Figure~\ref{fig:HighwayDrivingExample}. In this case study, we have a partially observable environment where the state information is only observable within a certain range. 
The objective of the agent is to drive along the highway at a high speed without colliding with other vehicles, for as long as possible.  The environment is characterized by a continuous state space that includes kinematic details of the ego vehicle and other vehicles on the road, amounting to a total of 15 parameters. The discrete action space available to the vehicle consists of five high-level actions: (1) moving right, (2) moving left, (3) accelerating, (4) remaining idle, and (5) braking. The agent receives a penalty of -10 for collisions, a 0.1 reward for maintaining its position in the rightmost lane, and a 0.4 reward for driving at high speeds (i.e., 20 to 30), which diminishes linearly for speeds outside the 20-30 range. Finally, the reward is normalized to the range of [0,1]. An episode ends under two conditions: (1) if the car collides with another vehicle, or (2) if it exceeds the maximum duration of episodes (30 time steps in our case) without any collision. An episode is deemed unsafe if the ego vehicle collides with another vehicle on the highway, constituting a failure to maintain a safe driving session. }

\subsection{Implementation}
\label{Sec:Implementation}

We implemented the RL agents for publicly available RL case studies using stable baselines~\cite{stable-baselines}. To develop these agents, we used a DQN policy network~\cite{mnih2013playing} with the standard settings provided by stable baselines. Specifically, we employed Double Q-learning~\cite{vanhasselt2015deep} and dueling DQN~\cite{wang2016dueling}.

We trained the \textit{Cart-Pole} RL agent for 70,000 time steps. The trained agent achieved an average reward of 193. On average, the pole remains upright for 193 time steps out of a maximum of 200.
The \textit{Mountain Car} agent was trained for 90,000 time steps. Considering the penalty of -1 applied at each time step in this specific case study, the agent achieves an average reward of -126, with an average episode duration of 112.
\added{We trained the \textit{Highway Driving} RL agent for 70,000 time steps. On average, the car drives safely without any collision for 29 time steps out of a maximum of 30.}

To train our safety monitor, we sampled episodes through random execution of the RL agent for each case study. Safety monitor's training data for \textit{Cart-Pole} includes 2200 episodes from which 215 are unsafe, for \textit{Mountain-Car} 2200 episodes with 279 unsafe episodes, \added{ and for \textit{Highway} case study 4000 episodes with 222 unsafe episodes}. \added{We should note that since our safety monitoring approach is black-box, it is designed to predict safety violations without depending on the training level of the DRL agents. This approach allows us to focus exclusively on the detection and prediction of safety violations, which is crucial for applications where the primary concern is maintaining operational safety rather than achieving training optimality.}

Also, we empirically determined that suitable abstraction levels were 0.11 for \textit{Cart-Pole},  five for \textit{Mountain-Car} \added{ and 0.2 for the \textit{Highway Driving} problem. Based on these abstraction levels, the number of abstract states is 1659 for \textit{Cart-Pole} with abstraction level 0.11 (compared to 424,962 concrete states), and the number of abstract states is 13,171 for \textit{Mountain-Car} with abstraction level 5 (compared to 247,035 concrete states). Finally, in the \textit{Highway Driving} problem the number of abstract states with abstraction level 0.2 is 136 (compared to 115,730 concrete states). } Also, with these levels, \textit{SMARLA} achieved the highest F1-score and the earliest prediction of safety violations (explained in detail in Section~\ref{Subsec:RQ4AbsLevel}).

\minor{It is important to note that some safety monitoring methods were not considered for comparison due to fundamental differences in their assumptions and objectives. For example, many approaches are white-box ~\cite{henzinger2020outside,michelmore2020uncertainty,michelmore2018evaluating}, assuming access to the internals of the agent, which contradicts our focus on black-box monitoring scenarios. Additionally, methods that depend on DNN models or image inputs rely on image transformations~\cite{xiao2020likelihood}, whereas \textit{SMARLA} is designed to be agnostic to the agent's input type, thus widening its applicability. Several existing methods, such as forward filtering-based and model-checking-based monitoring approaches~\cite{junges2021runtime}, rely on detailed models of the environment, which are often not available or are too complex to model. \textit{SMARLA} is designed for model-free RL algorithms, making it more applicable to real-world scenarios where environment models are difficult to develop. These distinctions justify the exclusion of such methods from our evaluation, as they do not align with the goals and assumptions of our work.}

\subsection{Evaluation and Results}

\subsubsection{RQ1. How accurately and early can we predict safety violations during the execution of episodes?} \label{Sec:RQ1}

\begin{figure}[ht]

\includegraphics[width=\linewidth]{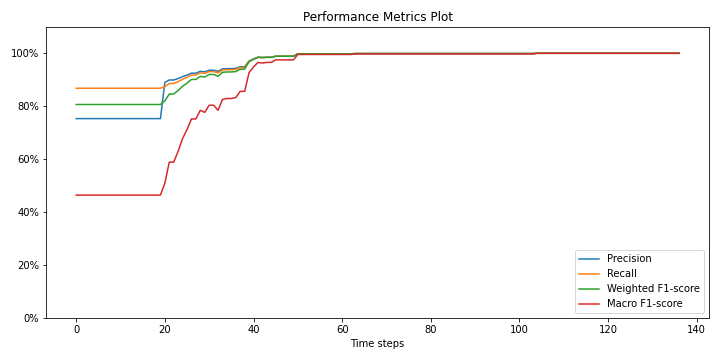}
    \caption{Performance of the safety violation prediction models in the \textit{Mountain-Car} case study }
    \label{Rq1_Mountain-Car_f1}
    \end{figure}

\begin{figure}[ht]

\includegraphics[width=\linewidth]{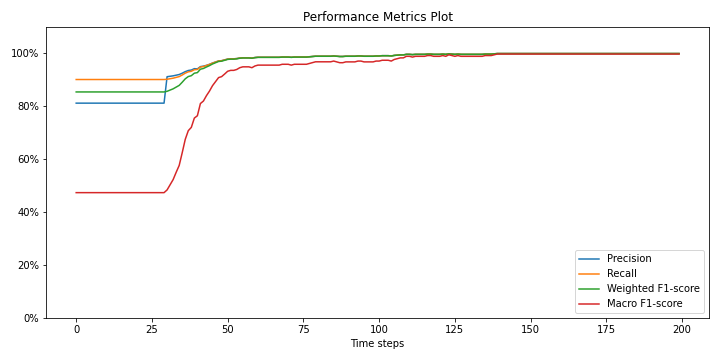}
    \caption{Performance of the safety violation prediction models in the \textit{Cart-Pole} case study }
    \label{Rq1_Cartpole_f1}
    \end{figure}

\begin{figure}[ht]

\includegraphics[width=\linewidth]{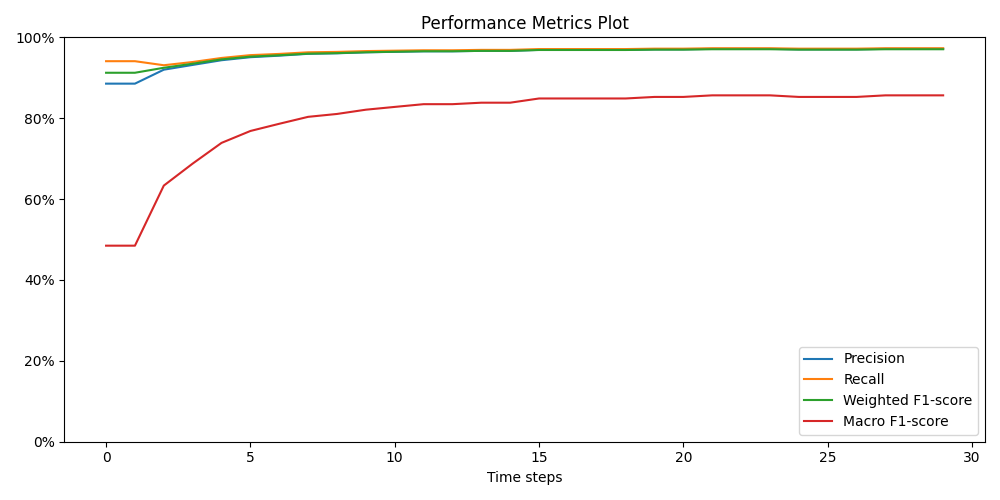}
    \caption{\added{Performance of the safety violation prediction models in the \textit{Highway} case study} }
    \label{Rq1_Highway_f1}
    \end{figure}
    
To answer this research question, we performed a series of $n$ random executions ($n=1000$) of the RL agent and extracted the corresponding episodes $e_{1 \leq i \leq n}$. To build our ground truth, these episodes were labeled as either safe or unsafe, taking into account the presence or absence of safety violations observed within each episode. We should note that in all case studies, there was a maximum of one safety violation per episode, occurring at the end of an unsafe episode as one of the termination criteria. We monitored the execution of each episode at each time step with \textit{SMARLA}. As described in Section~\ref{Subsec: Safety Violation}, when the upper bound of the confidence interval $Up(t)$ is greater than 50\% during the execution of the episode, \textit{SMARLA} classifies the episode as unsafe. For each case study, we computed the number of successfully predicted safety violations, measured the prediction precision, recall, and F1-scores at each time step, over the set of episodes, and depict the results in \added{Figures~\ref{Rq1_Mountain-Car_f1},~\ref{Rq1_Cartpole_f1} and~\ref{Rq1_Highway_f1}}. \added{Note that in this paper, we report the weighted precision, recall, and F1-scores to show the performance of \textit{SMARLA} in real-world situations, where operational data for predicting safety violations is typically imbalanced. Additionally, we report the macro F1-score, which calculates the F1-score for each class individually and then averages them. Unlike the weighted F1-score, which can be influenced by the majority class, the macro F1-score ensures that each class is treated equally. This provides a more balanced and accurate evaluation of our safety monitoring model's performance across all classes (i.e., safe and unsafe episodes), regardless of their prevalence. Note that in the following when we refer to the F1-score in our work, we are specifically referring to the F1-macro score, to cope with the class imbalance problem.} These figures show a consistent pattern where the precision, recall, and F1-score exhibit a general increase over time before eventually reaching a plateau.

\textit{SMARLA}'s accuracy is thus improving over time as episodes execute. \added{Ultimately, our safety monitor correctly predicted 99 (out of 99), 132 (out of 132), and 40 (out of 59) safety violations in the \textit{Cart-Pole}, \textit{Mountain-Car} and \textit{Highway Driving} case studies, respectively. We obtained highly accurate safety violation prediction results for \textit{Cart-Pole} and \textit{Mountain-Car} case studies after roughly 50 times steps, while in the \textit{Highway Driving} case study we obtained highly accurate predictions after 10 time steps.} In the \textit{Cart-Pole} case study, we achieved precision, recall, and a weighted F1-score of 98.4\%, as well as an F1-score of 95\%, after time step 59.  Furthermore, in the \textit{Mountain-Car} case study, we obtained a precision of 99.8\%, a recall of 99.8\%, a weighted F1-score of 99.8\% and an F1-score of 99.5\% after 50 time steps. \added{Finally, in the \textit{Highway Driving} case study we achieved a precision of 96.3\%, a recall of 96.6\%, a weighted F1-score of 96.4\% and an F1-score of 82.1\%, after time step 10. } In all case studies, results highlight \textit{SMARLA}'s capability to provide accurate predictions of safety violations, relatively early, roughly halfway through the episode's execution as described next. Reasons for the lower F1-score for \textit{Highway Driving} are discussed below and are not related to \textit{SMARLA}.

In detail, the precision, recall, and F1-score plateau at 98\% from time step 106 in \textit{Cart-Pole} where the average length of the episodes is 193 and the minimum length is 118. At time step 106, on average, 45\% of the steps within the episodes remain to execute until reaching an unsafe state. This indicates that there is significant time to apply safety mechanisms.

Similarly, in the \textit{Mountain-Car} case study, we observed early accurate predictions of safety violations with \textit{SMARLA}. This is indicated by the precision, recall, and F1-score plateauing from time step 50. The average and minimum episode lengths in this case study are 112 and 95. At this time step, an average of 55\% of the steps within the episodes remain to execute. These results further validate the safety violation prediction model's ability to anticipate safety violations long before they occur.

\added{Finally, in our \textit{Highway Driving} case study, by time step 15, the F1-score of \textit{SMARLA} reached 85\% while the weighted F1-score, precision and recall plateaus at 97\%. Also, at this time step, on average 48\% of the episodes remain to execute until reaching an unsafe state.} \minor{We observe that we achieved a relatively lower F1-score in \textit{Highway Driving} case study compared to the \textit{Cart-Pole} and \textit{Mountain-Car} case studies. Since this might be due to the lack of information to predict safety violations, we explored a more informative feature representation, that is features based on frequency of abstract states. 
}

\begin{figure}[ht]

\includegraphics[width=\linewidth]{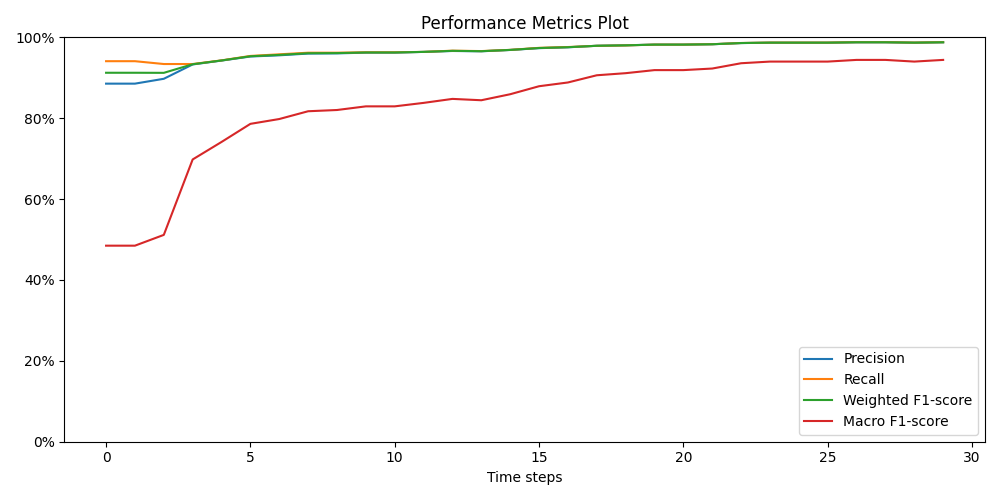}
    \caption{\minor{Performance of the safety violation prediction models in the \textit{Highway} case study with frequency-based features} }
    \label{Rq1_Highway_freq_f1}
    \end{figure}

\minor{Features based on the frequency of abstract states capture the number of times each abstract state is visited within an episode, providing a richer representation of the agent's behavior. An example of a transformed episode is as follows, where the episode is characterized by the number of occurrences of each abstract state:
\[ 
\begin{matrix}
& S^\phi_1  & S^\phi_2  & \cdots  &  S^\phi_j & \cdots & S^\phi_{n-1} & S^\phi_n\\
episode_{i}(t) & 0 & 1& \cdots &3&\cdots&  2 &0
\end{matrix}
\]
Using this frequency-based feature representation, we evaluated \textit{SMARLA}'s performance on the same test dataset of 1000 episodes for the \textit{Highway Driving} case study. The results are depicted in Figure~\ref{Rq1_Highway_freq_f1}}

\minor{Relying on frequency-based features in the \textit{Highway Driving} case study we achieved a precision of 96.2\%, a recall of 96.3\%, a weighted F1-score of 96.3\% and an F1-score of 82.9\%, after time step 10. By time step 20, the F1-score of \textit{SMARLA} reach 91.9\% while the weighted F1-score, precision and recall all reach 98.2\%. Also, at this time step, on average 30\% of the episodes remain to execute until reaching an unsafe state.} 

\minor{While binary features excel at early-stage predictions, the frequency of abstract state visits allows for improved prediction accuracy as more data becomes available (i.e., in later time steps when states are repeated). This is particularly useful in longer or more complex episodes, where additional data helps refine the model's predictions.}

\begin{table*}[t]
\small
\centering
\begin{tabular}{c|c|ccc|ccc|ccc|c}
 &
   &
  \multicolumn{3}{c|}{Decision time step} &
  \multicolumn{3}{c|}{Remaining time steps} &
  \multicolumn{3}{c|}{Remaining \% of time steps} &
   \\ \cline{3-11}
\multirow{-2}{*}{Decision criteria} &
  \multirow{-2}{*}{Case study} &
  \cellcolor[HTML]{D9D9D9}Min &
  \cellcolor[HTML]{D9D9D9}Avg &
  \cellcolor[HTML]{D9D9D9}Max & 
  \cellcolor[HTML]{D9D9D9}Min &
  \cellcolor[HTML]{D9D9D9}Avg &
  \cellcolor[HTML]{D9D9D9}Max &
  \cellcolor[HTML]{D9D9D9}Min &
  \cellcolor[HTML]{D9D9D9}Avg &
  \cellcolor[HTML]{D9D9D9}Max &
  \multirow{-2}{*}{\# FP} \\ \hline \hline
\multicolumn{1}{c|}{} &
  \multicolumn{1}{c||}{Mountain-Car} &
  15 &
  20.96 &
  38 &
  57 &
  74.21 &
  80 &
  59\% &
  77.16\% &
  83\% &
  2 \\
\multicolumn{1}{c|}{\multirow{-1}{*}{Upper bound}} &
  \multicolumn{1}{c||}{\cellcolor[HTML]{E7E6E6}Cart-Pole} &
  \cellcolor[HTML]{E7E6E6}22 &
  \cellcolor[HTML]{E7E6E6}41.36 &
  \cellcolor[HTML]{E7E6E6}105 &
  \cellcolor[HTML]{E7E6E6}67 &
  \cellcolor[HTML]{E7E6E6}104.5 &
  \cellcolor[HTML]{E7E6E6}164 &
  \cellcolor[HTML]{E7E6E6}38.70\% &
  \cellcolor[HTML]{E7E6E6}71.35\% &
  \cellcolor[HTML]{E7E6E6}82\% &
  \cellcolor[HTML]{E7E6E6}11 \\

\multicolumn{1}{c|}{} &
  \multicolumn{1}{c||}{Highway Driving } & 2  & 4.02  & 15  & 1  & 3.07  & 11  & 9.09\%  & 44.60\% & 66.67\% & 46 \\

\multicolumn{1}{c|}{} &
  \multicolumn{1}{c||}{\cellcolor[HTML]{E7E6E6}Highway Driving * } & 
  \cellcolor[HTML]{E7E6E6}2  & 
  \cellcolor[HTML]{E7E6E6}6.15  & 
  \cellcolor[HTML]{E7E6E6}20  & 
  \cellcolor[HTML]{E7E6E6}1  & 
  \cellcolor[HTML]{E7E6E6}3.96  & 
  \cellcolor[HTML]{E7E6E6}11  & 
  \cellcolor[HTML]{E7E6E6}5.56\%  & 
  \cellcolor[HTML]{E7E6E6}43.79\% & 
  \cellcolor[HTML]{E7E6E6}66.67\% & 
  \cellcolor[HTML]{E7E6E6}67 \\

\multicolumn{1}{c|}{} &
  \multicolumn{1}{c||}{ Mountain-Car} &
   20 &
   32.53 &
   63 &
   32 &
   62.65 &
   75 &
   33.33\% &
   65.15\% &
   78.12\% &
   1 \\
\multicolumn{1}{c|}{\multirow{-1}{*}{Output probability}} &
  \multicolumn{1}{c||}{\cellcolor[HTML]{E7E6E6}Cart-Pole} &
  \cellcolor[HTML]{E7E6E6}30 &
  \cellcolor[HTML]{E7E6E6}48.8 &
  \cellcolor[HTML]{E7E6E6}126 &
  \cellcolor[HTML]{E7E6E6}46 &
  \cellcolor[HTML]{E7E6E6}96.48 &
  \cellcolor[HTML]{E7E6E6}129 &
  \cellcolor[HTML]{E7E6E6}27\% &
  \cellcolor[HTML]{E7E6E6}66.58\% &
  \cellcolor[HTML]{E7E6E6}78.60\% &
  \cellcolor[HTML]{E7E6E6}2 \\

\multicolumn{1}{c|}{} &
  \multicolumn{1}{c||}{ Highway Driving} & 
   2  & 
   4.16  & 
   15  & 
   1  & 
   2.92  & 
   11  & 
   9.09\%  & 
   42.81\% & 
   66.67\% & 
   40 \\

\multicolumn{1}{c|}{} &
  \multicolumn{1}{c||}{\cellcolor[HTML]{E7E6E6}Highway Driving * } & 
  \cellcolor[HTML]{E7E6E6}2  & 
  \cellcolor[HTML]{E7E6E6}6.71  & 
  \cellcolor[HTML]{E7E6E6}22  & 
  \cellcolor[HTML]{E7E6E6}1  & 
  \cellcolor[HTML]{E7E6E6}3.55  & 
  \cellcolor[HTML]{E7E6E6}11  & 
  \cellcolor[HTML]{E7E6E6}5.56\%  & 
  \cellcolor[HTML]{E7E6E6}37.69\% & 
  \cellcolor[HTML]{E7E6E6}64.28\% & 
  \cellcolor[HTML]{E7E6E6}51 \\

\multicolumn{1}{c|}{} &
  \multicolumn{1}{c||}{Mountain-Car} &
  25 &
  44.39 &
  76 &
  19 &
  50.79 &
  70 &
  19.79\% &
  52.81\% &
  72.92\% &
  1 \\
\multicolumn{1}{c|}{\multirow{-1}{*}{Lower bound}} &
  \multicolumn{1}{c||}{\cellcolor[HTML]{E7E6E6}Cart-Pole} &
  \cellcolor[HTML]{E7E6E6}33 &
  \cellcolor[HTML]{E7E6E6}56.94 &
  \cellcolor[HTML]{E7E6E6}171 &
  \cellcolor[HTML]{E7E6E6}8 &
  \cellcolor[HTML]{E7E6E6}83.36 &
  \cellcolor[HTML]{E7E6E6}123 &
  \cellcolor[HTML]{E7E6E6}0.49\% &
  \cellcolor[HTML]{E7E6E6}61.34\% &
  \cellcolor[HTML]{E7E6E6}73.33\% &
  \cellcolor[HTML]{E7E6E6}0 \\

\multicolumn{1}{c|}{} &
  \multicolumn{1}{c||}{Highway Driving} & 2  & 3.6   & 10  & 1  & 2.96  & 11  & 9.09\%  & 45.13\% & 66.67\% & 38 \\

\multicolumn{1}{c|}{} &
  \multicolumn{1}{c||}{\cellcolor[HTML]{E7E6E6}Highway Driving * } & 
  \cellcolor[HTML]{E7E6E6}3  & 
  \cellcolor[HTML]{E7E6E6}6.40  & 
  \cellcolor[HTML]{E7E6E6}20  & 
  \cellcolor[HTML]{E7E6E6}1  & 
  \cellcolor[HTML]{E7E6E6}2.93  & 
  \cellcolor[HTML]{E7E6E6}10  & 
  \cellcolor[HTML]{E7E6E6}11.11\%  & 
  \cellcolor[HTML]{E7E6E6}31.81\% & 
  \cellcolor[HTML]{E7E6E6}57.14\% & 
  \cellcolor[HTML]{E7E6E6}39 \\

\end{tabular}
\captionsetup{width=.95\textwidth}
\caption{Overview of \textit{SMARLA}'s decision times and the remaining percentage of time steps before safety violations occur across different decision criteria and case studies (\# FP stands for the total number of false positives and \textit{Highway Driving}* refers to \textit{Highway Driving} case study using frequency-based features)}
\label{tab:Prediction_tme}
\end{table*}

\begin{tcolorbox}
\textbf{Answer to RQ1:} \textit{SMARLA} demonstrated high accuracy in predicting safety violations from RL agents. Moreover, such accurate predictions can be obtained early during the execution of episodes, thus enabling the system to prevent or mitigate damages. 
\end{tcolorbox}

\subsubsection{RQ2. How can the safety monitor determine when to trust the prediction of safety violations?} \label{Sec:Rq2Results}
In this research question, we investigate the use of confidence intervals as a means for the safety monitor to determine the appropriate time step to trigger safety mechanisms. This investigation is based on the same set of episodes randomly generated for RQ1 (Section~\ref{Sec:RQ1}). 
At each time step $t$, we collect the predicted probability of safety violation $P_{e_i}(t)$ and the corresponding confidence interval $[Low(t),Up(t)]$. The lower bound ($Low(t)$) and upper bound ($Up(t)$) of the confidence interval are computed using the methodology detailed in Section~\ref{Subsec: Safety Violation}.

To determine the best decision criterion for triggering safety mechanisms, we considered and compared the following alternative criteria:
\begin{itemize} 
    \item If the probability of safety violation, $P_{e_i}(t)$, is equal to or greater than 50\%, then the safety mechanism is activated.
    \item If the upper bound of the confidence interval at time step $t$ (based on the confidence level of 95\%) is above 50\% (i.e., $Up(t) \geq 50\%$), then the safety mechanism is activated. This is a conservative approach as the actual probability has a 97.5\% chance to be below that value. This may result in many false positives but it leads to early predictions of unsafe episodes and is unlikely to miss any unsafe episodes.
    \item If the lower bound of the confidence intervals at time step $t$ is above 50\% (i.e., $Low(t) \geq 50\%$), then the safety mechanism is activated. In this criterion, the actual probability has only a 2.5\% chance to be below that bound and we thus minimize the occurrence of false positives, at the cost of relatively late detection of unsafe episodes and more false negatives. 
\end{itemize}

Decision criteria identify the time step when the execution should be stopped and safety mechanisms should be activated. However, note that during our test, we continue the execution of the episodes until termination in order to extract the number of time steps until termination and the true label of episodes for our analysis.

\begin{figure}[t]
    \centering
    \includegraphics[width=\columnwidth]{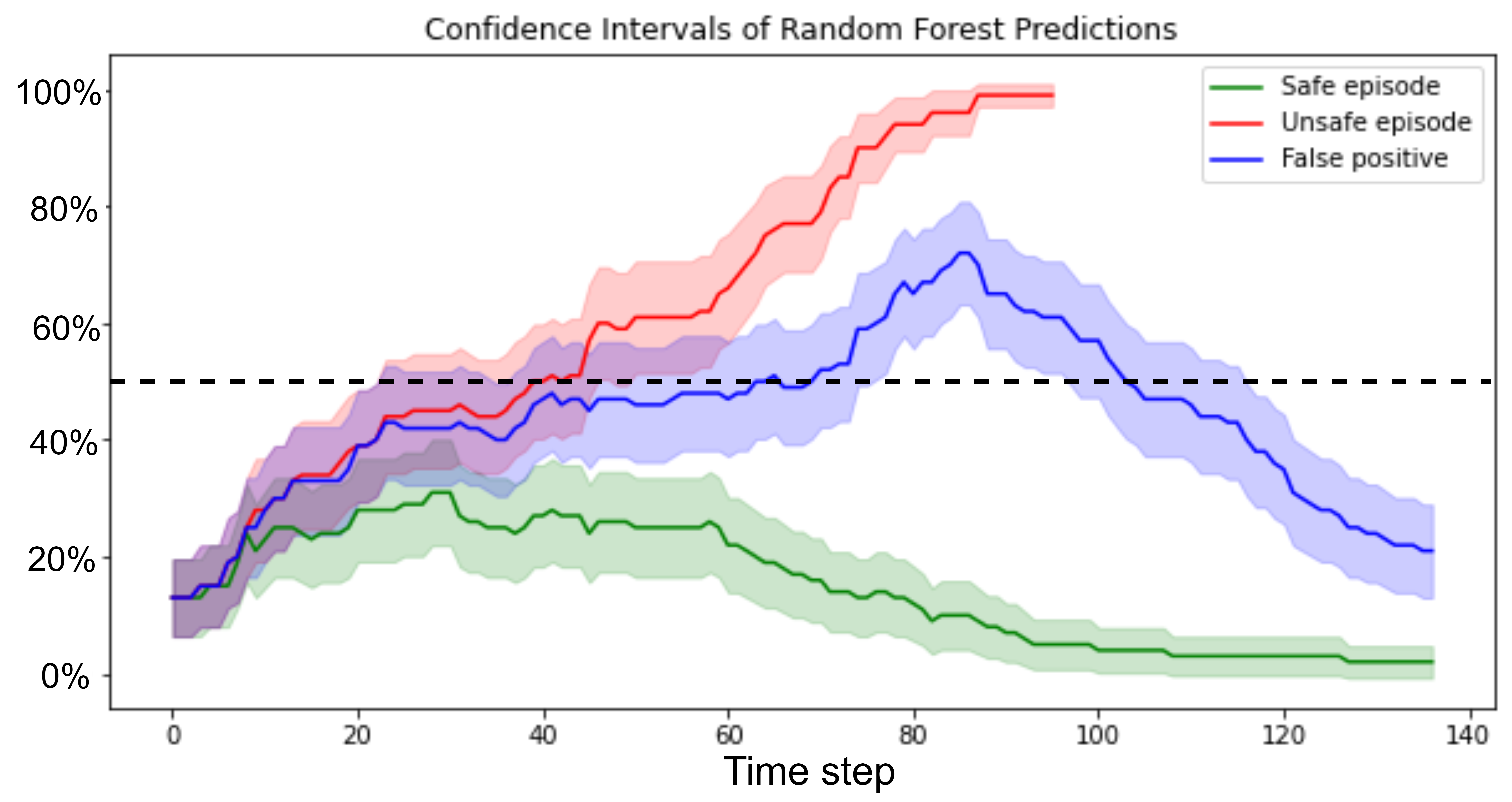}
    \caption{Confidence intervals of a safe episode (in green), an unsafe episode (in red) and a false positive episode (in blue) in the \textit{Mountain-Car} case study}
    \label{Rq2_CI example Mountain car}
\end{figure}

In Figure~\ref{Rq2_CI example Mountain car}, we provide representative examples of the estimated probability of safety violation over time for a safe episode, an unsafe episode, and a mispredicted episode (false positive) in the \textit{Mountain-Car} case study. The curves in the graph represent the output probabilities generated by the safety violation prediction model, while the shaded areas depict the corresponding 95\% confidence intervals. We observe that, as more information is acquired, probabilities either tend to increase or decrease depending on whether the episode is safe or unsafe. At some point, the confidence intervals tend to narrow down as enough information from episodes is collected to make an accurate prediction. As expected, in the figure, relying on the upper bound of the confidence interval leads to a much earlier safety violation prediction in the unsafe episode (depicted in red in the figure), compared to when considering the output probability or the lower bound of the confidence interval.

\begin{figure}[t]
    \includegraphics[width=\linewidth]{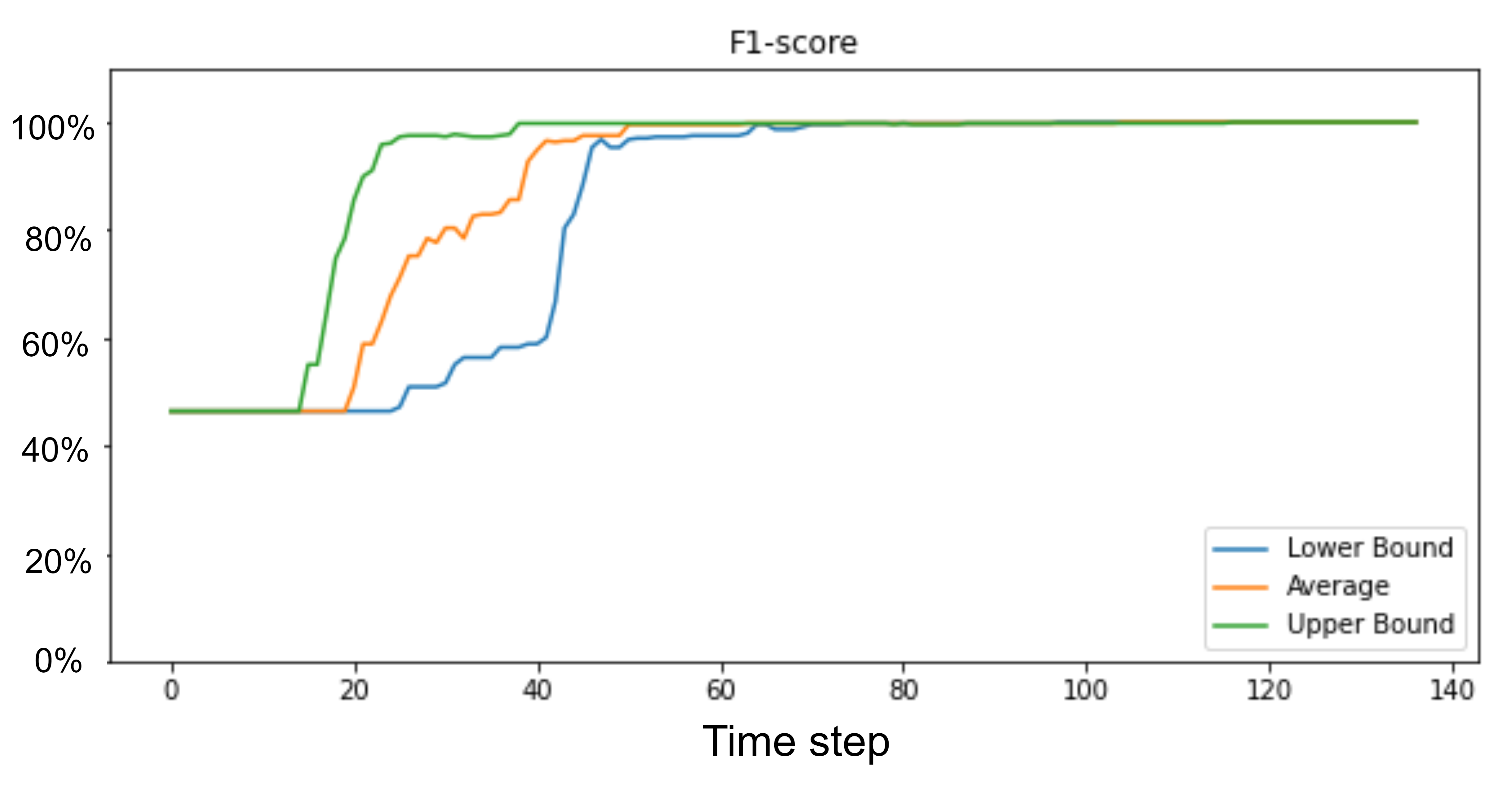}
    \caption{F1-score of the safety violation prediction model for different decision criteria in the \textit{Mountain-Car} case study}
    \label{Rq2_Mountain-Car_f1}
\end{figure}

\begin{figure}[t]
    \centering
    \includegraphics[width=\linewidth]{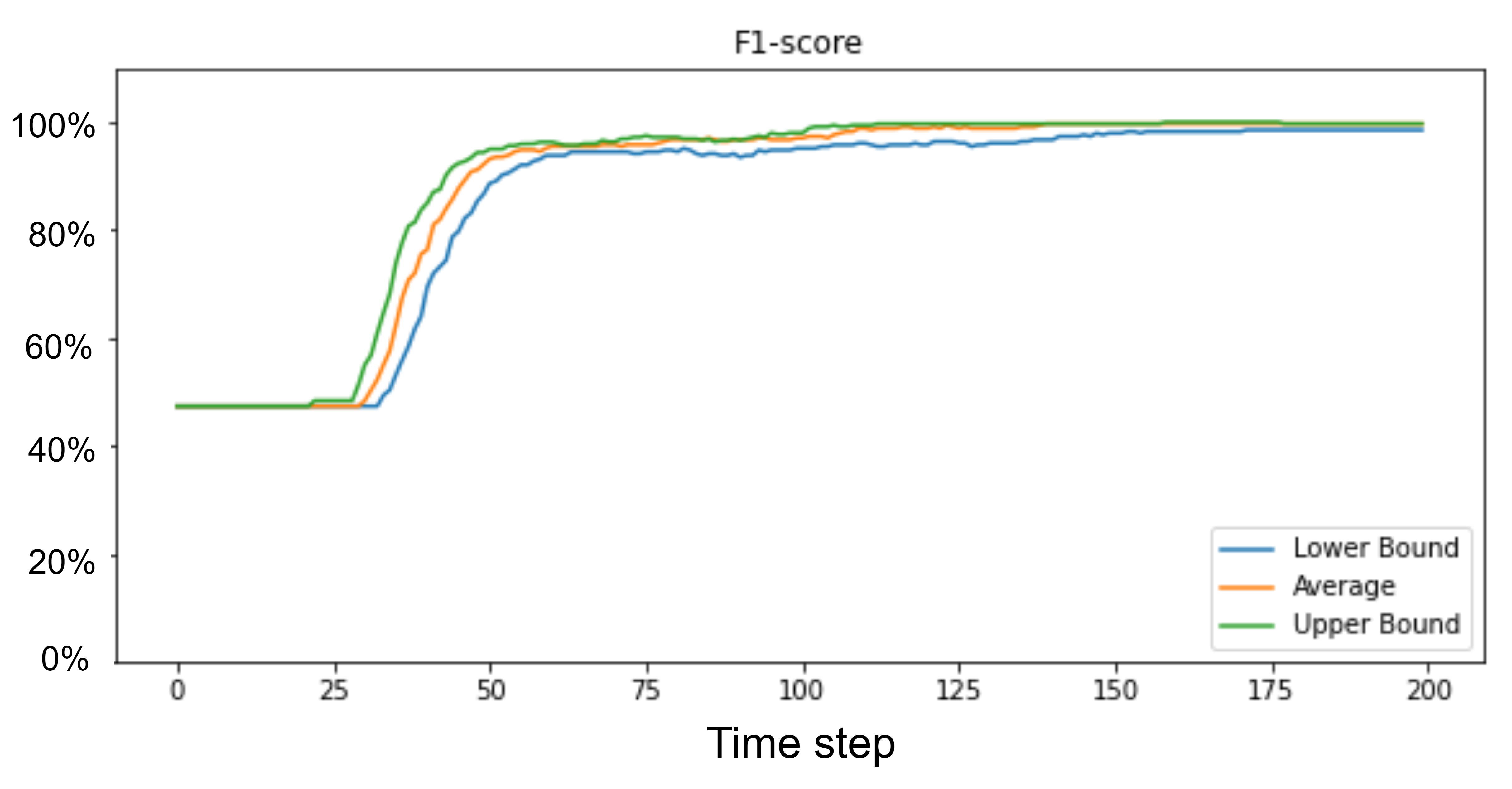}
    \caption{F1-score of the safety violation prediction model for different decision criteria in the \textit{Cart-Pole} case study}
    \label{Rq2_Cart-Pole_f1}
\end{figure}

\begin{figure}[ht]
    \centering
    \includegraphics[width=\linewidth]{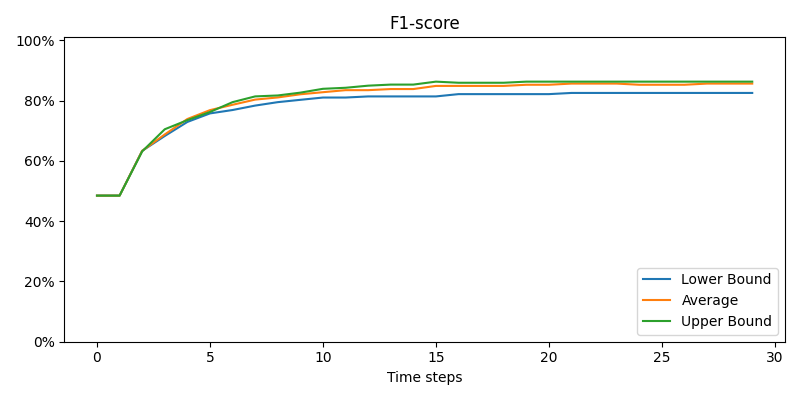}
    \caption{\added{F1-score of the safety violation prediction model for different decision criteria in the \textit{Highway} case study}}
    \label{Rq2_Highway_f1}
\end{figure}

\begin{figure}[ht]
    \centering
    \includegraphics[width=\linewidth]{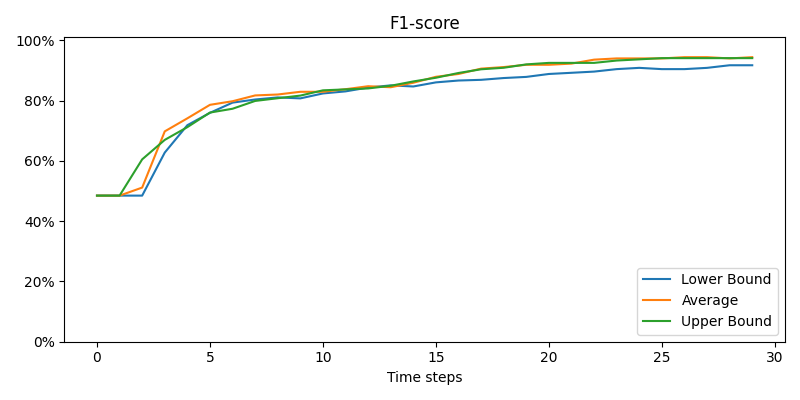}
    \caption{\minor{F1-score of the safety violation prediction model for different decision criteria in the \textit{Highway} case study using frequency-based feature representation}}
    \label{Rq2_Highway_frequency_f1}
\end{figure}

Now the question is how do the predictions based on the three above criteria compare in terms of accuracy. \added{Figures~\ref{Rq2_Mountain-Car_f1},~\ref{Rq2_Cart-Pole_f1} and~\ref{Rq2_Highway_f1} present a comparison of the F1-scores of the three predictions at each time step for the \textit{Mountain-Car}, \textit{Cart-Pole} and \textit{Highway Driving} case studies, respectively. }
\minor{Additionally, Figure~\ref{Rq2_Highway_frequency_f1} presents the comparison of decision criteria for the \textit{Highway Driving} case study, using frequency-based features.}
Though there are differences in the magnitude of the trend, results from our case studies consistently show that using the upper bound is the best choice as it leads to early accurate predictions.

More in detail, we observe that in the \textit{Cart-Pole} case study, the F1-score based on the upper bound is above 95\% from time step 50 and plateaus around 99\% at time step 102. On the other hand, when using the output probability of the safety violation prediction model, the F1-score is above 95\% starting from time step 59 and reaches its plateau of 99\% at time step 135. With the lower bound, however, F1-score is above 95\% from time step 99 and reaches a plateau of 98\% at time step 150.
Also, for the \textit{Mountain-Car} case study, with the upper bound we reach an F1-score above 95\% from time step 23 and plateau around 99\% at time step 38. In contrast, the F1-score based on the output probability of the safety violation prediction model is above 95\% from time step 41 and plateaus around 99\% at time step 50. Finally, with the lower bound, the F1-score is above 95\% from time step 46, and plateaus around 99\% at time step 69.


\added{In the Highway Driving case study, using the upper bound as a decision criterion allowed us to achieve an F1-score that plateaus at 85\% by time step 12. This early stabilization underscores the robustness of the upper bound criterion in rapidly identifying safety violations. Similarly, when using the output probability of the safety violation prediction model, we observed that the F1-score exceeds 82\% by time step 10 and reaches a plateau around 85\% by time step 15, demonstrating close and effective results. However, when applying the lower bound criterion, the F1-score stabilizes later, plateauing at 83\% by time step 21. This delay illustrates a more conservative approach, which, while reducing false positives, may also delay the prediction of safety violations.}

\minor{In the Highway Driving case study using frequency-based features, applying the upper bound as a decision criterion resulted in an F1-score of 82\% by time step 10. Achieving high accuracy predictions early on time highlights our ability to quickly and effectively detect safety violations. However, this accuracy continues to improve over time, reaching an F1-score of 90\% and 94\% at time step 18 and time step 25, respectively. Using the output probability of the safety violation prediction model yields similar performance to the upper bound, with the F1-score at 82\% by time step 10, increasing to 91\% by time step 18, demonstrating comparable effectiveness. In contrast, the lower bound criterion led to a slower stabilization, with the F1-score reaching 85\% by time step 15 and achieving an F1-score of 90\% by time step 23.}

We also observe that, for all three decision criteria in each case study, the F1-score is low during early steps due to lack of information. \added{It then increases sharply (around time steps 30-40 for \textit{Cart-Pole},  15-45 for \textit{Mountain-Car} and 2-5 for \textit{Highway Driving}) to reach a plateau shortly after 50 time steps for both \textit{Cart-Pole} and \textit{Mountain-Car}, while it plateaus after 15 time steps for the \textit{Highway Driving} case study. During that time frame, the difference among decision criteria for \textit{Mountain-Car} is more significant than for \textit{Cart-Pole} and \textit{Highway Driving}.  }
This discrepancy is due to the wider confidence intervals present in the \textit{Mountain-Car} case study, which can be explained by wider variance in decision tree predictions. Furthermore, the high F1-scores for all three decision criteria indicate that the confidence intervals significantly narrow once the plateau is reached. At that point, there is no overlap between the confidence intervals of unsafe and safe episodes.

Using these observations, we computed the average improvements, in terms of time steps required to achieve peak performance, when predicting safety violations by considering the upper bound of the confidence interval, in contrast to (1) the output probability and (2) the lower bound of the confidence interval. 
\added{Our findings indicate that using the upper bound of the confidence intervals results in an average decrease of 24\% for both \textit{Cart-Pole} and \textit{Mountain-Car}, 20\% for \textit{Highway Driving} in terms of time steps required to achieve peak performance, as compared to using the predicted probability. When compared to using the lower bound, the decrease is 32\% for \textit{Cart-Pole}, 45\% for \textit{Mountain-Car} and 43\% for \textit{Highway Driving}.}

\minor{Regarding \textit{Highway Driving} with frequency-based features, we observed similar performance in terms of prediction time steps when considering the upper bound and prediction probability. However, there was a 32\% improvement compared to the lower bound.}

We also investigated in all case studies three important metrics: (1) the decision time step, (2) the remaining time steps until the occurrence of a safety violation, and (3) the remaining percentage of time steps to execute until violation. For each metric, we present in Table~\ref{tab:Prediction_tme} the minimum, maximum, and average values.
The result in Table~\ref{tab:Prediction_tme} suggests that, relying on the upper bound, the average decision time step in the \textit{Mountain-Car} case study is 21. However, in the best-case scenario, safety violations are predicted as early as time step 15, while in the worst-case scenario, such violations are predicted at time step 38. Notably, the results demonstrate that when safety mechanisms are triggered, on average 74 times steps (77\%) of episodes remain to be executed. This observation suggests there is ample time to initiate safety mechanisms and hopefully prevent safety violations.

Also,  in the \textit{Cart-Pole} case study, employing the upper bound, the average prediction time step is determined to be 41. The earliest prediction of a safety violation occurs at time step 22, while the latest prediction is observed at time step 105. Remarkably, on average, there are still 104 time steps remaining to be executed when safety mechanisms are triggered. This accounts for approximately 71\% of the average length of episodes, once again suggesting there is significant time available for carrying out safety measures. 
\added{Similarly, in the \textit{Highway Driving} case study, when using the upper bound of confidence intervals, the average prediction time step is 4, the earliest prediction of a safety violation occurs at time step 2 while the latest prediction is made at time step 15. Also, at the time of prediction, 44.6\% of the average length of episodes still remains to be executed. } 

\minor{Regarding the Highway Driving case study using frequency of abstract states as features, with the upper bound of confidence intervals, the average prediction time step is 6, the earliest prediction of a safety violation occurs at time step 2 while the latest prediction is made at time step 20. Also, at the time of prediction, 43.8\% of the average length of episodes still remains to be executed.}
\minor{Overall, the above results demonstrate that \textit{SMARLA} can predict safety violations early and accurately, suggesting it is a good solution for ensuring system safety when relying on RL agents.} 
Furthermore, in the \textit{Cart-Pole} case study, our observations revealed the occurrence of 11 false positives leading to a false positive rate (FPR) of 1\%, when relying on the upper bound. In contrast, only two false positives were identified with the output probability ($FPR = 0.2\%$), and none when employing the lower bound.
Regarding the \textit{Mountain-Car} case study, two false positives were observed ($FPR = 0.2\%$) when employing the upper bound. In contrast, using the output probability or the lower bound resulted in only one false positive ($FPR = 0.1\%$). 

\added{Finally, in the \textit{Highway Driving} case study, we obtained 46 false positives leading to an FPR of 4.8\% when using the upper bound and 40 false positives with an FPR of 4.2\% considering the output probability. Furthermore, we identified 38 false positives ($FPR = 4\%$), when using the lower bound.}
\minor{Similarly, when using frequency-based features, we observed 67 false positives, resulting in an FPR of 7.1\% with the upper bound, and 51 false positives with an FPR of 5.4\% when considering the output probability. Additionally, we identified 39 false positives ($FPR = 4.1\%$) when applying the lower bound.
}

To summarize, while relying on the upper bound of confidence intervals leads to an earlier prediction of safety violations, it also introduces a higher rate of false positives compared to using the predicted probability and the lower bound. Therefore, considering the trade-off between earlier detection of safety violations and the number of false positives, the selection of an appropriate decision criterion relies heavily on the level of criticality of the RL agent. For instance, in certain scenarios, prioritizing early detection of safety violations and allowing for a longer time frame to apply safety mechanisms may be of critical importance, even at the expense of a slightly higher false positive rate. Conversely, in other cases, there might be a preference to sacrifice time in order to optimize accuracy and minimize the occurrence of false positives. The selection of the appropriate decision criterion depends on the specific context and the relative importance of early detection and prediction accuracy. In our case studies, the increase in false positives appears to be limited, and therefore using the upper bound is the best option.

\begin{tcolorbox}
\textbf{Answer to RQ2:}  Considering the upper bound of the confidence intervals leads to a significantly earlier and still highly accurate detection of safety violations. This provides a longer time frame for the system to apply preventive safety measures and mitigate potential damages. This, however, comes at the expense of a slightly higher false positive rate.  
\end{tcolorbox}

\subsubsection{RQ3. What is the effect of the prediction threshold on the safety monitoring system?}
\label{Subsec:RQ3Threshold}

\begin{figure*}[htbp]
    \centering
    \begin{subfigure}[b]{0.33\textwidth}
        \centering
        \includegraphics[width=\textwidth]{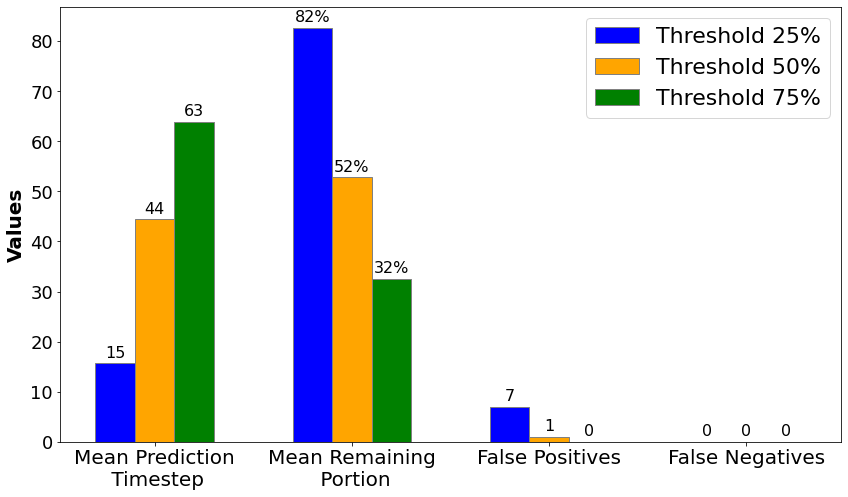}
        \caption{\added{\textit{Mountain-Car} lower bound}}
        \label{fig:sub1}
    \end{subfigure}
    \hfill
    \begin{subfigure}[b]{0.33\textwidth}
        \centering
        \includegraphics[width=\textwidth]{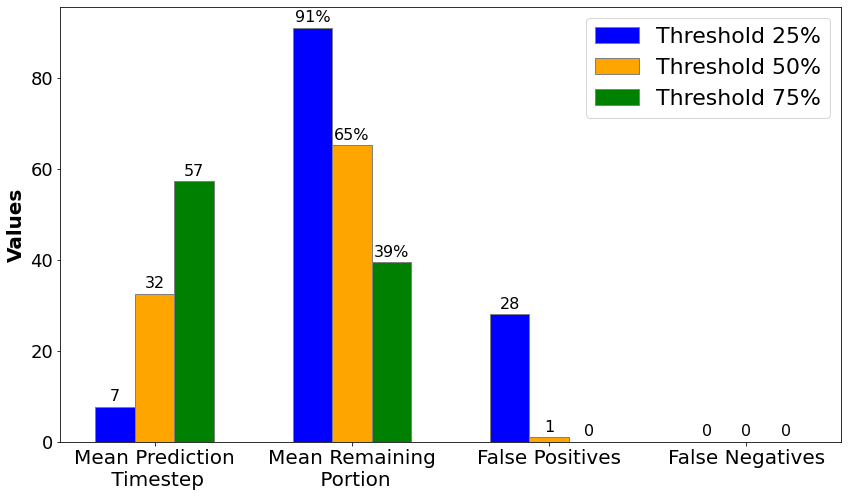}
        \caption{\added{\textit{Mountain-Car} prediction probability}}
        \label{fig:sub2}
    \end{subfigure}
    \hfill
    \begin{subfigure}[b]{0.33\textwidth}
        \centering
        \includegraphics[width=\textwidth]{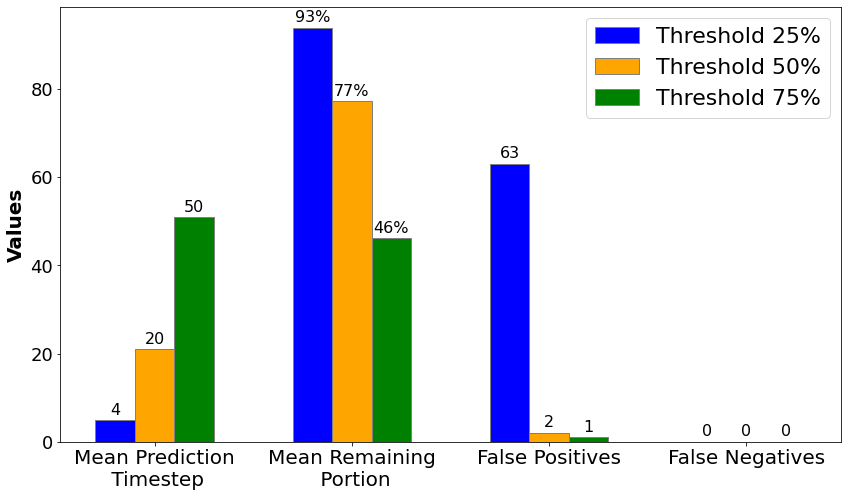}
        \caption{\added{\textit{Mountain-Car} upper bound}}
        \label{fig:sub3}
    \end{subfigure}
    
    \vspace{0.15cm}
    
    \begin{subfigure}[b]{0.33\textwidth}
        \centering
        \includegraphics[width=\textwidth]{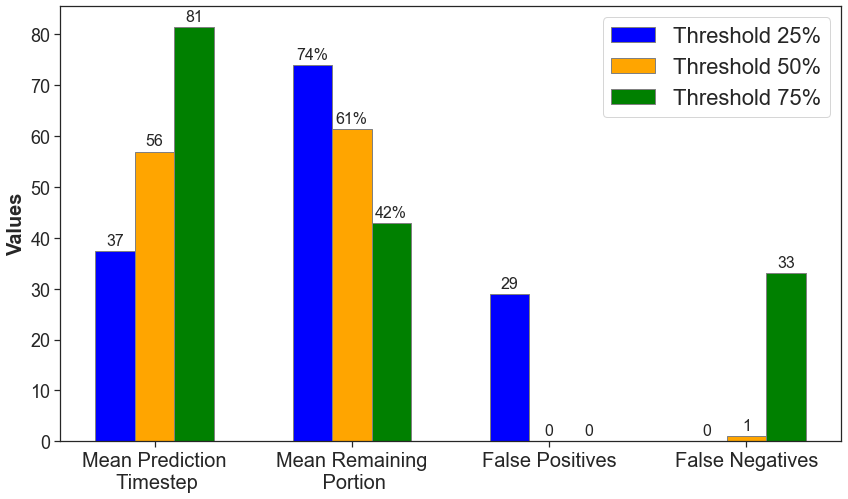}
        \caption{\added{\textit{Cart-Pole} lower bound}}
        \label{fig:sub4}
    \end{subfigure}
    \hfill
    \begin{subfigure}[b]{0.33\textwidth}
        \centering
        \includegraphics[width=\textwidth]{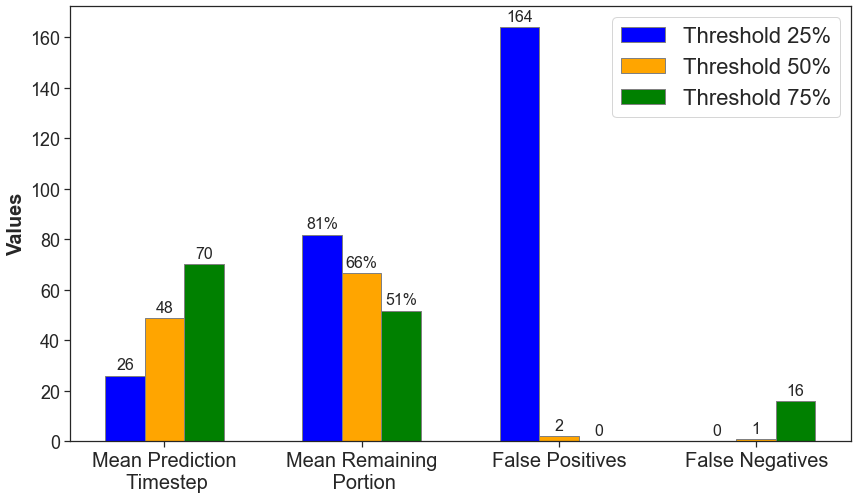}
        \caption{\added{\textit{Cart-Pole} prediction probability}}
        \label{fig:sub5}
    \end{subfigure}
    \hfill
    \begin{subfigure}[b]{0.33\textwidth}
        \centering
        \includegraphics[width=\textwidth]{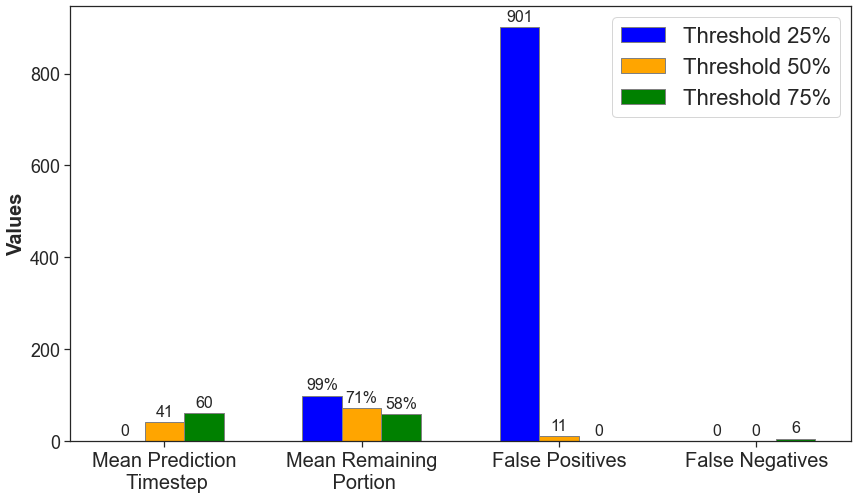}
        \caption{\added{\textit{Cart-Pole} upper bound}}
        \label{fig:sub6}
    \end{subfigure}

    \vspace{0.15cm}

    \begin{subfigure}[b]{0.33\textwidth}
        \centering
        \includegraphics[width=\textwidth]{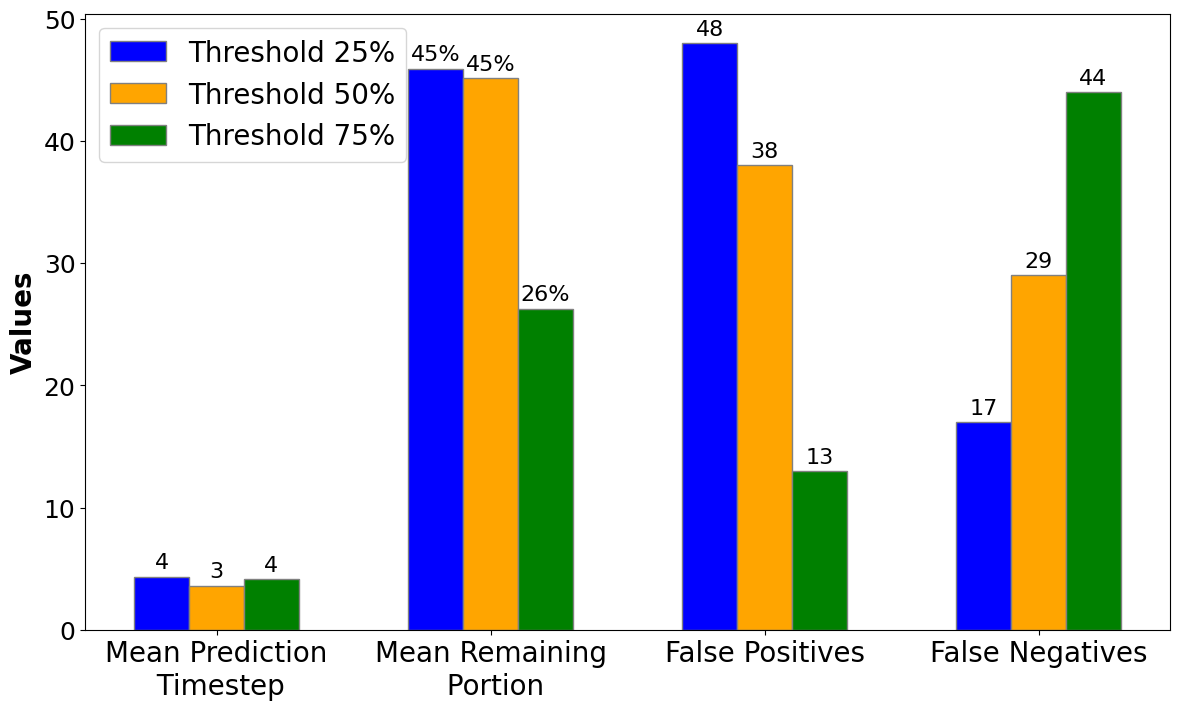}
        \caption{\added{\textit{Highway Driving} lower bound}}
        \label{fig:sub7}
    \end{subfigure}
    \hfill
    \begin{subfigure}[b]{0.33\textwidth}
        \centering
        \includegraphics[width=\textwidth]{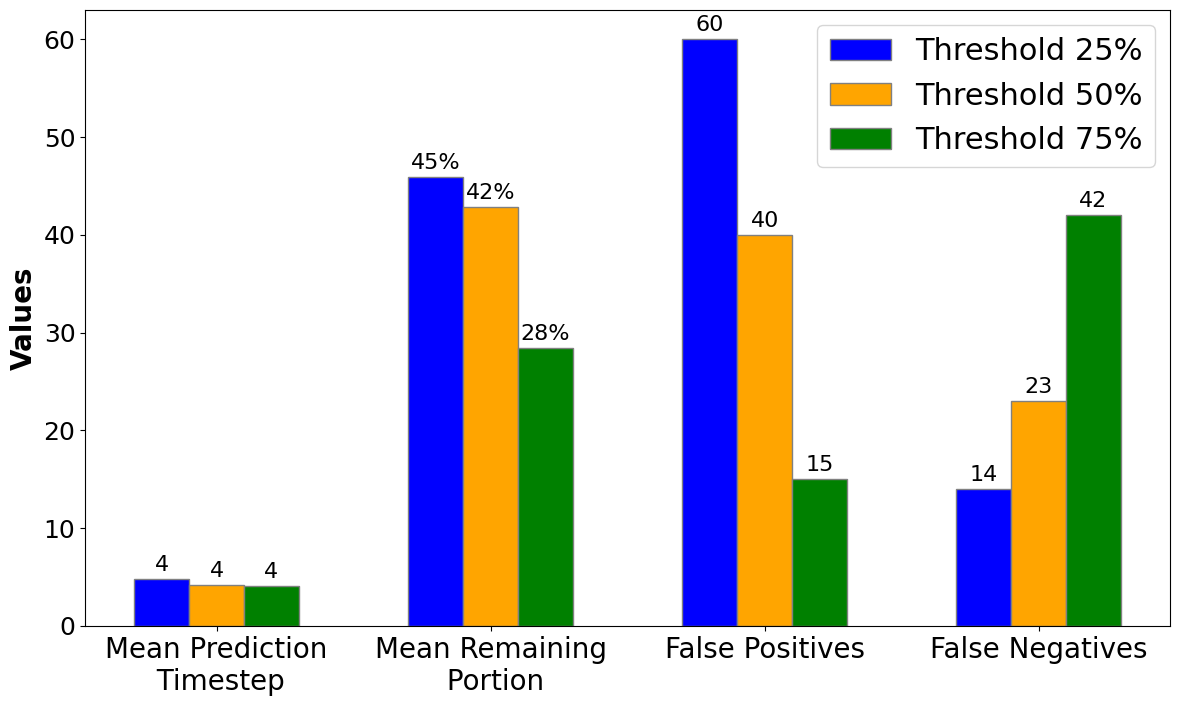}
        \caption{\added{\textit{Highway Driving} prediction probability}}
        \label{fig:sub8}
    \end{subfigure}
    \hfill
    \begin{subfigure}[b]{0.33\textwidth}
        \centering
        \includegraphics[width=\textwidth]{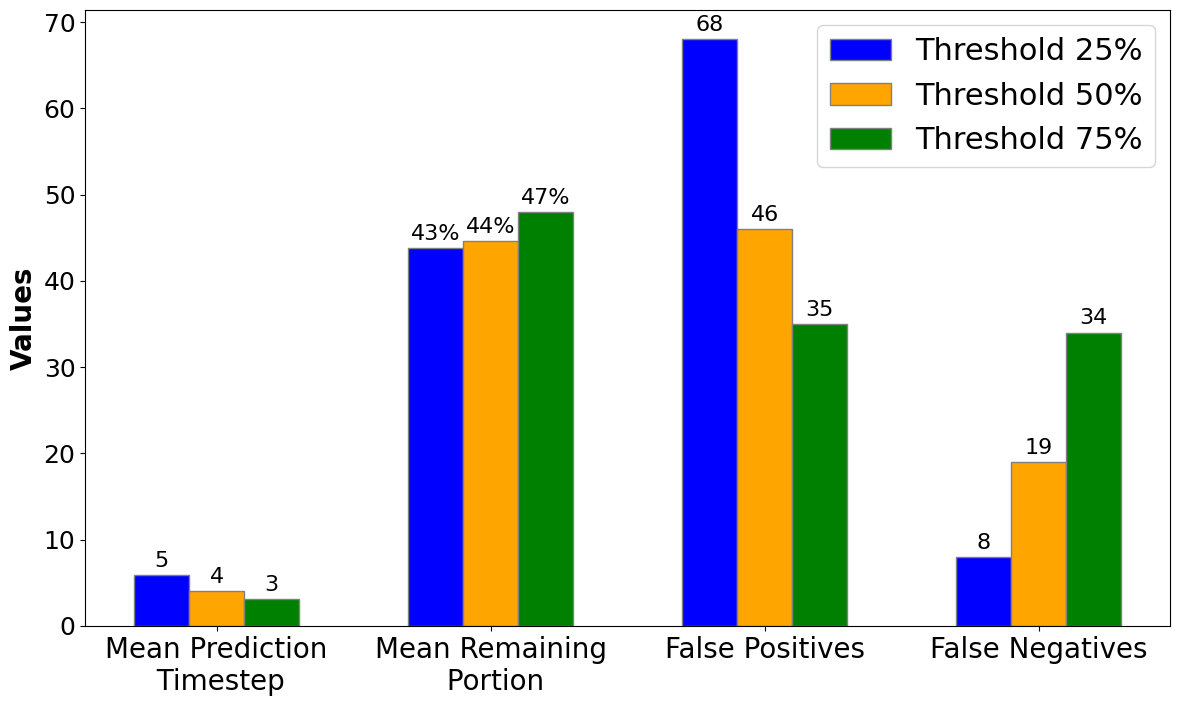}
        \caption{\added{\textit{Highway Driving} upper bound}}
        \label{fig:sub9}
    \end{subfigure}

    \vspace{0.15cm}

    \begin{subfigure}[b]{0.33\textwidth}
        \centering
        \includegraphics[width=\textwidth]{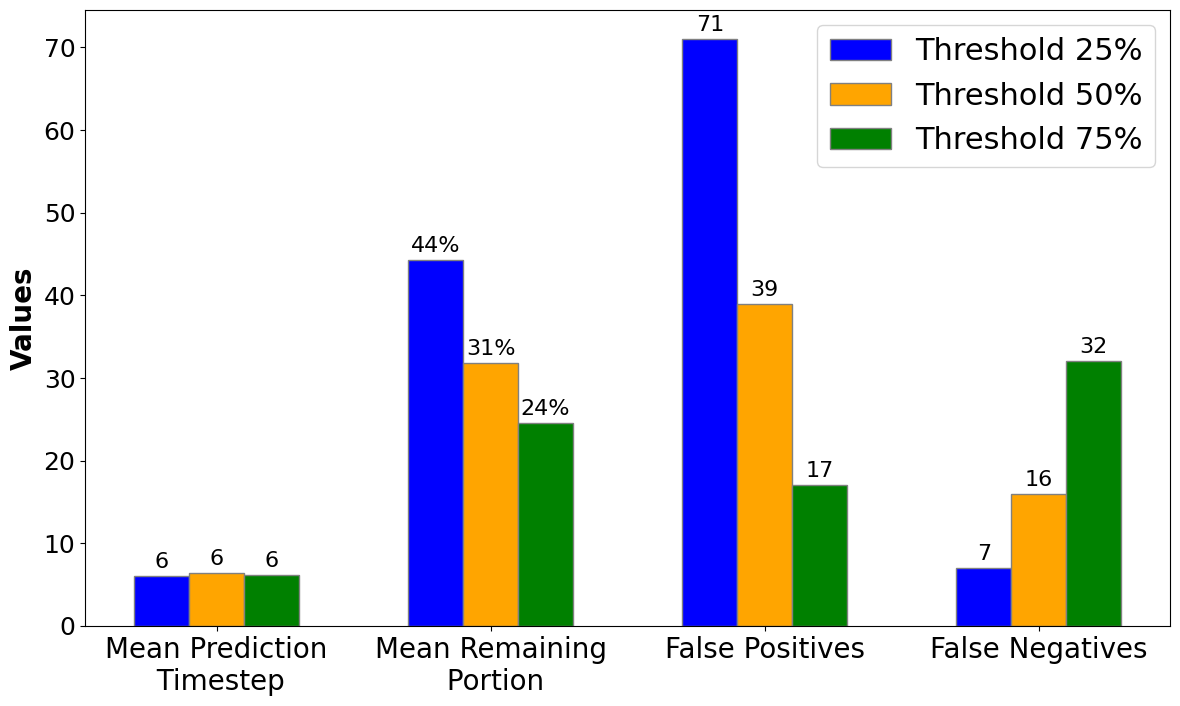}
        \caption{\minor{\textit{Highway Driving}* lower bound}}
        \label{fig:sub10}
    \end{subfigure}
    \hfill
    \begin{subfigure}[b]{0.33\textwidth}
        \centering
        \includegraphics[width=\textwidth]{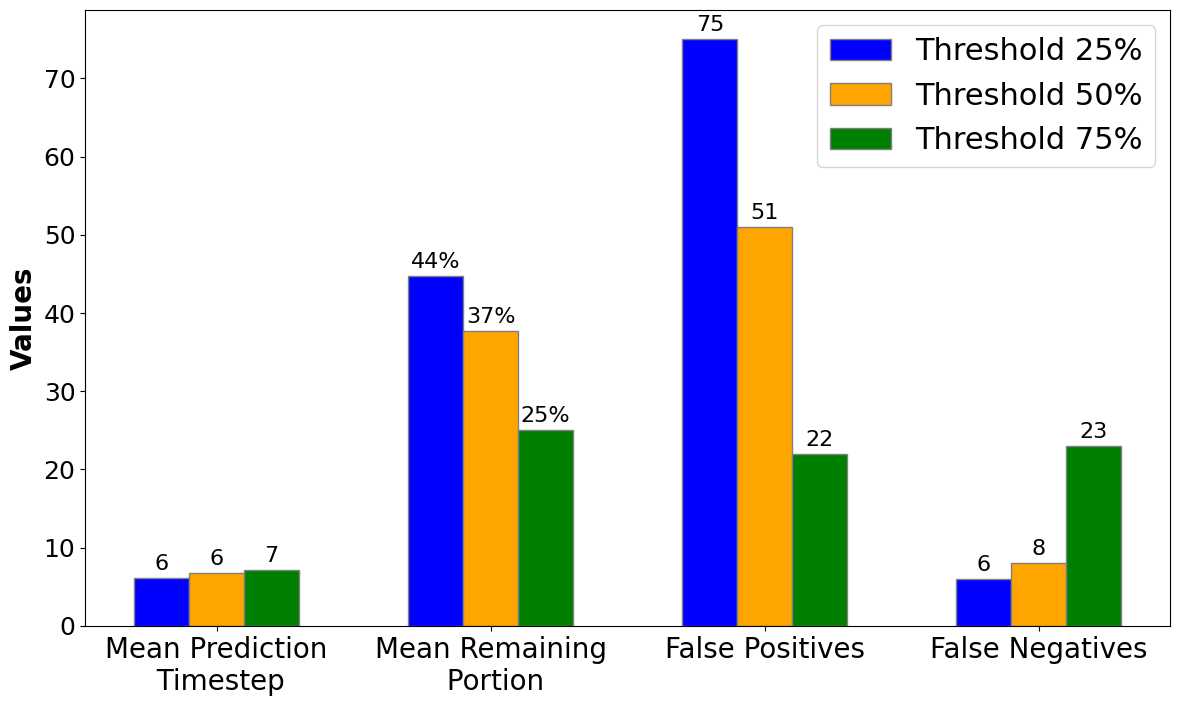}
        \caption{\minor{\textit{Highway Driving}* prediction probability}}
        \label{fig:sub11}
    \end{subfigure}
    \hfill
    \begin{subfigure}[b]{0.33\textwidth}
        \centering
        \includegraphics[width=\textwidth]{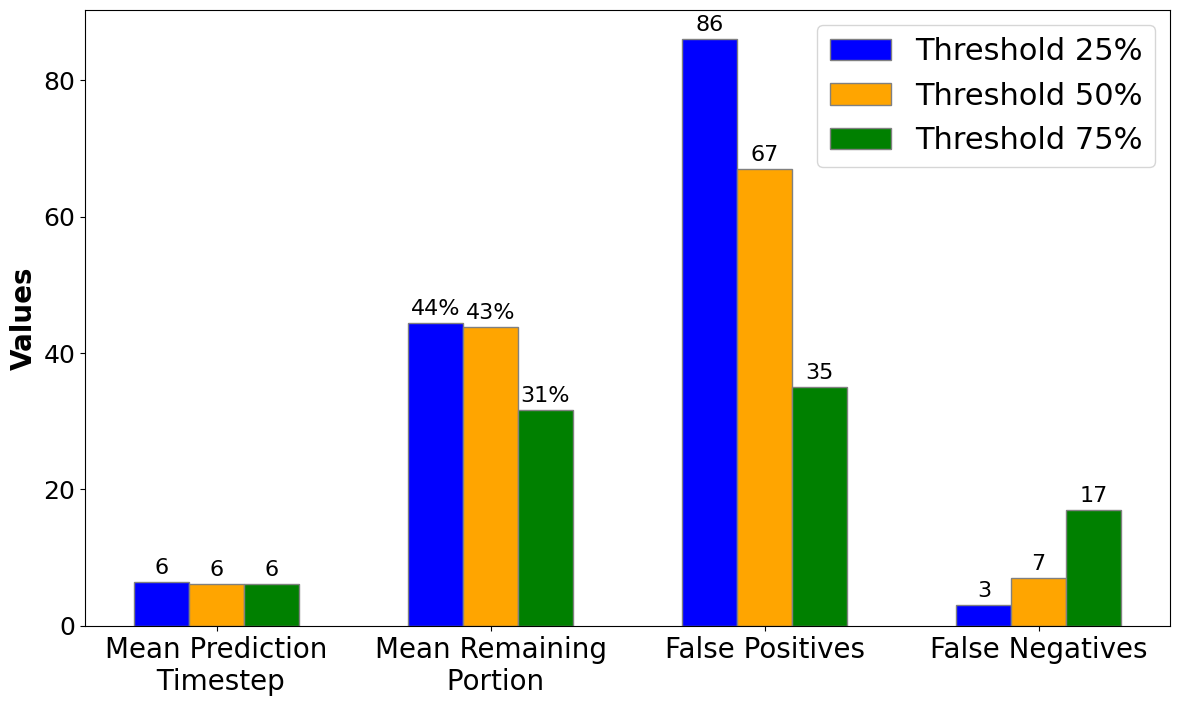}
        \caption{\minor{\textit{Highway Driving}* upper bound}}
        \label{fig:sub12}
    \end{subfigure}

    \captionsetup{width=\textwidth}
    \caption{\minor{Comparison of 25\%, 50\% and 75\% threshold values for \textit{Mountain-Car}, \textit{Cart-Pole}, \textit{Highway Driving} and \textit{Highway Driving} using frequency-based features (\textit{Highway Driving}*)}}
    \label{fig:main}
\end{figure*}

\added{We aim in this research question to study the impact of varying the prediction threshold values on our safety monitoring approach. We evaluate the prediction times and performance of \textit{SMARLA} using different prediction thresholds based on the same set of episodes randomly generated in RQ1 (Section~\ref{Sec:RQ1}). Specifically, for each case study, we consider three prediction thresholds $\theta \in \{25\%, 50\%, 75\%\}$ and assess the performance of \textit{SMARLA} according to the three decision criteria explained in Section~\ref{Sec:Rq2Results} ($P(t) \geq \theta$, $Up(t) \geq \theta$, and $Low(t) \geq \theta$). For each case study and evaluation criteria, we report (1) the average decision time step, (2) the average remaining percentage of time steps to execute until violation, (3) the number of false positives of the safety monitor, and (4) the number of false negatives. Results are reported in Figure~\ref{fig:main}. } 

\added{As visible from the figure, we obtain consistent results over the three decision criteria and case studies. As expected, varying the prediction threshold does impact the average prediction time step, the percentage of the remaining portion of the episode to be executed, as well as the number of false positives and false negatives. More precisely, having a lower prediction threshold, such as $\theta=25\%$, leads to an earlier decision time to predict safety violations, which may be critical for taking preemptive safety measures, but increases the occurrence of false positives. On the other hand, higher thresholds, such as $\theta=75\%$, reduce false positives at the expense of delayed safety violation detection. Thus, choosing a high threshold can potentially compromise the system's ability to promptly initiate safety mechanisms and effectively prevent safety violations. Finally, we found that the intermediate threshold $\theta=50\%$ offers a balanced solution for all case studies, minimizing false positives and false negatives while still providing timely predictions. We therefore rely on such a threshold value in our experiments as it offers the best trade-off between accurate and early predictions of safety violations. We should note, however, that determining a suitable prediction threshold is context-dependent and relies heavily on the level of criticality of the RL agent. In practice, we recommend building a testing dataset that encompasses both safe and unsafe episodes based on random executions of the agent.  Then, we recommend trying different prediction thresholds and conducting a sensitivity analysis specific to the deployment environment of the DRL agent, to determine an optimal threshold that offers the best trade-off between timely and accurate prediction of safety violations.}

\begin{tcolorbox}
\added{\textbf{Answer to RQ3:}  The performance of safety monitoring is sensitive to the prediction threshold $\theta$. Higher prediction thresholds reduce false positives but lead to a delayed prediction of safety violations and higher false negatives, whereas lower thresholds lead to earlier predictions and lower false negatives but at the cost of increased false positives. Therefore, selecting an optimal threshold is crucial to balance timely and accurate predictions, enhancing the system's effectiveness in predicting safety violations. Such selection is specific to the deployment environment of the DRL agent but a threshold value around $\theta=50\%$ is likely to be balanced.  }
\end{tcolorbox}

\subsubsection{RQ4. What is the effect of the abstraction level on the safety monitoring system?}
\label{Subsec:RQ4AbsLevel}

To answer this research question, we studied how different levels of state abstraction affect the performance of the safety violation prediction model in the training phase and in operation.

\begin{figure}[ht]
    \centering
    \includegraphics[width=\columnwidth]{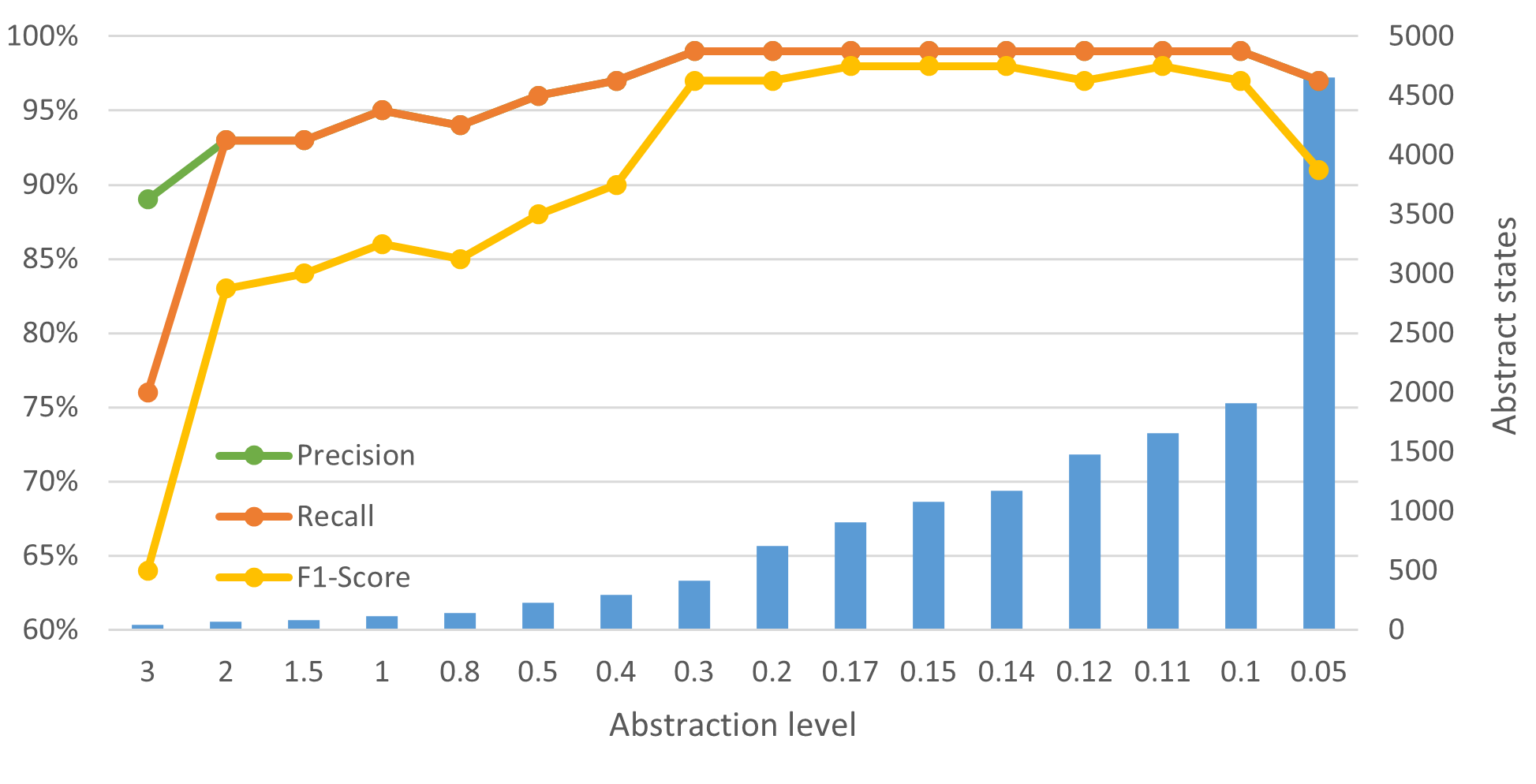}
    \caption{Precision, recall, and F1-score achieved after the training of the safety violation prediction model and the number of abstract states for the \textit{Cart-Pole} case study across different abstraction levels}
    \label{RQ3-abstraction_Cart-Pole}
\end{figure}

\begin{figure}[ht]
    \centering
    \includegraphics[width=\columnwidth]{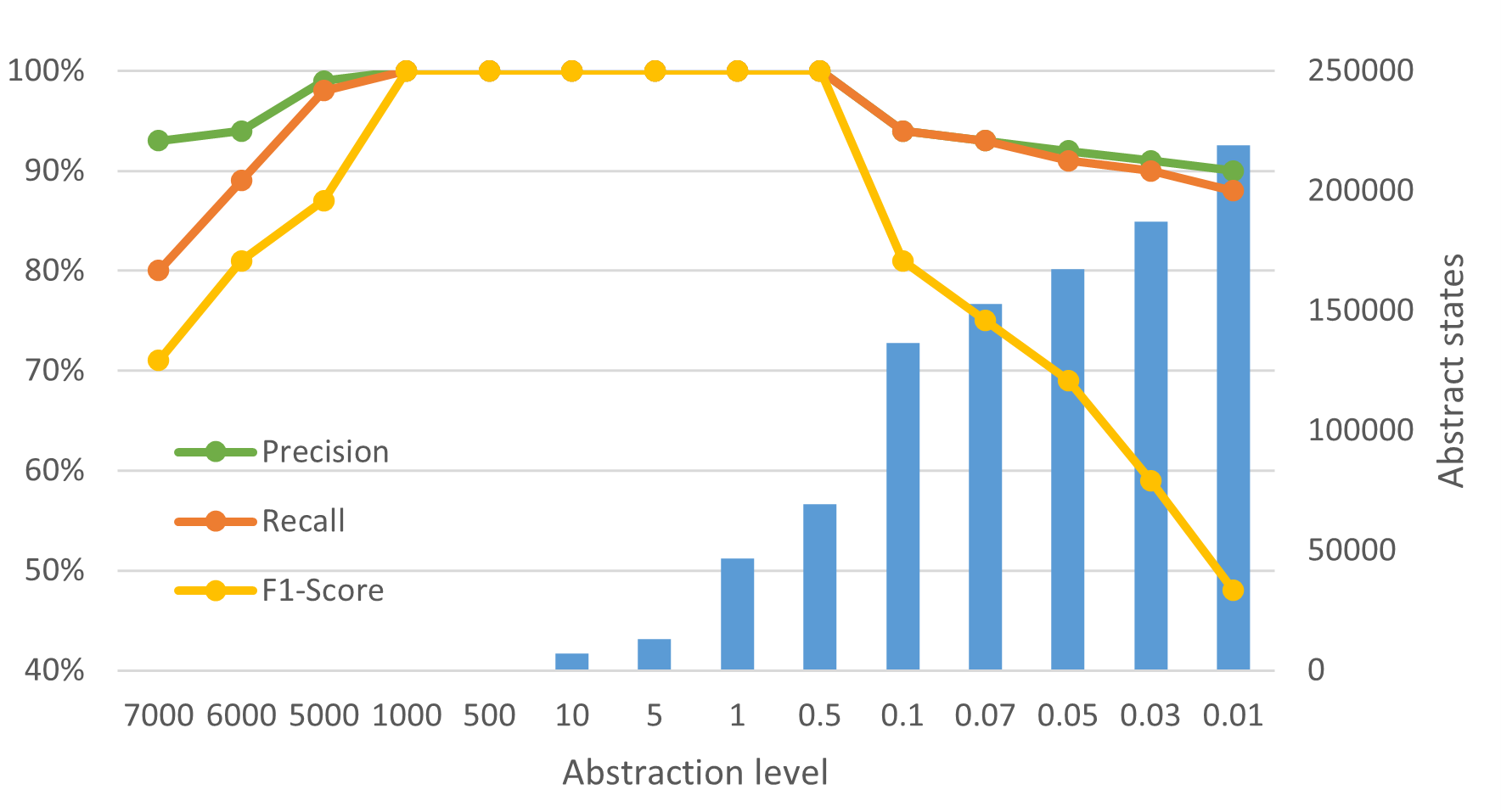}
    \caption{Precision, recall, and F1-score achieved after the training of the safety violation prediction model and the number of abstract states for the \textit{Mountain-Car} across different abstraction levels}
    \label{RQ3-abstraction_Mountian-Car}
\end{figure}

\begin{figure}[ht]
    \centering
    \includegraphics[width=\columnwidth]{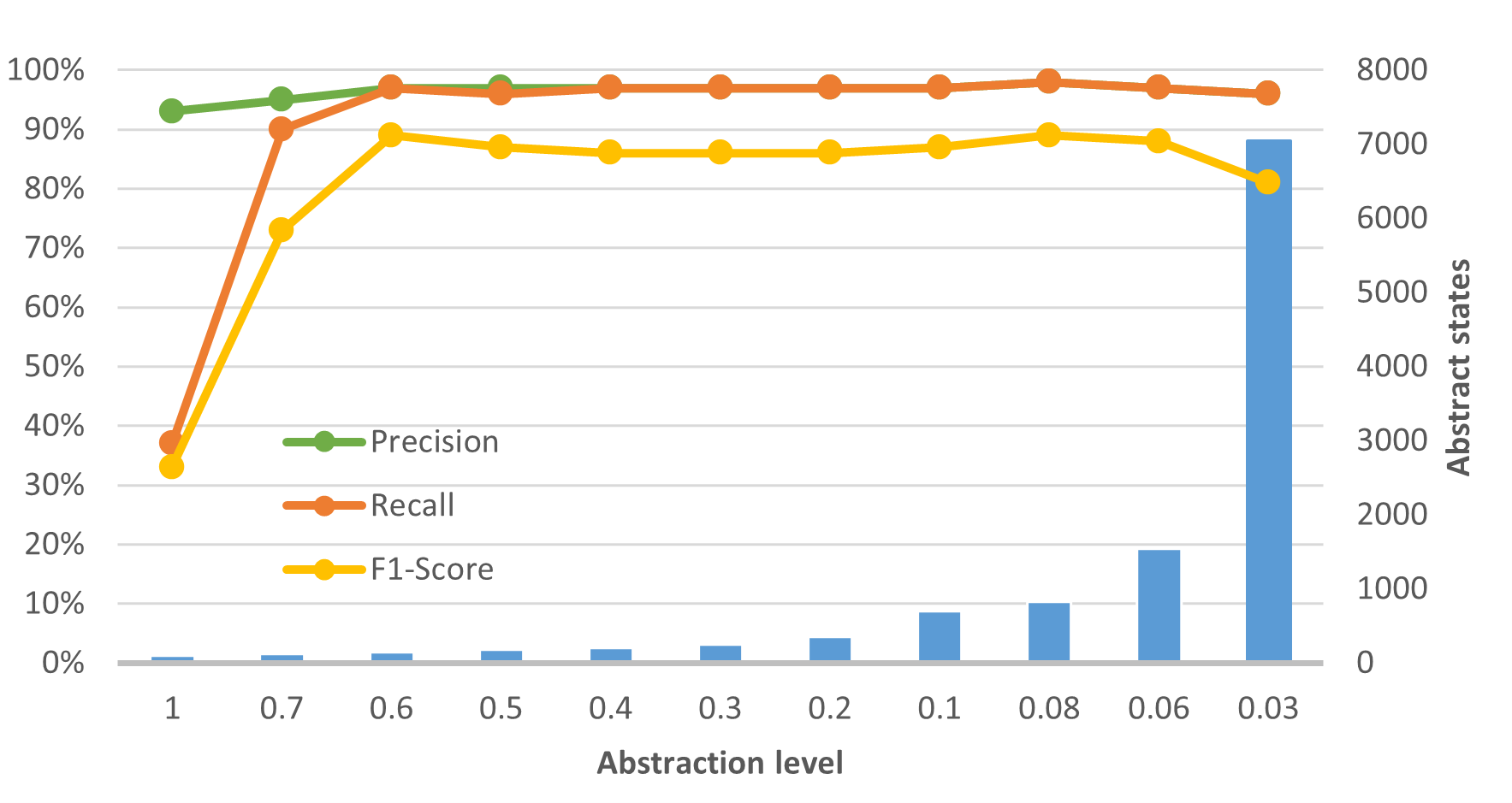}
    \caption{\added{Precision, recall, and F1-score achieved after the training of the safety violation prediction model and the number of abstract states for \textit{Highway Driving} across different abstraction levels}}
    \label{fig:RQ3-abstraction_Highway}
\end{figure}
\begin{figure}[ht]
    \centering
    \includegraphics[width=\columnwidth]{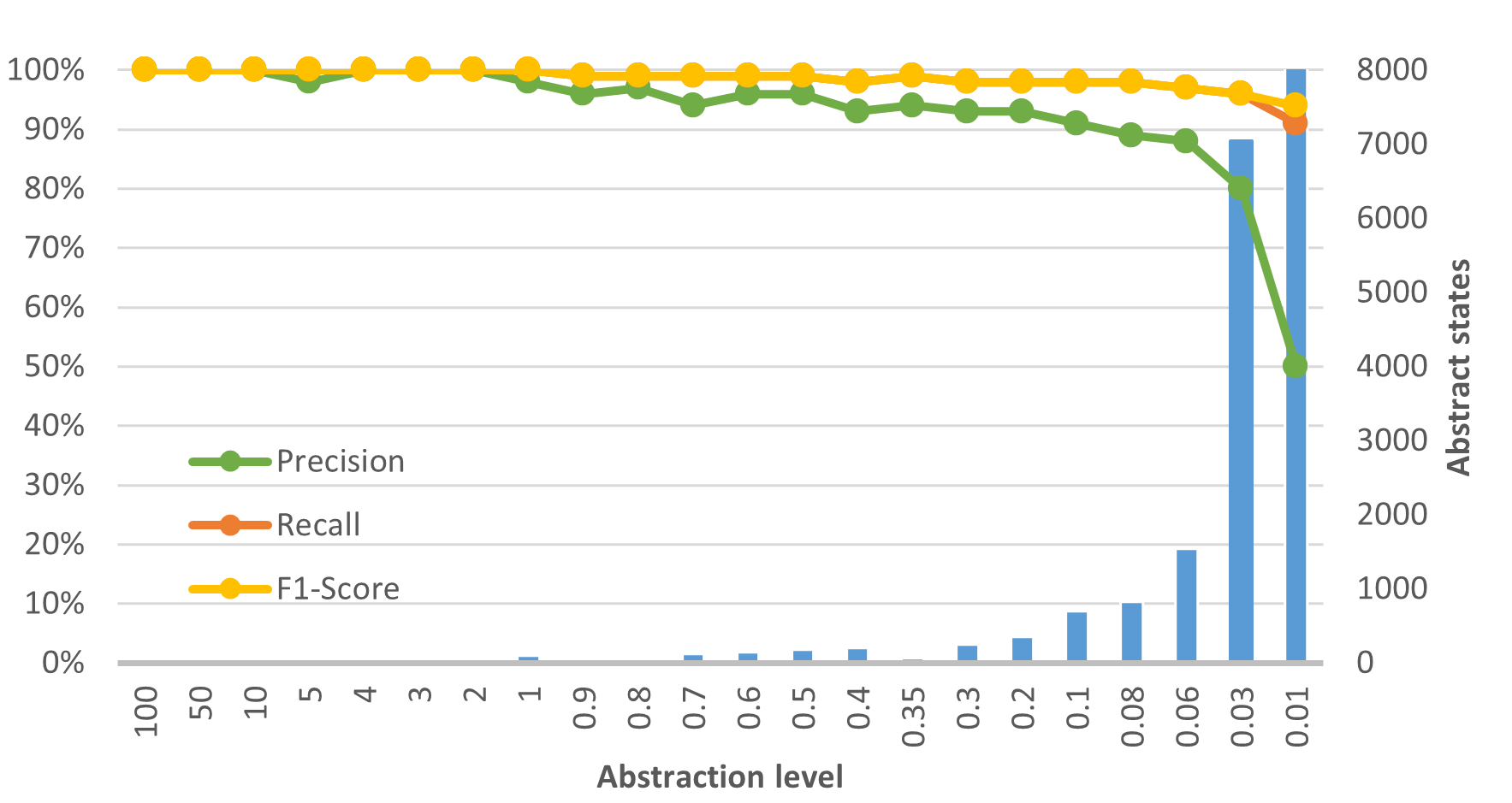}
    \caption{\minor{Precision, recall, and F1-score achieved after the training of the safety violation prediction model and the number of abstract states for \textit{Highway Driving} across different abstraction levels using frequency of abstract states as features}}
    \label{fig:RQ3-abstraction_Highway_freq}
\end{figure}

\textbf{The accuracy of the \textit{Random Forest} model after training with different abstraction levels.} This aspect involves evaluating the performance of the \textit{Random Forest} model once it has been trained on the available training data. We randomly sampled 70\% of the dataset to train and 30\% to compute the F1-scores of the models using different levels of abstraction ($d$).
A lower abstraction level implies finer-grained states, while higher abstraction levels represent coarser ones that lead to a smaller feature space.

\added{Based on the analysis depicted in Figures~\ref{RQ3-abstraction_Cart-Pole},~\ref{RQ3-abstraction_Mountian-Car}, and~\ref{fig:RQ3-abstraction_Highway}, we observed that higher abstraction levels lead to lower accuracy in predicting safety violations, above a threshold of 0.3 for \textit{Cart-Pole},  1000 for \textit{Mountain-Car} and 0.6 for the \textit{Highway Driving} case study.} This is attributed to the smaller feature space associated with higher levels of abstraction. \minor{Note, however, that this does not apply to features based on the frequency of abstract states. As the results suggest (Figure~\ref{fig:RQ3-abstraction_Highway_freq}), these features yield higher accuracy even at higher abstraction levels. This is because these features are more informative, even when the number of abstract states is small.} As the abstraction level decreases, the feature space grows larger, allowing for more precise information to be captured by features. Consequently, the accuracy of the safety violation prediction model tends to increase until it eventually plateaus and then starts to decrease. This decrease occurs due to the very large number of abstract states, making it more challenging to learn in a larger feature space. 
This suggests that there is an optimal range of abstraction that yields the highest accuracy in predicting safety violations. Going beyond this optimal range can reduce the performance of safety monitoring.

\minor{In the \textit{Highway Driving} case study using the frequency of abstract states, we observed that very high levels of abstraction ($d \geq 0.9$) achieve high accuracy but are not practical. This is because the probability of safety violations tends to consistently decrease, especially for safe episode. Specifically, at the start of the episode, the model predicts a high probability of safety violations (above the prediction threshold). As the safety monitor gathers more information, this probability drops in safe episodes, allowing the model to distinguish between safe and unsafe situations. However, this decreasing trend makes it difficult to predict safety violations and stop execution at early time steps during the execution, since the probability is above the prediction threshold. To address this, we introduced an additional criterion for choosing abstraction levels with frequency-based features, ensuring that the initial probability of safety violation remains below the prediction threshold.}

This optimal range of abstraction level depends on the environment, the RL agent, and the reward. This arises from the fact that the calculation of Q-values, which are used in the abstraction process, relies heavily on the reward signal. Consequently, the choice of reward function significantly influences the optimal range of abstraction levels.
Indeed, our empirical analysis revealed that abstraction levels ranging from 0.1 to 0.3 result in the highest accuracy for \textit{Cart-Pole}. \minor{Similarly, the highest accuracy for \textit{Mountain-Car}, \textit{Highway Driving} with binary features, and \textit{Highway Driving} with frequency features is obtained with  abstraction levels from 0.5 to 1000, 0.06 to 0.6, and 0.2 to 0.8, respectively. Therefore, for the next experiments, we consider the abstraction levels within the optimal ranges of each case study. }

\textbf{The performance of the model in operation with different abstraction levels.} This part focuses on evaluating the performance of the safety violation prediction model during the execution of episodes. We analyze how well the trained \textit{Random Forest} models perform in operation across different time steps while considering different abstraction levels.
The main focus for such evaluation is the model's ability to accurately predict safety violations early. 

\begin{figure}[ht]
    \centering
    \includegraphics[width=\columnwidth]{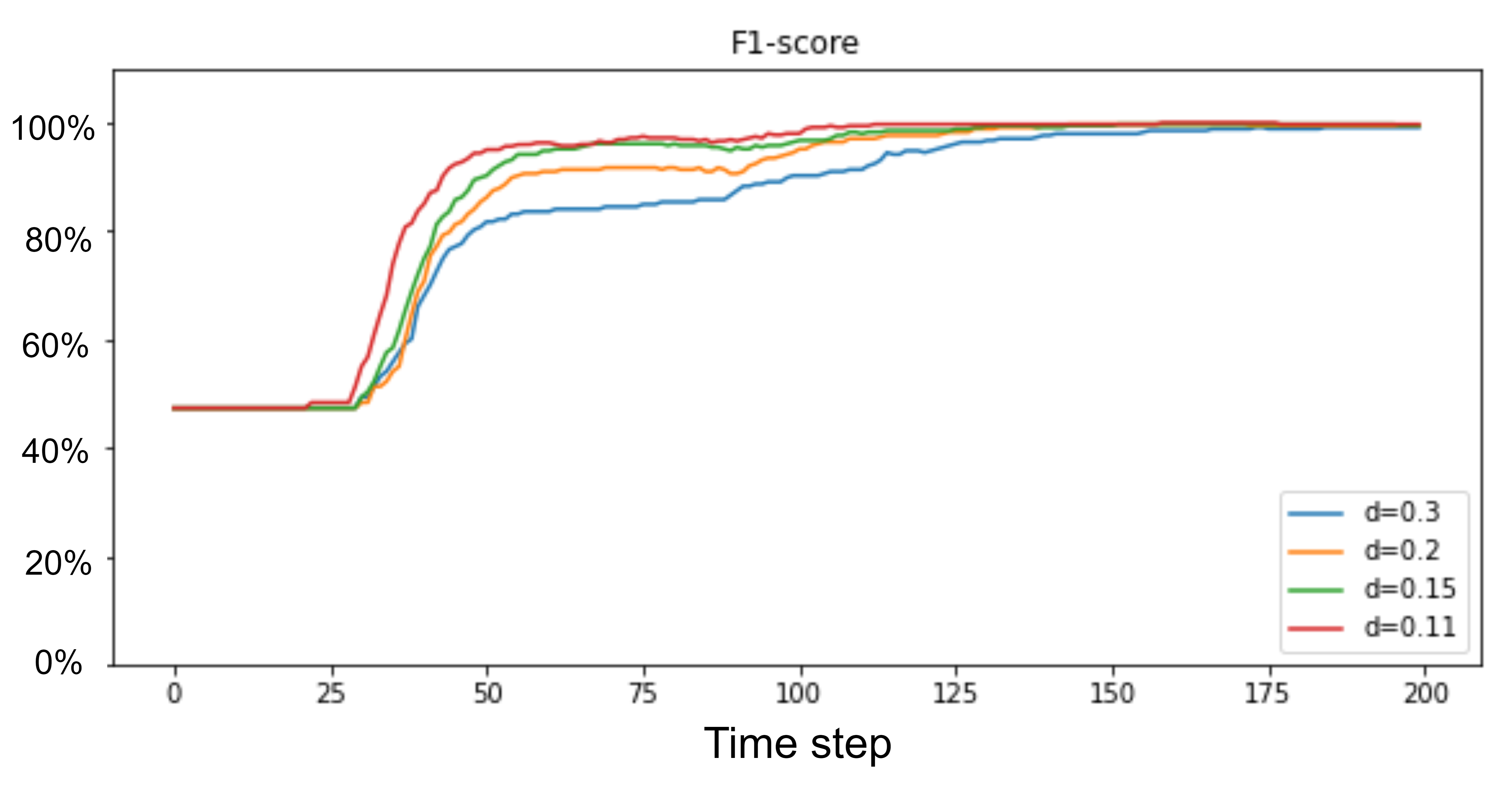}
    \caption{Performance of the safety violation prediction models in operation for the \textit{Cart-Pole} case study across different levels of abstraction}
    \label{Rq3 HD in action Cart-Pole}
\end{figure}

\begin{figure}[ht]
    \centering
    \includegraphics[width=\columnwidth]{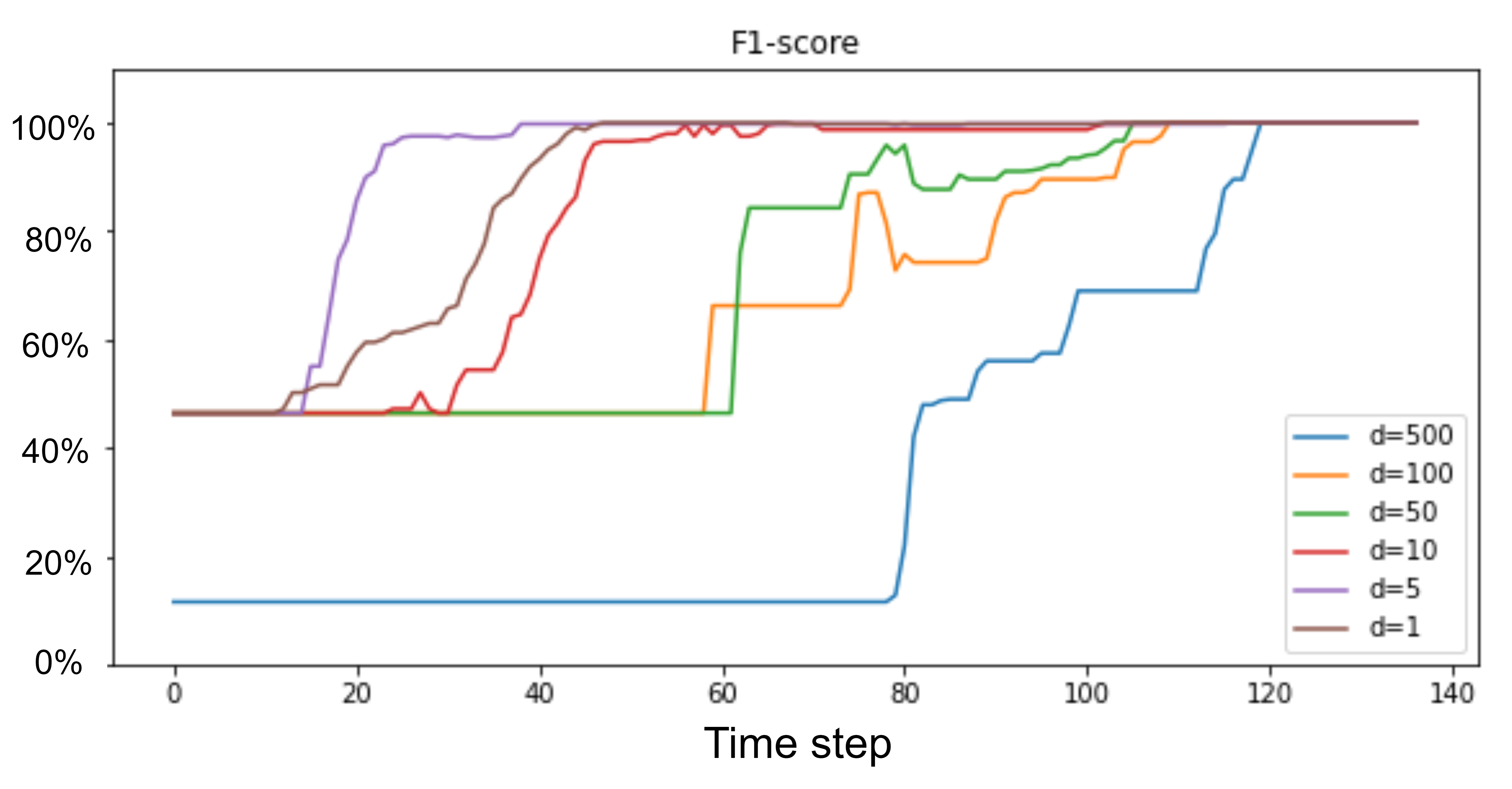}
    \caption{Performance of the safety violation prediction models in operation for the \textit{Mountain-Car} case study across different levels of abstraction}
    \label{Rq3 HD in action Mountain-Car}
\end{figure}

\begin{figure}[ht]
    \centering
    \includegraphics[width=\columnwidth]{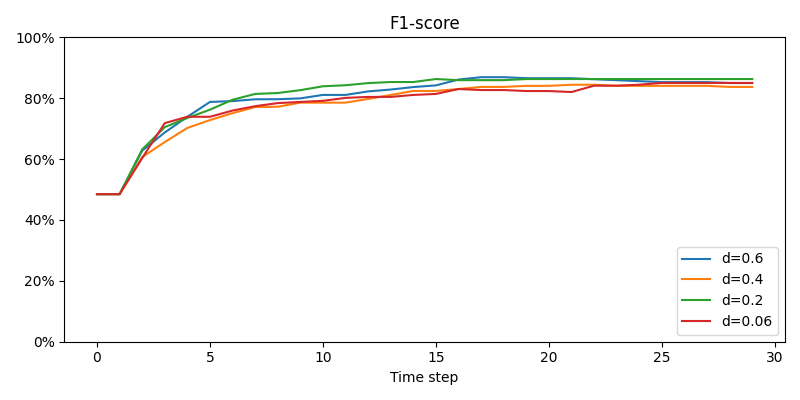}
    \caption{\added{Performance of the safety violation prediction models in operation for the \textit{Highway Driving} case study across different levels of abstraction}}
    \label{Rq3 HD in action Highway}
\end{figure}

\begin{figure}[ht]
    \centering
    \includegraphics[width=\columnwidth]{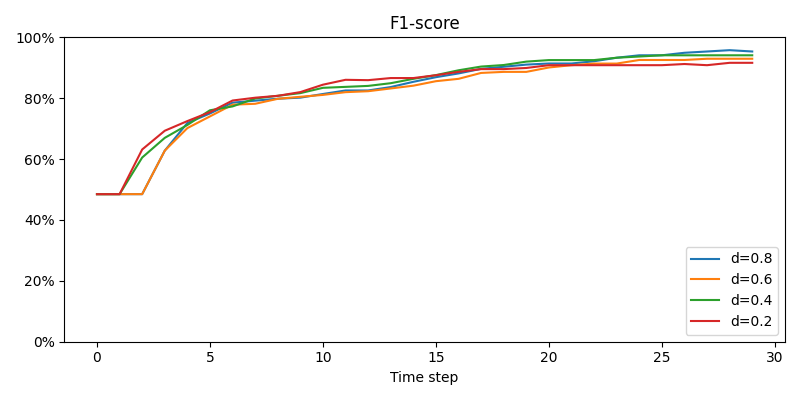}
    \caption{\minor{Performance of the safety violation prediction models in operation for the \textit{Highway Driving} case study, across different levels of abstraction, using frequency of abstract states as features}}
    \label{Rq3 HD in action Highway Freq}
\end{figure}

The F1-score of the safety monitoring models \added{across all} case studies, considering various levels of abstraction, are presented in \added{Figures~\ref{Rq3 HD in action Cart-Pole},~\ref{Rq3 HD in action Mountain-Car},~\ref{Rq3 HD in action Highway}, and~\ref{Rq3 HD in action Highway Freq}}. 
As visible, the performance of the safety violation prediction model is highly sensitive to the selected abstraction level during the training phase, especially in the case of \textit{Mountain-Car}. Despite selecting only abstraction levels that maximize the model's performance during training, they required different numbers of time steps to achieve the highest accuracy in predicting safety violations. This sensitivity highlights the importance of carefully selecting the appropriate abstraction level for optimal model performance.
\minor{Based on Figures~\ref{Rq3 HD in action Cart-Pole},~\ref{Rq3 HD in action Mountain-Car},~\ref{Rq3 HD in action Highway} and~\ref{Rq3 HD in action Highway Freq}, we observe that the most suitable abstraction level is $d=0.11$ for \textit{Cart-Pole},  $d=5$ for \textit{Mountain-Car}, $d=0.2$ for \textit{Highway Driving}, and $d=0.4$ for \textit{Highway Driving} using frequency based features, as they exhibit the most accurate and earliest prediction of safety violations compared to other abstraction levels.} 
This indicates that these abstraction levels are particularly effective at capturing relevant features at the right level of granularity to support learning and the prediction of unsafe episodes.

In summary, in this research question, we studied the performance of \textit{SMARLA} through a systematic two-step process. 
In the first step, we analyzed the accuracy of \textit{SMARLA} in predicting safety violations post-training, and we derived a range of optimal abstraction levels corresponding to the highest F1-score. Next, we investigated the number of time steps \textit{SMARLA} requires to achieve its peak accuracy. Note that, the ultimate objective is to identify the proper abstraction level which enables the safety monitor to reach its highest accuracy in predicting safety violations at the earliest time step possible.

\begin{tcolorbox}
\textbf{Answer to RQ4:} 
The accuracy of safety violation prediction models is sensitive to the selected abstraction level and, therefore, the latter should be carefully selected to have optimal monitoring results.
\end{tcolorbox}

\subsubsection{Abstraction level selection procedure}
\label{Subsec: abstraction level selection procedure}

To achieve early and accurate predictions of the safety monitor in practice, it is crucial to determine a proper abstraction level.
Thus, based on the results of RQ4~\ref{Subsec:RQ4AbsLevel}, we propose the following procedure to select a proper abstraction level for our monitoring approach. This procedure is a one-time process that involves a systematic two-step approach. Initially, in the first step, a Coarse-to-Fine search technique~\cite{schaeffer2008coarse} is used for finding the optimal range of abstraction levels that maximize the F1-score of the safety violation prediction model. Subsequently, in the second step, within this optimal range, the abstraction level yielding the earliest prediction of safety violation is identified.

In detail, this process starts by training the safety violation prediction models with approximately 70\% of the training episodes. During this phase, the safety violation prediction model is trained using a diverse range of abstraction levels, systematically varied in increments. The assessment metric is the F1-score of the safety violation prediction model. It is noteworthy that abstraction levels can be mapped to the number of abstract states, thereby facilitating the determination of the range of abstraction levels to be explored. As a practical guideline, using the Coarse-to-Fine approach, it is recommended to start with a wide range spanning from several hundred to approximately 100,000 states. The process involves iteratively considering new values close to abstraction levels that previously yielded high F1-scores, and therefore gradually narrowing down the range to ultimately identify the optimal range of abstraction levels.

However, the choice of the range of levels to cover depends also on the complexity of the environment; more complex environments may necessitate a larger number of abstract states to ensure accurate prediction of safety violations.
Subsequently, within the recommended range, the optimal subset of abstraction levels can be determined by selecting levels that maximize the F1-score of the models when evaluated on the remaining 30\% of episodes. \minor{The results of this step for  \textit{Cart-Pole}, \textit{Mountain-Car}, \textit{Highway Driving}, as well as  \textit{Highway Driving} with frequency-based features are depicted in Figures~\ref{RQ3-abstraction_Cart-Pole},~\ref{RQ3-abstraction_Mountian-Car},~\ref{fig:RQ3-abstraction_Highway} and~\ref{fig:RQ3-abstraction_Highway_freq}, respectively. 
In the \textit{Cart-Pole}, the optimal range of abstraction levels lies between 0.1 and 0.3, whereas for the \textit{Mountain-Car}, it is between 0.5 and 1000, and for \textit{Highway Driving} case study, it is between 0.6 and 0.06.}

In the second phase, it is imperative to conduct experiments with different abstraction levels within the optimal range when utilizing the safety violation prediction model during the execution of the episode. The primary objective is to pinpoint the level of abstraction that enables the safety monitor to make precise predictions of safety violations at the earliest time step possible. 
\added{Figures~\ref{Rq3 HD in action Cart-Pole},~\ref{Rq3 HD in action Mountain-Car} and~\ref{Rq3 HD in action Highway} illustrate the outcomes of the second phase, where we extracted the F1-score of safety monitoring models considering various abstraction levels within the optimal range. As a result, the optimal abstraction levels that enable accurate and early predictions of the safety monitoring approach for the \textit{Cart-Pole}, \textit{Mountain-Car} and \textit{Highway Driving} case studies are $d=0.11$, $d=5$ and $d=0.2$, respectively.}

\subsubsection{Qualitative Analysis}
\label{Sec:RQA}

\begin{figure}[t]
    \centering
    \includegraphics[width=\linewidth]{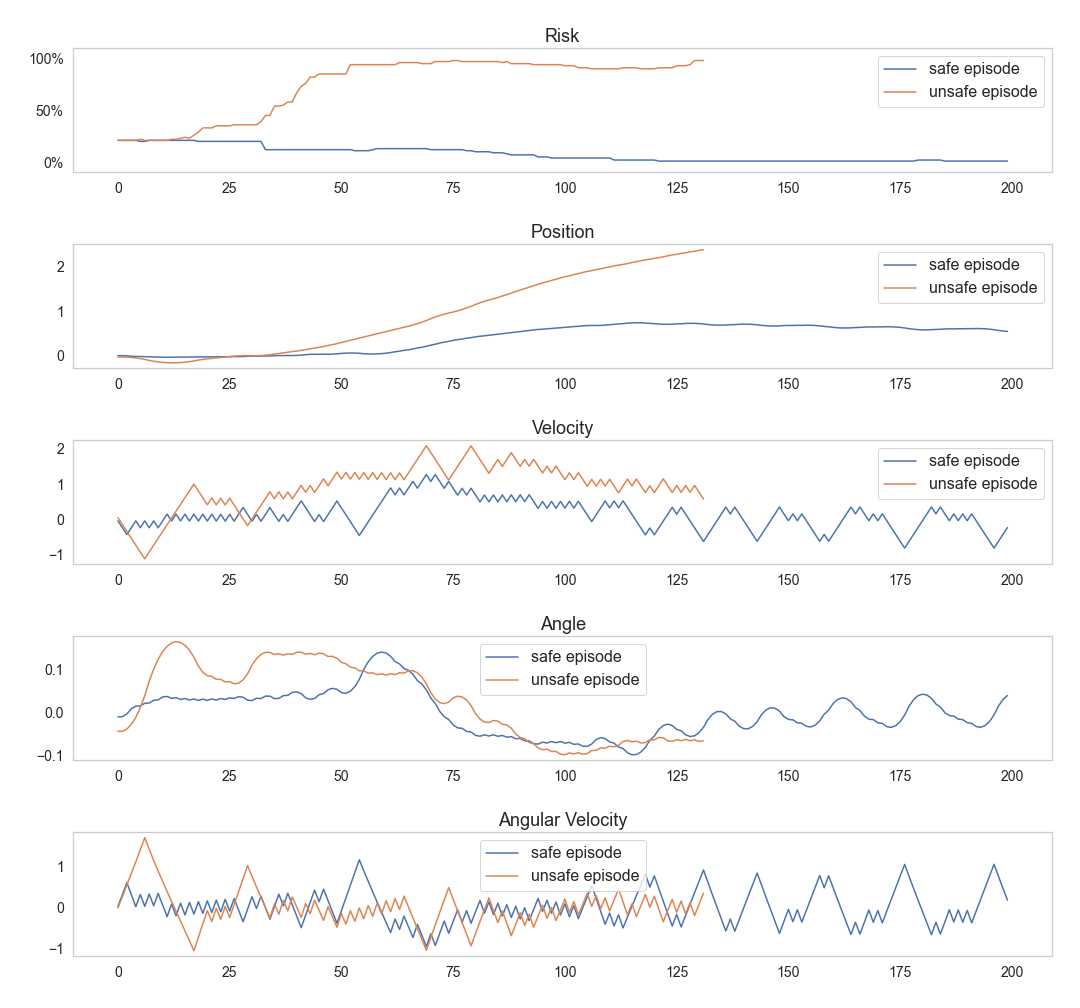}
    \caption{\added{Examples for \textit{Cart-Pole}}}
    \label{fig:QA_CP}

\end{figure}

\added{To further shed light on the prediction results of \textit{SMARLA}, we manually sampled representative episodes from each case study and qualitatively analyzed the corresponding risk plots. The primary objective of this analysis was to understand how the probability of safety violations evolves in relation to the dynamics of the agent in each scenario, providing deeper insights into the behavior of the agent and the conditions under which the risk of safety violation is increased.
For example, in the \textit{Cart-Pole} case study, we analyzed how the probability of safety violations evolves along with the position and velocity of the cart, as well as the angle and angular velocity of the pole. We provide an example of the safety violation prediction plot for a safe and an unsafe episode in Figure~\ref{fig:QA_CP}. This Figure illustrates how the probability of safety violation fluctuates with changes in the cart's dynamics. It is important to recall that an episode is considered unsafe if the cart moves away from the center by a distance above 2.4 units as it crosses the safe operational boundary and can cause damages, regardless of the accumulated reward (detailed in Section~\ref{subsec:Cartpole}). 
As shown in Figure~\ref{fig:QA_CP}, the risk plot shows that in the safe episode, the probability of safety violations remains relatively low throughout the episode as the cart stays within the safe boundary. In contrast, in the unsafe episode, the probability of safety violations increases significantly as the cart approaches the safety boundary limit of 2.4. 
The velocity plot shows that in the safe episode, the velocity of the cart fluctuates within a moderate range, while in the unsafe episode, the cart's velocity reaches higher magnitudes (both positive and negative), indicating more pronounced movements. The angle and angular velocity plots further illustrate that in the unsafe episode, the angle of the pole deviates more significantly, and the angular velocity shows higher variations compared to the safe episode. More specifically the probability of safety violation increases when the angle of the pole is leaning toward the right and the cart is moving to the right to recover the pole from falling (time steps 40 to 50). This is clearly visible when comparing with another safe episode plotted in the same figure where, despite the pole starting to lean toward the right (time step 60), the agent immediately moves the cart to the right and successfully prevents the pole from falling in a much shorter distance from the reference point, so the probability of safety violations is decreased as depicted in Figure~\ref{fig:QA_CP}.}

\begin{figure}[t]
    \centering
    \includegraphics[width=\linewidth]{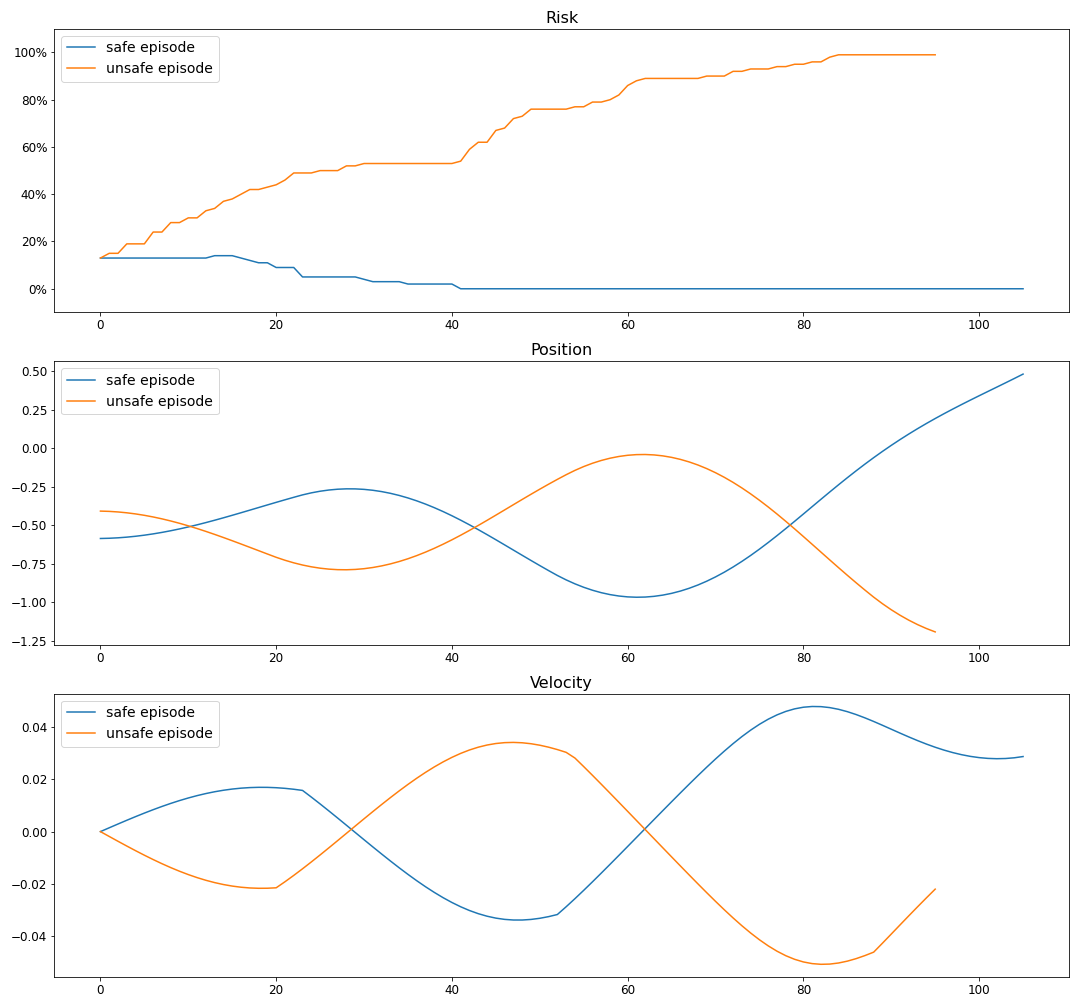}
    \caption{\added{Examples for \textit{Mountain-Car}}}
    \label{Fig:QA_MC}
\end{figure}

\added{Regarding the \textit{Mountain-Car} case study, we also analyzed the probability of safety violations along with the position and velocity of the car. Examples of safe and an unsafe episodes is shown in Figure~\ref{Fig:QA_MC}. We should recall that a safety violation happens when the car crosses the left border of the environment, as this poses potential damage to the car (detailed in section~\ref{subsec:MTC}).}
\added{In the unsafe episode, the agent initially gains momentum in the opposite direction compared to the safe episodes. The probability of safety violation significantly increases when the car is on the right slope, gaining momentum toward the left border. The momentum becomes so high that the car cannot stop before passing the left border. As the car gets closer to this boundary, the probability of safety violation continues to rise, where finally the car passes the left border and a safety violation occurs.}

\begin{figure*}[ht]
    \centering
    \includegraphics[width=\linewidth]{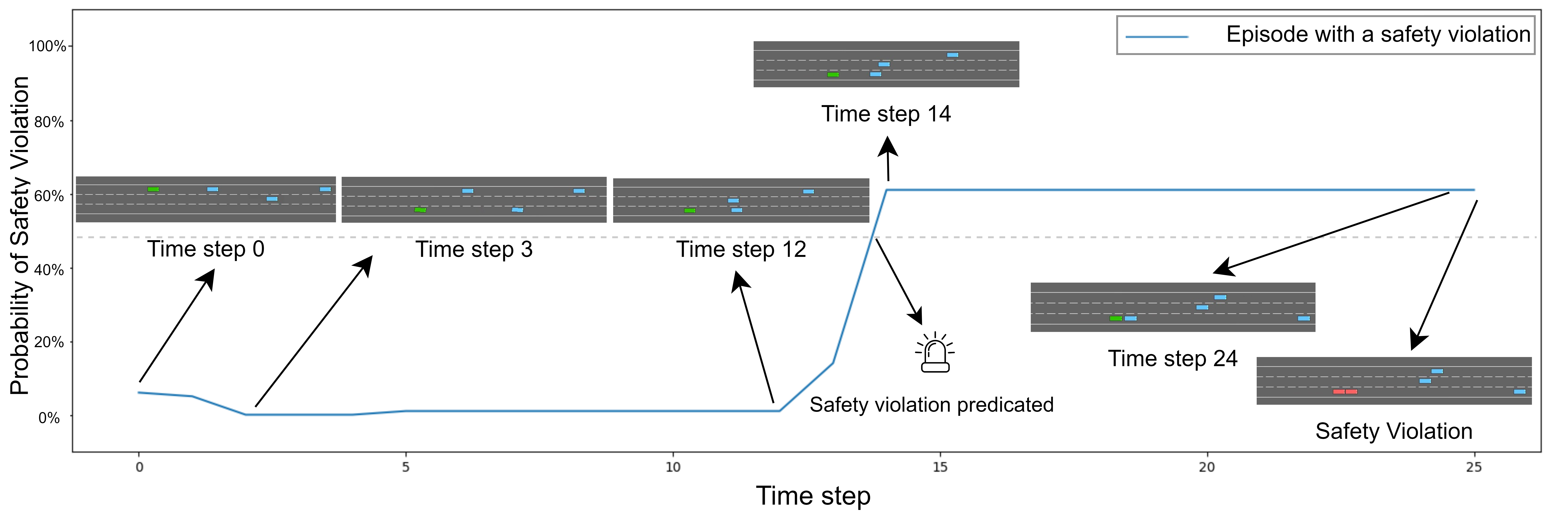}
    \caption{\added{Examples for \textit{Highway Driving}}}
    \label{Fig:QA_HW}
\end{figure*}

\added{For the third case study, we analyzed the probability of safety violations in the \textit{Highway Driving} scenario where the agent must navigate at high speeds while avoiding other vehicles. We should recall that, in this environment, a safety violation occurs when the agent collides with another vehicle (Section~\ref{Sec:CaseHighway}).}

\added{Figure~\ref{Fig:QA_HW} illustrates the probability of safety violation in an unsafe episode along with snapshots of the environment captured during the execution. In this episode, the agent starts driving in the third lane, where there is another actor vehicle at a safe distance from the ego vehicle. The agent decides to move to the rightmost lane to gain reward and increases speed to safely overtake the other vehicles (time steps 0 to 3). Then, another vehicle appears in the rightmost lane, cruising at a lower speed than the ego vehicle. At time step 12, when the ego vehicle is approaching the vehicle in front at high speed, the probability of safety violation begins to rise. At time step 14, the probability exceeds the 50\% threshold, and we predict the safety violation as the ego vehicle continues to approach the vehicle in front. The execution continues until the final time step, where a collision occurs between the two vehicles. }

\added{This qualitative analysis illustrates how \textit{SMARLA} effectively predicts potential safety violations for DRL agents. By examining the agent's dynamics and the probability of safety violations, we provide insights into \textit{SMARLA}'s capability to provide accurate early predictions of safety violations during the execution of episodes. Such early detection allows the system to take proactive measures to prevent or mitigate potential damages, thereby enhancing the overall safety and reliability of DRL agents.}

\added{Additionally, it demonstrates the practical application of SMARLA and qualitatively validates its prediction results. Through such a qualitative analysis, one can understand the conditions under which the risk of safety violation is increased, and thus provide insights into how these risks might be mitigated in operation.}

\section{Discussion}
\label{Sec:Discussion}
\minor{We propose \textit{SMARLA}, a black-box safety monitoring approach designed for reinforcement learning agents, and demonstrate its high accuracy in predicting safety violations of RL agents during their execution. We also show that such accurate prediction can be made early long before the actual occurrence of violations, allowing for timely incident prevention and mitigation. 
In our work, we rely on state abstraction to reduce state space and learn change patterns in agent Q-values to predict unsafe episodes. We should emphasize that learning the  patterns of unsafe episodes with abstraction does not imply that all episodes following such patterns represent known unsafe concrete states. State abstraction allows us not to depend on the specific occurrence of concrete states and thus generalize patterns to concrete states that may not have been seen during training. Although no approach, including ours, can claim to fully predict or mitigate all unknown unsafe states, the use of state abstraction enhances the safety monitor's capability to recognize potential risks by leveraging similarities in abstract states and learned Q-values from past patterns.}  

\minor{In addition, predicting known unsafe states early is not just valuable, it is essential to achieving acceptable safety levels as recommended by several safety standards for different safety critical domains such as automotive and medical sectors. In these domains, effective risk management depends on the system’s ability to detect and address known unsafe states early in operation. For example, in autonomous driving, Tesla's Autopilot must manage a variety of known unsafe states, such as vehicle skidding, excessive braking distances, or unprotected left turns. Even though these conditions are known safety violations, their occurrence depends on dynamic environmental variables. If an autonomous vehicle detects early signs of hydroplaning (a state where the tires lose contact with the road due to water), it can engage preventive measures, such as reducing speed or adjusting the trajectory, long before the vehicle becomes uncontrollable. A delayed response could lead to accidents that might have been prevented with earlier detection. In such scenarios, knowing the unsafe state is not sufficient as early detection of safety violations becomes crucial.
In this context, state abstraction techniques, like the $Q^*$-irrelevance abstraction employed in our approach, allow for generalization over known patterns, enabling the system to detect potentially unsafe behavior that may not match previous episodes exactly but follows recognizable safety violation patterns. This provides a level of generalization over possible unsafe episodes and helps the system adapt to new but related situations. }

\minor{In short, we offer a light-weight solution to build a safety monitor based on testing results of DRL agents. This is important in practice as testing data contain useful information that can be exploited for monitoring. We should note that the more comprehensive the testing, the better the safety monitoring, as we cannot learn from what we have not observed. However, we alleviate this limitation with state abstraction, although this challenge cannot be fully eliminated.}

\section{Threats to Validity}
\label{Sec:Threats}

In this section, we discuss the different threats to the validity of our study and describe how we mitigated them. 

\noindent{\textbf{Internal threats}} concern the causal relationship between the treatment and the outcome. One such threat is the choice of an inappropriate state abstraction level. To mitigate this threat, we explored different abstraction levels and identified the ones that  (1) significantly reduce the state space, (2) optimize the accuracy of the ML model, and (3) enable the earliest and most accurate predictions of safety violations during the RL agent execution. Another threat pertains to newly seen abstract states during operation (i.e., abstract states that were not included in the training dataset of the safety violation prediction model), which may impact the accuracy of the safety monitoring system. To mitigate the risk of missing abstract states, we trained our ML model with a large and diverse dataset of RL episodes and use a state abstraction method that considers a large number of concrete states (roughly 450 000 in \textit{Cart-Pole} and 250 000 in \textit{Mountain-Car}). We will study, in future work, the impact of retraining the ML model (to include newly seen abstract states) on the performance of our safety monitoring approach. The scarcity of unsafe episodes poses a threat to accurate safety violation model training. As future work, we will explore extracting unsafe episodes from the later stages of the RL agent's training phase, as they align more closely with the agent's final policy.

\noindent{\textbf{Conclusion threats}} are concerned with the relationship between treatment and outcome. One important threat to consider is the potential impact of a large monitor's overhead on its applicability in real-time scenarios.
Such overhead refers to the additional computational resources and time required by safety monitoring. To mitigate this threat, \textit{SMARLA} relies on state abstraction to reduce the state space and computation overhead. It also uses \textit{Random Forest}, which is a lightweight machine learning model known for its fast inference capabilities~\cite{10.1023/A:1010933404324,2023}. This choice helps minimize the computational burden associated with monitoring, allowing for fast processing and response times. Note that, in our case studies, inference times (i.e., safety violation predictions) are in the range of a few milliseconds, thus confirming our expectations. Finally, our monitoring approach is black-box. It does not require access to the internals of the RL agent, which greatly simplifies processing and reduces computation overhead.

\noindent{\textbf{Reliability threats}} concern the replicability of our study results. We rely on publicly available ML models and RL environments. We provide online all the materials required to replicate our study results.

\noindent{\textbf{External threats}} concern the generalizability of our study. Due in part to the high computational expense of our experiments, we relied on \added{three} case studies in our work, which may threaten the generalizability of our results. Indeed, experiments for all research questions took about four weeks using a system with a Core-i9 processor, 32GB of memory, and an Nvidia GeForce RTX 3070 GPU with 8GB of memory. 
However, to mitigate this threat, we relied on benchmark problems widely used in many RL-related studies~\cite{Behzadan2019AdversarialEO,Behzadan2019SequentialTF,DBLP:journals/corr/abs-2006-05032,pattanaik2018robust,yan2024forgingvisionfoundationmodels,10565991,10.1007/978-3-031-48121-5_18,10077125}. Future work includes applying our monitoring approach to different RL problems for broader generalization of the results. 
\added{Another threat pertains to the ability of \textit{SMARLA} to detect unsafe episodes that are not included in the training set. To mitigate this threat, we use state abstraction and the agent's Q-values to predict safety violations. In our approach, we assume that patterns of Q-value changes observed in training episodes with safety violations can also occur under operational conditions. This allows \textit{SMARLA} not to depend on the specific occurrence of concrete states and extend its predictive capability beyond the concrete states seen during training. This is performed by recognizing and responding to analogous change patterns even in the presence of unseen states, thus generalizing patterns to concrete states that have possibly not been encountered. Furthermore, it is worth noting that we have explained strategies to deal with unseen abstract states in the deployment environment in Section~\ref{Sec:FunctionalFault}. }

\section{Related Work} 
\label{Sec:RW}

This section provides an overview of the most relevant literature on safety monitors. We discuss related works on (1) safety monitors for RL agents, (2) safety monitors for AI/ML-based systems in general \added{, and (3) safety monitoring for Cyber-Physical Systems (CPS). The first topic is obviously directly related to our work, while topic (2) is a bit more general and topic (3) is indirectly related as RL agents are often deployed on Cyber-Physical Systems.} 

\subsection{Safety monitors for RL} Several approaches have been proposed in the literature on safe reinforcement learning~\cite{mindom2021assessing,gu2022review,garcia2015comprehensive}, but only a few of them proposed safety monitors to predict safety violations of the agent at runtime. 
Junges~\textit{et al.}~\cite{junges2021runtime} proposed two strategies for monitoring MDPs: forward filtering-based monitoring and model checking-based monitoring. The former involves estimating future possible states of the system based on historical observations. 
In contrast, the latter uses model-checking techniques to assess the probability of reaching certain states in the MDP. 
These strategies are only applicable when a detailed model of the environment is available, in contrast to \textit{SMARLA} which works on a broader type of model-free RL algorithms which are more widely used in practice as environment models are challenging to develop and validate~\cite{wang2020deep,semeraro2023human}.

Other approaches have been proposed for the online safety shielding of RL agents where a shield layer is added to the agent and blocks unsafe actions during the execution~\cite{konighofer2022online,melcer2022shield}. In general, online shielding operates by dynamically calculating, during runtime, the collection of potential states that may be reached in the immediate future. Within this set of states, the safety of all feasible actions is assessed and leveraged for shielding purposes as soon as any of the aforementioned states is encountered. In that context, Konighofer~\textit{et al.}~\cite{konighofer2022online} proposed an online shielding technique for RL agents that employs MDPs and probabilistic model-checking to evaluate all possible actions in the agent's states. By blocking unsafe actions and analyzing the impact of shielding on learning performance and policy safety, they demonstrate improved learning performance with shielding and recommend its application during both the training and execution phases of RL agents.
Other shielding approaches have been proposed to increase the safety of RL agents during the training phase~\cite{mallozzi2019runtime,elsayed2021safe}. For example, Mallozzi~\textit{et al.}~\cite{mallozzi2019runtime} introduced \textit{WiseML}, a runtime safety monitoring framework for RL agents. The goal of this framework is to prevent the agent from selecting unsafe actions during training by relying on manual specifications of safety violation patterns using linear-time temporal logic. The monitoring system blocks safety violations, penalizes the agent's reward for unsafe actions, and enhances the learning performance of the agent by accelerating the convergence to its goal. Also, Elsayed~\textit{et al.}~\cite{elsayed2021safe} proposed a shielding approach for multi-agent reinforcement learning. Their goal is to enforce safety specifications expressed in linear temporal logic and learn safe RL policies in multi-agent environments.

Our research objectives differ from shielding-based studies as we propose a black-box safety monitor for RL agents that does not force the agent to select non-optimal actions (i.e., actions that deviate from the optimal policy) and continue the execution relying on the policy of the agent. Rather, our focus is on the early prediction of safety violations to provide engineers with the flexibility to implement a range of safety mechanisms, including the option to apply early corrective or preventive safety measures.

\subsection{Safety monitors for AI/ML} Different approaches have been proposed in the literature to monitor the safety of ML-based systems due to concerns about their reliability, especially in safety-critical systems~\cite{cheng2019runtime,henzinger2019outside,aslansefat2020safeml,ferreira2021benchmarking,stocco2020misbehaviour}. Some studies focus on detecting distributional shift (between training inputs and data seen in operation time)~\cite{cheng2019runtime,henzinger2019outside,aslansefat2020safeml,ferreira2021benchmarking} resulting in DNN mispredictions during operation~\cite{stocco2020misbehaviour,stocco2022thirdeye,hussain2022deepguard}, aiming to maintain accuracy and reduce safety risks. These approaches compare neurons activation patterns~\cite{cheng2019runtime,henzinger2019outside} or features distributions~\cite{aslansefat2020safeml} of inputs during operation with those seen during training. Other approaches involve monitoring DNN outputs and verifying their confidence levels to predict DNNs mispredictions~\cite{weiss2021fail}. \added{Other studies such as \textit{Likelihood Regret}~\cite{xiao2020likelihood} detect out-of-distribution DNN input images by reconstructing inputs using autoencoders. However, unlike \textit{SMARLA}, this method does not support the early prediction of safety violations for DRL agents.} 
Furthermore, Stocco~\textit{et al.}~\cite{stocco2020misbehaviour} introduced \textit{SafeOracle}, an unsupervised monitoring tool for autonomous driving systems. \textit{SafeOracle} aims to predict a DNN's misbehavior by building a proxy for the DNN's output confidence level at runtime. 
\added{The authors have proposed an autoencoder-based image reconstruction technique to forecast the subsequent input images and predict out-of-distribution inputs through a reconstruction error.} Additionally, they have trained an anomaly detector on the nominal image inputs of the DNN, which observes the reconstruction error to identify any anomaly.
In their approach, they combine confidence estimation, probability distribution fitting, and time series analysis, and show that they were able to identify 77\% of safety violations up to six seconds in advance. \added{The same applies for \textit{DeepGuard}~\cite{hussain2022deepguard} which relies on measuring the reconstruction error of input images using autoencoders to predict safety violations. \textit{SelfOracle} and \textit{DeepGuard} are designed for DNN systems primarily focused on classification or regression, where the input spaces consist of images. These methods require image transformations to effectively train the autoencoders and employ distance metrics specific to images to measure reconstruction error and estimate the probability of safety violations.
Our approach, however, is fundamentally different in terms of inputs and methodology. Further, in our case studies, the DRL agent and the underlying DNN do not take images as inputs. Instead, they process state parameters such as the position and velocity of the agent. This distinction is crucial as the nature of the input data in our approach differs significantly from that in traditional image-based DNN safety monitors. In fact, our approach is agnostic to the type of DRL agent’s inputs since we only rely on the agent’s Q-values. Furthermore, we monitor the behavior of the agents to predict violations based on a training dataset that contains both safe and unsafe episodes, as opposed to relying solely on out-of-distribution inputs.}

\subsection{Safety monitoring for CPS}

\added{Safety monitoring of CPS is crucial for ensuring their reliable and secure operation, particularly when integrating deep learning models. Various approaches have been proposed to address the safety monitoring of CPS~\cite{xie2023mosaic,henzinger2020outside,bartocci2018specification,strickland2018deep,michelmore2020uncertainty}.
For example, Xie \textit{et al.}~\cite{xie2023mosaic} proposed \textit{Mosaic}, a model-based safety analysis framework specifically designed for AI-enabled CPSs. This framework addresses the safety challenges that arise from the integration of AI-based controllers in CPSs, which introduce uncertainties and potential safety risks due to their random exploration nature and lack of systematic explanations for their behavior.
\textit{Mosaic} constructs an abstract MDP model to represent the AI-CPS, enabling safety analysis through two main components: online safety monitoring and offline model-guided falsification. 
Online safety monitoring employs probabilistic model checking to predict safety issues during real-time operations and switches control to a predefined safety controller if necessary. Offline model-guided falsification searches for counterexamples that violate safety specifications using a combination of global and local search strategies.
Our method, \textit{SMARLA}, differs from \textit{Mosaic} in several aspects. \textit{Mosaic} is model-dependent, requiring the construction of an MDP model. This process can be complex and computationally expensive, especially for high-dimensional systems. In contrast, \textit{SMARLA} is model-free, eliminating the need for constructing abstract models, thereby simplifying the safety analysis process and reducing computational overhead. Furthermore, \textit{Mosaic} focuses on general AI-CPS applications but does not specifically address RL environments. \textit{SMARLA}, on the other hand, is specifically designed for RL environments, providing tailored safety solutions for RL applications. 
In summary, while \textit{Mosaic} offers a framework for the safety analysis of AI-CPS through model-based methods, \textit{SMARLA} addresses key limitations by being model-free, specifically designed for RL environments, and suitable for black-box RL agents. This provides a more versatile and efficient solution for ensuring safety in reinforcement learning applications.
}
\added{Henzinger \textit{et al}.~\cite{henzinger2020outside} proposed an abstraction-based monitoring framework for neural networks, focusing on novelty detection during runtime. Their method utilizes box abstractions to monitor the behavior of neural networks by observing hidden layers. Specifically, they define intervals (or boxes) around the values seen during training for each neuron in the monitored layers. If an input causes the network to produce values outside these pre-defined boxes, it is flagged as a novelty. 
This method balances false positives and novel input detection through adjustable parameters. However, it requires access to the internal structure of the neural network, making it a white-box approach. This limitation restricts its applicability to scenarios where such access is available. In contrast, \textit{SMARLA} is black-box and does not require knowledge of the model’s internals, making it more versatile. While the proposed framework is primarily designed for classification tasks, \textit{SMARLA} is tailored to reinforcement learning environments.}
\added{Furthermore, Strickland \textit{et al.}~\cite{strickland2018deep} proposed an LSTM-based safety monitoring approach for CPS. The model analyses the driving environment images as well as the vehicle's state to predict potential collisions.  However, this method is tailored to image inputs, whereas \textit{SMARLA} is versatile and not restricted to any specific type of input data. Michelmore \textit{et al.}~\cite{michelmore2020uncertainty,michelmore2018evaluating} relied on the evaluation of uncertainty measures, in real-time within end-to-end controllers, and Bayesian inference methods for autonomous driving to predict safety violations. However, these approaches require access to the internal weights of the DNN models, and are thus white-box.}

\added{Several other approaches have been proposed in the literature on the runtime verification of CPS to check the compliance of such systems with safety requirements. For example, Grundt \textit{et al.}~\cite{grundt2022towards} presented a formalization approach for spatio-temporal requirements in autonomous driving systems and proposed a runtime verification method to ensure these systems comply with complex requirements during validation runs. Their monitoring system continuously evaluates the behavior of the autonomous driving system to provide real-time compliance verdicts with the system's requirements. The same applies to Zapridou \textit{et al.}~\cite{zapridou2020runtime} where the authors propose a runtime verification approach to ensure that autonomous driving systems, using an adaptive cruise control system as a case study, operate safely and effectively within simulated urban driving environments. This is achieved by continuously monitoring the system's behavior against predefined safety and performance specifications expressed in signal temporal logic. 
The objectives of the above approaches substantially differ from our work because they concentrate on the runtime verification of autonomous driving systems. Unlike \textit{SMARLA}, which predicts the probability of safety violations at each time step during the operation of RL agents, these methodologies focus on verifying system compliance with predefined requirements at runtime. Therefore, we do not include these approaches as a baseline for comparison.}

Overall, \textit{SMARLA} stands out by providing a novel black-box safety monitoring approach that is tailored to RL agents and prioritizes early prediction of safety violations, thus providing the flexibility to implement a wide range of safety mechanisms. The use of state abstraction and ML models allows for efficient and accurate safety violation prediction, without requiring detailed environment models, making it well-suited for a broader range of model-free RL algorithms that are commonly used.

\section{Conclusion}
\label{Sec:Conclusion}
In this paper, we propose \textit{SMARLA}, a black-box safety monitoring approach for reinforcement learning agents. We rely on a machine learning model to predict safety violations of RL agents early during their execution. \added{Our approach relies on Q-values and is agnostic to the input of the RL agent.}  We employ state abstraction to reduce the state space, enabling improved learnability to predict violations. \added{ We quantitatively and qualitatively evaluate our safety monitoring approach on \added{three} widely used RL benchmarks}. Our results demonstrate the high accuracy of \textit{SMARLA} in predicting the safety violations of RL agents during their execution. Additionally, our empirical results show that such accurate prediction can be made early long before the actual occurrence of violations, allowing for timely damage prevention and mitigation. 
In future work, we intend to expand our evaluation to include additional RL case studies. Furthermore, to further improve accuracy and early predictions, our aim is to investigate the use of other types of features that consider temporal information regarding the agent’s states and actions in RL episodes. For instance, we can consider the sequence of abstract states instead of only considering their presence or absence as features.

\section*{Acknowledgements}

This work was supported by a research grant from General Motors as well as the Canada Research Chair and Discovery Grant programs of the Natural Sciences and Engineering Research Council of Canada (NSERC). Lionel Briand was partly supported by the Science Foundation Ireland grant 13/RC/2094-2.

\bibliographystyle{IEEEtran}
\bibliography{main.bib} 

\pagebreak

\begin{IEEEbiography}[{\includegraphics[width=1in,height=1.25in,clip,keepaspectratio]{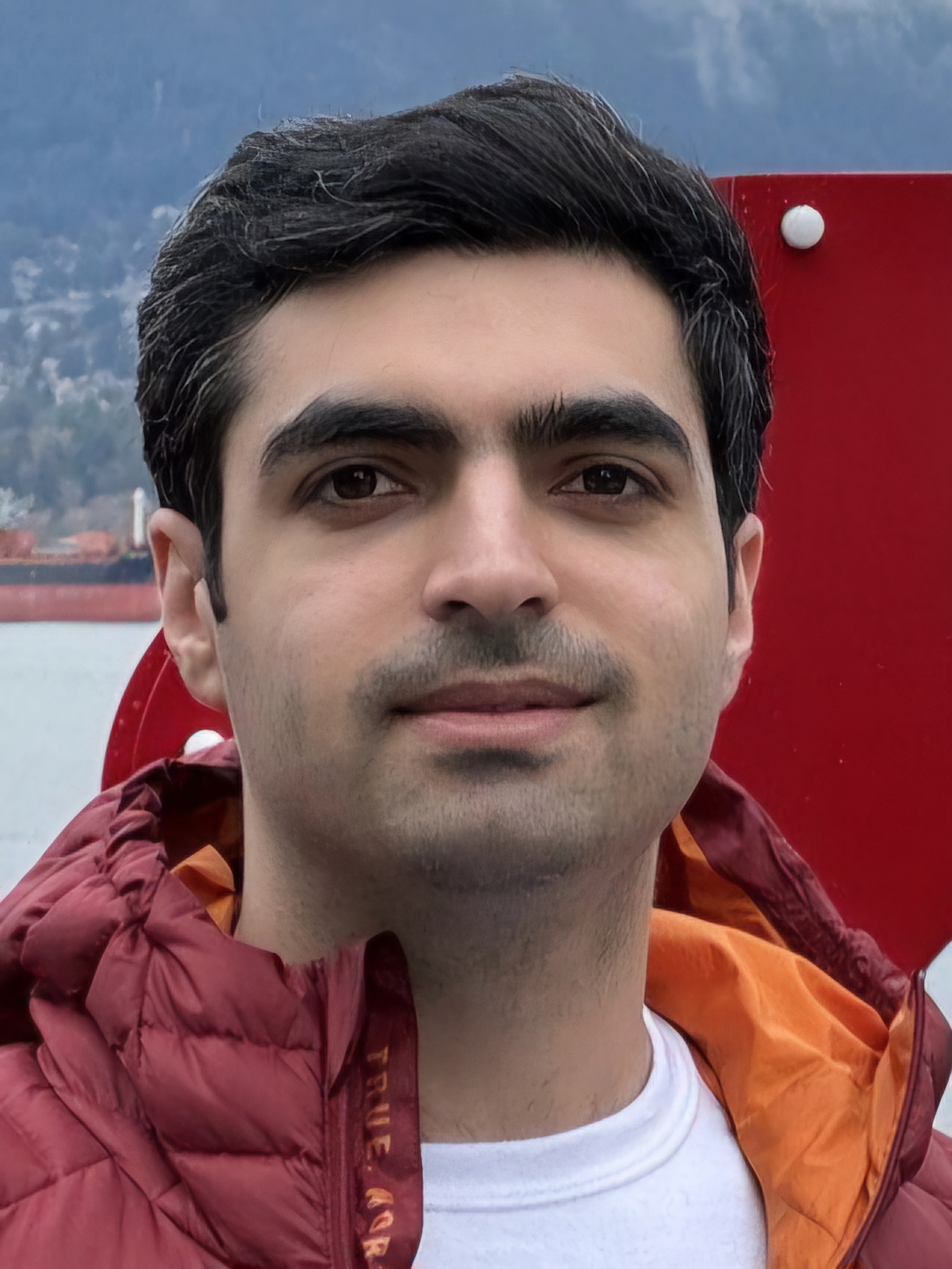}}]{Amirhossein Zolfagharian}
is a member of Nanda Lab and is currently pursuing a PhD in the School of EECS, University of Ottawa. He gained valuable practical experience from his internship at General Motors' research and development lab in the United States. Throughout his academic career, he has been the recipient of several academic awards, including a PhD admission scholarship, an international doctoral scholarship from the University of Ottawa, and an honorable award for admission to the master's program in computer science at Amirkabir University of Technology. In 2017, he was ranked as the 4th-best student among all computer science students at Amirkabir University, placing him in the top 5\% of his class in GPA. His research interests primarily focus on machine learning, empirical software engineering, and testing and verification of RL-based systems.

\end{IEEEbiography}
\begin{IEEEbiography}
[{\includegraphics[width=1in,height=1.25in,clip,keepaspectratio]{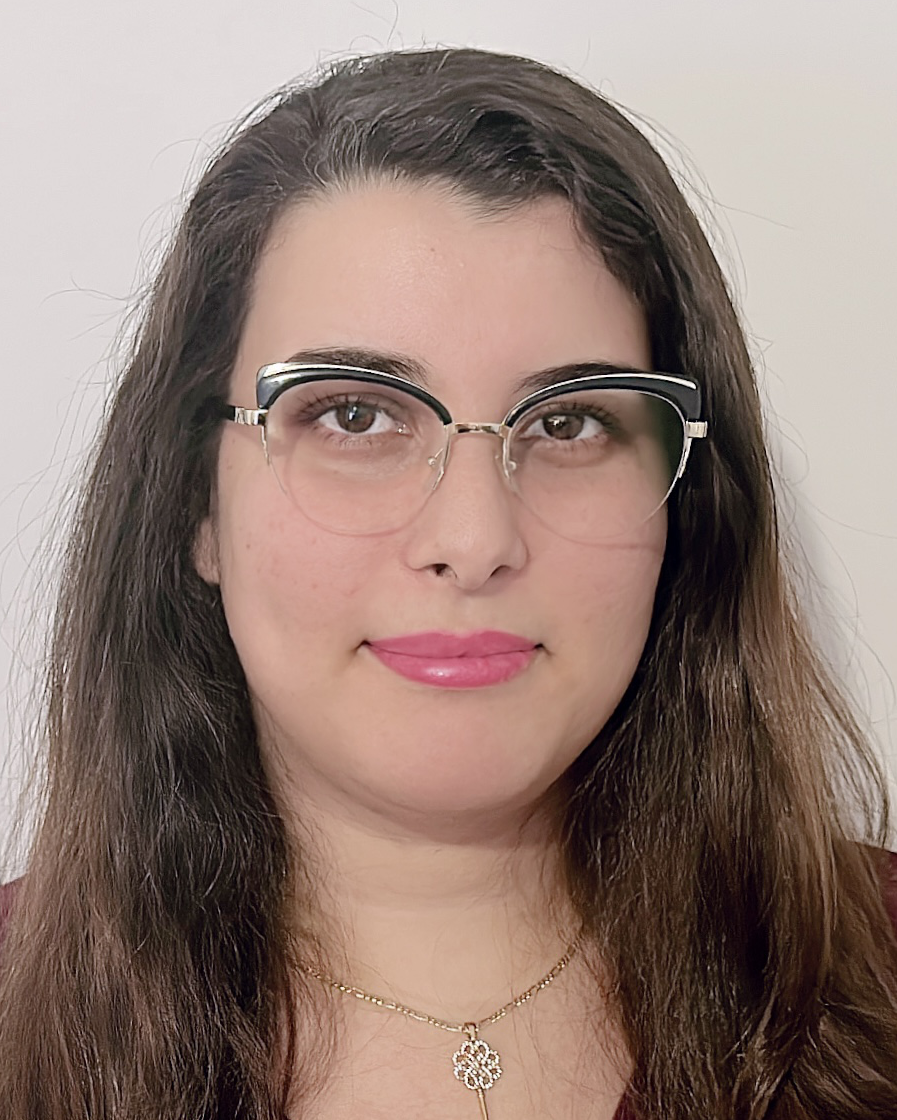}}]
{Manel Abdellatif} is a Professor of Software Engineering at the École de Technologie Supérieure (ÉTS) in Canada. She earned her PhD in Computer Science from Polytechnique Montreal, followed by a postdoctoral fellowship at the School of Electrical Engineering and Computer Science (EECS) at the University of Ottawa. She has received multiple best paper awards for her research contributions and established several impactful collaborations with industry partners. Her research focuses on trustworthy AI, service-oriented computing, and software maintenance and evolution.

\end{IEEEbiography}
 

\begin{IEEEbiography}
[{\includegraphics[width=1in,height=1.25in,clip,keepaspectratio]{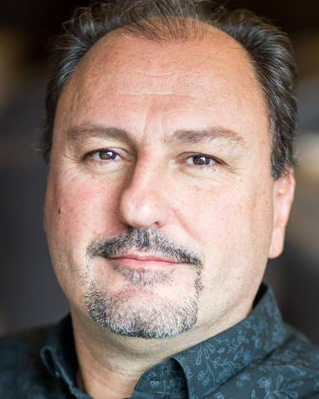}}]
    {Lionel C. Briand} is professor of software engineering and has shared appointments between (1) The University of Ottawa, Canada, and (2) The Lero SFI Centre---the national Irish centre for software research---hosted by the University of Limerick, Ireland. In collaboration with colleagues, for over 30 years, he has run many collaborative research projects with companies in the automotive, satellite, aerospace, energy, financial, and legal domains. Lionel has held various engineering, academic, and leading positions in seven countries.  He currently holds a Canada Research Chair (Tier 1) on "Intelligent Software Dependability and Compliance" and is the director of Lero, the national Irish centre for software research. Lionel was elevated to the grades of IEEE Fellow and ACM Fellow for his work on software testing and verification. Further, he was granted the IEEE Computer Society Harlan Mills award, the ACM SIGSOFT outstanding research award, and the IEEE Reliability Society engineer-of-the-year award. He also received an ERC Advanced grant in 2016 on modelling and testing cyber-physical systems, the most prestigious individual research award in the European Union and was elected a fellow of the Academy of Science, Royal Society of Canada in 2023. More details can be found at: http://www.lbriand.info.
\end{IEEEbiography}


\begin{IEEEbiography}
[{\includegraphics[width=1in,height=1.25in,clip,keepaspectratio]{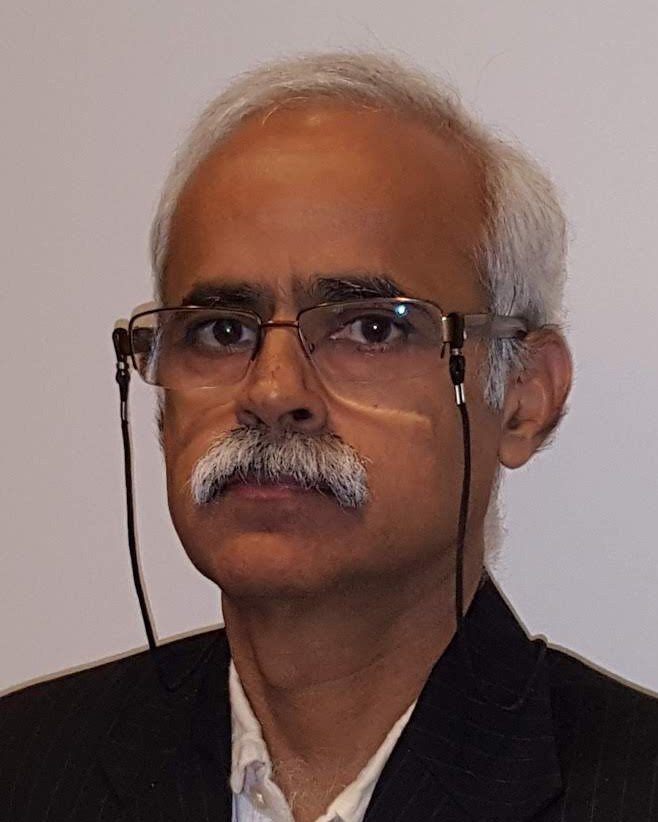}}]
    {Ramesh S}, Senior Technical Fellow,  has been with General Motors Research and Development (R\&D) for more than 15 years conducting and leading advanced research projects in the areas of model-based development of embedded systems and software, rigorous verification and validation, and more recently, AI/ML based systems. Prior to joining GM R\&D, he was a full professor in the department of Computer Science and Engineering at the Indian Institute of Technology Bombay, India, where he co-founded a Centre for Formal Design and Verification of Software. He has published more than 125 research papers in international journals and conferences and authored many patents in the areas of modeling, analysis, and verification of embedded systems and software. He has been on the program committees of several international research conferences and on the editorial boards of journals. He leads an USCAR committee and serves as an expert in ISO and SAE committees to develop guidelines for AI/ML based systems.
\end{IEEEbiography}

\end{document}